\documentclass{article}
\usepackage[margin=1in]{geometry}
\usepackage{parskip}

\usepackage[round]{natbib}

\usepackage[utf8]{inputenc} %
\usepackage[T1]{fontenc}    %
\usepackage{booktabs}       %
\usepackage{enumitem}
\setenumerate[1]{label=(\arabic*)}

\usepackage{graphicx}
\usepackage{hyperref}%
\usepackage{color}
\usepackage{amsmath,amsthm,amssymb}
\usepackage{bm}
\hypersetup{colorlinks=true, linkcolor=blue, citecolor=blue}
\usepackage[nameinlink,capitalize,noabbrev]{cleveref} %

\theoremstyle{plain}
\newtheorem{theorem}{Theorem}%
\newtheorem{lemma}{Lemma}
\newtheorem{corollary}{Corollary}
\newtheorem{proposition}{Proposition}

\newtheorem{condition}{Condition}
\crefname{condition}{Condition}{Conditions}
\Crefname{condition}{Condition}{Conditions}

\newenvironment{namedcondition}[1]
  {\renewcommand\thecondition{#1}\begin{condition}}
  {\end{condition}}

\theoremstyle{definition}

\newtheorem{assumption}{Assumption}
\crefname{assumption}{Assumption}{Assumptions}
\Crefname{assumption}{Assumption}{Assumptions}

\theoremstyle{remark}
\newtheorem{remark}{Remark}
\newtheorem{example}{Example}
\crefname{example}{Example}{Examples}
\Crefname{example}{Example}{Examples}

\newcommand{\RR}{\mathbb{R}}

\newcommand{\NN}{\mathbb{N}}
\newcommand{\ZZ}{\mathbb{Z}}
\newcommand{\PP}{\mathbb{P}}

\newcommand{\EE}{\mathbb{E}}

\newcommand{\dd}{\mathrm{d}}
\newcommand{\eps}{\varepsilon}

\newcommand{\calX}{\mathcal{X}}
\newcommand{\calY}{\mathcal{Y}}

\newcommand{\calM}{\mathcal{M}}
\newcommand{\calS}{\mathcal{S}}

\newcommand{\calH}{\mathcal{H}}

\newcommand{\calA}{\mathcal{A}}
\newcommand{\calD}{\mathcal{D}}

\DeclareMathOperator{\const}{const}
\DeclareMathOperator{\embed}{\Phi_{embed}}%
\DeclareMathOperator{\eval}{ev}
\DeclareMathOperator{\id}{id}
\DeclareMathOperator{\diam}{diam}
\DeclareMathOperator{\dist}{dist}

\DeclareMathOperator{\lip}{Lip}
\DeclareMathOperator{\colip}{coLip}

\DeclareMathOperator{\iso}{Iso}
\DeclareMathOperator{\homeo}{Homeo}

\newcommand{\gen}{\mathsf{gen}}
\newcommand{\bias}{\mathsf{bias}}
\newcommand{\var}{\mathsf{var}}

\newcommand{\ssq}[1]{[\![#1]\!]} %

\newcommand{\Himp}{\calH_{\text{imp}}}

\newcommand{\Habs}{\calH_{\text{abs}}} %
\newcommand{\Con}{\mathcal{C}}

\newcommand{\errimp}{\eps_{\text{imp}}}
\newcommand{\errmodel}{\eps_{\text{model}}}

\newcommand{\xdom}{\calX}

\newcommand{\rad}{\mathfrak{R}}
\newcommand{\erad}{\hat{\mathfrak{R}}}

\newcommand{\Lhat}{\hat{L}}
\newcommand{\hhat}{\hat{h}}
\newcommand{\fhat}{\hat{f}}

\DeclareMathOperator{\relu}{ReLU}

\newcommand{\biase}{\alpha}
\newcommand{\biasp}{\beta}
\newcommand{\varp}{\gamma}

\newcommand{\kopt}{k^\ast}

\allowdisplaybreaks

\title{\textbf{Why and When Deep is Better than Shallow: Implementation-Agnostic State-Transition Model of Deep Learning}}

\author{%
\textbf{Sho Sonoda}${}^{1,2}$ \hfill\small{\url{sho.sonoda@riken.jp}}\\
\textbf{Yuka Hashimoto${}^{3,1}$} \hfill\small{\url{yuka.hashimoto@ntt.com}}\\
\textbf{Isao Ishikawa}${}^{4,1}$ \hfill\small{\url{ishikawa.isao.5s@kyoto-u.ac.jp}}\\
\textbf{Masahiro Ikeda}${}^{5,1}$ \hfill\small{\url{ikeda@ist.osaka-u.ac.jp}}\\
\small{${}^1$\textit{RIKEN AIP} }%
\small{${}^2$\textit{CyberAgent, Inc} }%
\small{${}^3$\textit{NTT, Inc} }%
\small{${}^4$\textit{Kyoto University} }%
\small{${}^5$\textit{The University of Osaka}}\\%\hfill\phantom{a}\\
\phantom{\large{aaaaaaaaaaaaaaaaaaaaaaaaaaaaaaaaaaaaaaaaaaaaaaaaaaaaaaaaaaaaaaaaaaaaaaaa}}%
}

\date{September 26, 2026}

\begin{document}

\maketitle

\begin{abstract}
We ask when adding hidden layers improves generalization, in a model that keeps the layers fixed and varies only their number.  A hidden layer is a self-map of a state space, a depth-$k$ network composes at most $k$ hidden layers with an output layer, and depth is compared within the nested family $\mathcal H_0\subset\mathcal H_1\subset\cdots$ built from one class $F$ of hidden layers; this compares a deep network with shallower networks built from the same layers, not with wider ones.  Our message is that the statistical cost of depth is the metric entropy of the set $B(k,F)$ of compositions.  The estimation error is bounded by an entropy integral over $B(k,F)$ with constants that do not depend on the depth, and the bound is matched from below when the output layer can see the hidden states.  For Lipschitz layers on a bounded state space this entropy grows at most polynomially in $k$, and it stays bounded, or grows only like $\log k$, under contraction, equicontinuity, or nilpotent structure.  Balanced against the approximation error, this gives depths $k^\ast(n)$ that grow with the sample size, and it separates the models whose estimation error is independent of the depth from those whose estimation error grows with it.  Deep ReLU networks, unrolled solvers, and chain-of-thought computation are worked out; for the last two the number of steps is derived rather than assumed.  All statements are machine-checked in Lean~4; the formalization and its blueprint are available at \url{https://shosonoda.github.io/lean-deepgen/}.
\end{abstract}

\section{Introduction}

Deeper models often generalize better, and learning theory does not make
this automatic: every added layer enlarges the hypothesis class, and naive
complexity bounds grow exponentially in the depth \(k\).  This paper asks a
precise version of the question.  Fix a class \(F\) of hidden layers and a
class \(H\) of output layers, let \(\calH_k\) be the networks with an output
layer from \(H\) and at most \(k\) hidden layers from \(F\), and consider the
nested family \(\calH_0\subset\calH_1\subset\cdots\).  What is the
statistical cost of the depth \(k\) in this family, and when is the best
depth \(\kopt(n)\) larger than one and growing with the sample size \(n\)?
This compares a deep network with shallower networks built from the same
layers.  It is the design question for models whose layers are shared by
construction, such as unrolled solvers, iterative denoisers, samplers of
diffusion models, neural operators, and step-by-step reasoning, where the
depth is the number of iterations or steps.  It is not a comparison of deep
and wide networks at equal parameter count; width enters through the layer
class \(F\) and is held fixed when depth varies.

\begin{figure}[!b]
    \centering
    \includegraphics[width=0.55\linewidth]{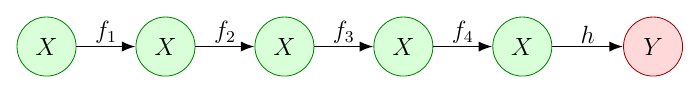}
    \caption{A network of depth \(k=4\) in the state-transition model.  The input and every hidden representation are points of one state space \(\xdom\); each hidden layer is a map \(f_i:\xdom\to\xdom\); the output layer \(h:\xdom\to\RR\) reads the answer from the last state.  Depth is the number of hidden maps composed before the output layer.}
    \label{fig:network}
\end{figure}

\paragraph{The model.}
We describe a network as a state-transition model (\cref{fig:network}).
The \emph{state space} \(\calX\) is a metric space in which the input and
all hidden representations live, for instance a bounded subset of
\(\RR^m\).  A \emph{hidden layer} is a continuous map \(f:\calX\to\calX\),
and \(F\) is the set of admissible hidden layers.  The
\emph{depth-\(k\) hidden class} \(B(k,F)\) is the set of all compositions
\(f_m\circ\cdots\circ f_1\) of at most \(k\) maps from \(F\), including the
identity.  An \emph{output layer} is a map \(h:\calX\to\RR\) that reads the
answer from the last state, and \(H\) is the set of admissible output
layers.  The \emph{hypothesis class} at depth \(k\) is
\(\calH_k:=H\circ B(k,F)=\{h\circ f:h\in H,\ f\in B(k,F)\}\).  The
\emph{target class} \(\Con\) is the set of functions the learner is compared
with; it need not be contained in any \(\calH_k\).  Only the metric of
\(\calX\) and the composition of maps in \(F\) are used, so the same
objects describe fully connected networks, unrolled solvers, and symbolic
computation.

\paragraph{Message and evidence.}
The statistical cost of depth is the metric entropy of \(B(k,F)\), the
logarithm of the number of maps in \(B(k,F)\) that can be told apart at a
given resolution.  The estimation error is independent of the depth when
composition creates no new distinguishable maps, and it grows with the
depth when it does.  Three results
support this message.
\begin{enumerate}
\item
(\Cref{sec:bounds}.)  Under three standing assumptions, the
depth-dependent part of the estimation error is bounded by an entropy
integral over \(B(k,F)\), with a constant that depends only on the
output-layer class and not on the depth, and when the output layer can
tell the hidden states apart on the sample the same entropy appears in a
lower bound (\cref{thm:hidden-decomp}, \cref{cor:matching-main}).
\item
(\Cref{sec:growth-conditions}.)  The entropy of \(B(k,F)\) has a universal
upper bound, the layerwise covering bound: for Lipschitz parametric layers
on a bounded state space it grows at most polynomially in \(k\), and the
exponentially growing Lipschitz constant of a composition enters only
through its logarithm (\cref{prop:envelope}).  Five mechanisms decide where
a layer class sits below this bound: contraction, equicontinuity, and
nilpotent control make the entropy constant or logarithmic in \(k\),
separation makes it linear, and memory-preserving expansion attains the
bound (\cref{thm:mechanisms-main}).
\item
(\Cref{sec:optdepth,sec:discussion}.)  Balancing the resulting variance
profiles against the approximation error gives depths \(\kopt(n)\) that
grow with \(n\) (\cref{cor:balancing-main}).  If the estimation bound is
independent of the depth, the model may be run to convergence; if it grows
polynomially, the depth has to be balanced against the bias.  Three
examples carry this out: for deep ReLU networks the Lipschitz constant of
a layer selects the mechanism, and for unrolled solvers and
chain-of-thought computation both sides of the balance are derived within
the framework, so the step counts are consequences rather than inputs
(\cref{ex:relu-main,ex:cot-main}; unrolled solvers in \cref{sec:app-ode}).
\end{enumerate}

\paragraph{Background and position.}
Two lines of work motivate the model.  The first is the theory of depth in
deep ReLU networks, summarized below, which established sharp rates for
one architecture but not a notion of depth that survives a change of
architecture.  The second is the family of models in which depth is an
iteration count: unrolled optimization and fixed-point solvers, deep
equilibrium models, neural ODEs, the samplers of diffusion and
flow-matching models, Picard-type neural operators, and chain-of-thought
reasoning in language models.  For these models the layer is fixed by
design and the question ``how many steps'' is asked at deployment time,
so a theory that keeps the layer class fixed and varies only the depth
speaks to them directly; \cref{sec:lit-overview} reviews both lines and
places the known ReLU results in the coordinates of this paper, and
\cref{sec:examples-regime} works out the teacher--student and
neural-operator examples that connect them.

\paragraph{Relation to the theory of deep ReLU networks.}
Most rigorous results on depth are proved for one implementation, the
fully connected ReLU network: depth separation in approximation
\citep{Eldan2016,Telgarsky2016,Yarotsky2017,Siegel2023}, VC-dimension,
norm-based, and compression-based estimation bounds
\citep{Bartlett2019,Bartlett2017,Golowich2018,Arora2018compression,Suzuki2020compression},
and minimax rates on smoothness classes and hierarchical compositions
\citep{Schmidt-Hieber2020,suzuki2018besov-adaptivity}.  These results are
sharp, but their depth dependence is read off an explicit parameterization,
and a change in the layer structure, the state space, or the parameter
sharing requires a new proof.  Our framework isolates depth in \(B(k,F)\),
separately from the layer class \(F\); the known ReLU rates then appear as
one case of the analysis (the polynomial profile), and the question of
what makes depth cheap is answered by the geometry of composition rather
than by the parameter count.  A detailed comparison with the literature is
given in \cref{sec:lit-overview}.

\paragraph{Roadmap.}
\Cref{sec:setting} fixes the learning problem.  \Cref{sec:bounds} reduces
the cost of depth to the entropy of \(B(k,F)\); the precise statements and
proofs are in \cref{sec:proof-bv,sec:proof-hidden,sec:sudakov-proof}.
\Cref{sec:growth-conditions} classifies the growth of this entropy; the
formal conditions, examples, and proofs are in
\cref{sec:growth-conditions-proof,sec:proof-envelope,sec:proof-profiles}.
\Cref{sec:optdepth} derives the balancing depths; the four regimes and
their calculations are in \cref{sec:proof.tradeoff}.
\Cref{sec:discussion} explains how to apply the framework to a concrete
model.  The worked examples are in \cref{sec:examples-regime} (teacher--student
models, neural operators, and the implementation layer),
\cref{sec:app-cot} (chain-of-thought computation), \cref{sec:app-ode}
(unrolled solvers and diffusion samplers), and \cref{sec:app-relu} (deep
ReLU networks, whose Lipschitz constant selects the growth mechanism).  Every statement of the paper is formalized and
machine-checked in Lean~4; the source and a blueprint linking each
statement to its formal counterpart are available at
\url{https://shosonoda.github.io/lean-deepgen/}.

\section{Setting}
\label{sec:setting}

\paragraph{Learning problem.}
We observe an i.i.d.\ sample \(\calD=((X_1,Y_1),\dots,(X_n,Y_n))\) from a
distribution \(P\) on \(\calX\times\RR\), and write \(S=(X_1,\dots,X_n)\)
for the inputs.  The loss \(\ell:\RR\times\RR\to[0,b]\) is bounded and
Lipschitz in the prediction (\cref{ass:loss-main} below; clipped squared
loss, hinge, ramp, or logistic loss on a bounded range; cross-entropy
needs clipping, \cref{rem:loss-app}).  The risk and the empirical risk are
\(L[f]:=\EE\,\ell(f(X),Y)\) and \(\Lhat[f]:=n^{-1}\sum_i\ell(f(X_i),Y_i)\).
The learner picks an \(\eta\)-approximate empirical risk minimizer
\(\fhat\in\calH_k\), that is, \(\Lhat[\fhat]\le\inf_{f\in\calH_k}\Lhat[f]+\eta\),
and outputs an implementation \(\hhat:=\embed(\fhat)\), where the
\emph{implementation map} \(\embed\) sends each abstract predictor to a
concrete one, for instance a finite-width network.  Two errors are attached
to the depth \(k\):
\[
\errimp(k):=\sup_{f\in\calH_k}\|f-\embed(f)\|_\infty,
\qquad
\errmodel(k):=\inf_{f\in\calH_k}L[f]-\inf_{c\in\Con}L[c],
\]
the \emph{implementation error} and the \emph{approximation error}.  The
latter is a gap between risks; a uniform approximation rate transfers to it
through the Lipschitz loss (\cref{rem:approx-vs-risk-app}),
\begin{equation}
\errmodel(k)\le\beta_\ell\,\sup_{c\in\Con}\inf_{f\in\calH_k}\|f-c\|_\infty .
\label{eq:approx-transfer}
\end{equation}

\paragraph{Distances, covering numbers, and complexity.}
Two distances between hidden maps \(f,g:\calX\to\calX\) are used: the
uniform distance \(d_\infty(f,g):=\sup_{x}d(f(x),g(x))\) and the empirical
distance \(d_S(f,g):=(n^{-1}\sum_id(f(X_i),g(X_i))^2)^{1/2}\), which only
looks at the sample and satisfies \(d_S\le d_\infty\).  For a set \(A\) of
maps and a resolution \(\eps>0\), the \emph{covering number} \(N(A,d,\eps)\)
is the smallest number of balls of radius \(\eps\) (in the distance \(d\))
needed to cover \(A\); it counts the maps in \(A\) that can be told apart at
resolution \(\eps\), and \(\log N(A,d,\eps)\) is the \emph{metric entropy}.
For a class \(G\) of real-valued functions, the empirical Rademacher
complexity
\[
\erad_S(G):=\EE_\sigma\Bigl[\sup_{g\in G}\frac1n\sum_{i=1}^n\sigma_ig(X_i)\Bigr],
\]
with independent random signs \(\sigma_i\), measures how well \(G\) can fit
random signs on the sample; it is the standard measure of statistical
complexity in uniform-convergence bounds.  Finer conventions (external
covering numbers, the semigroup view of \(B(k,F)\)) are recorded in
\cref{sec:app-remarks}.

\paragraph{Standing assumptions.}
Three assumptions are used throughout; each is stated once here with the
place where it enters.

\begin{assumption}[Bounded Lipschitz loss]
\label{ass:loss-main}
The loss \(\ell:\RR\times\RR\to[0,b]\) is measurable and
\(\beta_\ell\)-Lipschitz in its first argument, with \(b\) and
\(\beta_\ell\) fixed with the loss, so that neither depends on the depth.
It enters the bias--variance decomposition, \cref{thm:hidden-decomp}(i).
\end{assumption}

\begin{assumption}[Output layer does not amplify hidden distances]
\label{ass:sg-increment-main}
For a hidden map \(f\) let \(Z_f:=\sup_{h\in H}n^{-1}\sum_i\sigma_ih(f(X_i))\)
be the Rademacher average of the output layers applied after \(f\).  There
is a constant \(L_H\), depending only on \(H\) and the metric of \(\calX\),
such that for all compositions \(f,g\) of maps from \(F\), of any depth,
and all \(t>0\),
\(\PP_\sigma(|Z_f-Z_g|>t)\le2\exp\bigl(-nt^2/(2L_H^2d_S(f,g)^2)\bigr)\).
It holds with \(L_H\) equal to the Lipschitz constant of the feature map
for norm-bounded linear output layers on any feature map, and for finite
classes of Lipschitz output layers, including bounded scalarizations of a
softmax head (\cref{sec:proof-hidden}).  Its point is that \(L_H\) is fixed
with the output layer and is the same at every depth: the hidden layers
stay inside \(B(k,F)\) and never enter \(L_H\).  It enters the
hidden--output decomposition, \cref{thm:hidden-decomp}(ii).
\end{assumption}

\begin{assumption}[Bounded state space, Lipschitz layers]
\label{ass:geometry-main}
The state space has finite diameter \(D:=\diam(\calX)\), and every map in
\(F\) is \(\Lambda\)-Lipschitz for some \(\Lambda\ge0\).  Boundedness is
provided in practice by bounded activations, normalization layers, or
clipping.  It enters the growth analysis of
\cref{sec:growth-conditions} and the profiles of \cref{sec:optdepth}.
\end{assumption}

\section{The cost of depth is the entropy of \texorpdfstring{\(B(k,F)\)}{B(k,F)}}
\label{sec:bounds}

This section proves the first claim: the statistical cost of depth is the
metric entropy of \(B(k,F)\).  The argument has three steps, collected in
one theorem.  The generalization error splits into an implementation, an
approximation, and a statistical term; the statistical term splits into an
output-layer part, which does not depend on the depth, and a hidden-layer
part, which is an entropy integral over \(B(k,F)\); and a converse shows
that this entropy is paid whenever the output layer can see the hidden
states.

\begin{theorem}[The statistical cost of depth]
\label{thm:hidden-decomp}
Fix \(k\ge0\) and a sample \(S\).
\begin{enumerate}
\item[(i)]
(Bias--variance decomposition.)  Under \cref{ass:loss-main}, if
\(\calH_k\) is separable in the uniform norm and pointwise bounded, then
for every \(\delta\in(0,1)\), with probability at least \(1-\delta\), the
\(\eta\)-approximate empirical risk minimizer satisfies
\begin{equation}
L[\hhat]-\inf_{c\in\Con}L[c]
\le
\beta_\ell\,\errimp(k)+\errmodel(k)+\eta+4\beta_\ell\,\erad_S(\calH_k)
+6b\sqrt{\tfrac{2\log(4/\delta)}{n}} .
\label{eq:bv-informal}
\end{equation}
\item[(ii)]
(Hidden--output decomposition.)  Under \cref{ass:sg-increment-main},
\begin{equation}
\erad_S(\calH_k)
\le
\erad_S(H)
+\frac{12L_H}{\sqrt n}\,\mathsf V_k(S),
\qquad
\mathsf V_k(S):=\int_0^{\diam_S(B(k,F))}\sqrt{\log N\bigl(B(k,F),d_S,\eps\bigr)}\,\dd\eps .
\label{eq:hidden-output-main}
\end{equation}
\item[(iii)]
(Converse under output-visible separation.)  If for each map \(g\) in
\(B(k,F)\) some output layer \(h_g\in H\) realizes the hidden geometry on
the sample, \(\|h_g\circ g-h_{g'}\circ g'\|_S\ge\kappa\,d_S(g,g')\) and
\(\|h_g\circ g\|_{S,\infty}\le R_{\rm out}\) for all \(g,g'\), then, with
\(M\) the packing number and \(c\) a universal constant,
\begin{equation}
\erad_S(\calH_k)
\ge
c\sup_{\eps>0}\min\Bigl\{\kappa\eps\sqrt{\tfrac{\log M(B(k,F),d_S,2\eps)}{n}},\ \tfrac{\kappa^2\eps^2}{R_{\rm out}}\Bigr\}.
\label{eq:sudakov-informal}
\end{equation}
\end{enumerate}
The precise statements, with all regularity conditions, are
\cref{thm:bv} in \cref{sec:proof-bv}, the restatement of (ii) in
\cref{sec:proof-hidden}, and \cref{thm:sudakov-type} in
\cref{sec:sudakov-proof}.
\end{theorem}

\paragraph{Reading the theorem.}
Part (i) says that the statistical cost of depth is entirely carried by
\(\erad_S(\calH_k)\): the implementation error is controlled separately
(\cref{prop:implementation} in \cref{sec:examples-regime}) and the
approximation error is a property of the target class.  Part (ii) removes
the output layer from this cost: the first term is the complexity of the
output layer alone, and the second is the \emph{entropy integral} of
\(B(k,F)\), which adds up, over all resolutions \(\eps\) up to the diameter
of \(B(k,F)\) on the sample, the square root of the number of maps in
\(B(k,F)\) that are distinguishable at resolution \(\eps\).  It is Dudley's
chaining bound \citep[Chapter~8]{Vershynin2018} applied to the process
\(f\mapsto Z_f\), whose increments are controlled by
\cref{ass:sg-increment-main}; the constant is \(12L_H\).  Part (iii) is a
Sudakov-type minoration \citep[Chapter~7]{Vershynin2018}: the packing
number \(M(B(k,F),d_S,2\eps)\) is the largest number of maps in \(B(k,F)\)
at mutual distance at least \(2\eps\) on the sample, and it is
comparable to the covering number, so (ii) and (iii) bound the hidden
term by the same quantity from both sides.  The converse concerns the
complexity of the class, not the risk of every algorithm; it is informative
exactly when the output layer sees the hidden states, and when hidden
representations collapse the entropy of \(B(k,F)\) is not paid and the
upper bound is correspondingly loose (\cref{sec:sudakov-collapse}).

\begin{corollary}[Matching depth dependence]
\label{cor:matching-main}
Let the hypotheses of \cref{thm:hidden-decomp}(iii) hold for every
\(k\) with \(\kappa,R_{\rm out}\) independent of \(k\), and fix a resolution
\(\eps_0>0\).  If \(B(k,F)\) contains \(e^{\Theta(k)}\) maps that are
\(2\eps_0\)-separated on the sample, then \(\erad_S(\calH_k)\gtrsim\sqrt{k/n}\)
for \(n\gtrsim k\); if it contains \(k^{\Theta(1)}\) such maps, then
\(\erad_S(\calH_k)\gtrsim\sqrt{\log k/n}\) for \(n\gtrsim\log k\).  When the
covering numbers obey the matching upper bounds of \cref{sec:growth-conditions},
the hidden term therefore has the same depth dependence, \(\sqrt k\) or
\(\sqrt{\log k}\), in the upper and the lower bound
(\cref{cor:matching} in \cref{sec:sudakov-proof}).
\end{corollary}

The separation in (iii) has to be visible on the sample, not only in the
uniform distance: \cref{sec:app-cot} gives a family whose maps are
separated in \(d_\infty\) but not in \(d_S\), for every sample.  After this
section, everything the depth does to the estimation error is contained in
two quantities, the covering numbers of \(B(k,F)\) at every resolution and
its diameter; the next section determines how they grow with \(k\).

\section{How the entropy of \texorpdfstring{\(B(k,F)\)}{B(k,F)} grows with the depth}
\label{sec:growth-conditions}

This section proves the second claim.  By \cref{thm:hidden-decomp}(ii),
the depth-dependent part of the estimation bound is
\(\var(k,n):=12L_H\mathsf V_k(S)/\sqrt n\), and its growth with \(k\) is the
growth of the entropy integral \(\mathsf V_k(S)\); we call the function
\(k\mapsto\mathsf V_k(S)\), up to constants, the \emph{profile} of the
variance bound.  Since \(d_S\le d_\infty\), it suffices to bound the
covering numbers of \(B(k,F)\) in the uniform distance together with its
diameter.  Two results do this: a bound that holds for every Lipschitz
layer class (\cref{prop:envelope}), and a list of the structures that
lower the growth below it or attain it (\cref{thm:mechanisms-main}).
Throughout, \cref{ass:geometry-main} is in force.

\subsection{A layerwise covering bound: depth costs at most polynomially}
\label{sec:envelope}

\begin{theorem}[Layerwise covering bound]
\label{prop:envelope}
Let every map in \(F\) be \(\Lambda\)-Lipschitz, \(\Lambda\ge0\).  Then for
every \(k\ge0\) and \(\eps>0\),
\begin{equation}
N\bigl(B(k,F),d_\infty,\eps\bigr)
\le
1+k\,N\Bigl(F,d_\infty,\frac{\eps}{1+\Lambda+\cdots+\Lambda^{k-1}}\Bigr)^{k}.
\label{eq:envelope}
\end{equation}
Consequently, if the state space is bounded and \(F\) is parametric, that
is, \(N(F,d_\infty,\eps)<\infty\) with
\(\log N(F,d_\infty,\eps)\le p\log(1/\eps)+O(1)\), then
\(\mathsf V_k(S)=O(\sqrt{kp\log k})\) when \(\Lambda\le1\) and
\(\mathsf V_k(S)=O(k\sqrt{p\log\Lambda})\) when \(\Lambda>1\); if instead
\(\log N(F,d_\infty,\eps)\le c\,\eps^{-q}\) with \(q<2\), then
\(\mathsf V_k(S)=O(k^{(1+q)/2})\) for \(\Lambda\le1\) and
\(O(k^{1/2}\Lambda^{qk/2})\) for \(\Lambda>1\)
(explicit constants in \cref{cor:envelope-profiles}, \cref{sec:proof-envelope}).
\end{theorem}

The proof is the telescoping argument that also controls the implementation
error: replacing each of \(k\) layers by a nearby one moves the composition
by at most \(\eps\).  Three consequences are worth stating.  First, for
parametric layers on a bounded state space, depth costs at most
polynomially, and the exponentially growing Lipschitz constant
\(\Lambda^k\) of a composition enters the covering numbers only through
its logarithm; for norm-bounded ReLU networks this is the order of the
VC-dimension bounds of \citet{Bartlett2019} up to a factor \(\sqrt k\) and
logarithms, obtained here without any piecewise-linear structure and hence
for every layer class.  Second, the bound is attained: a construction whose
layers write a symbol, shift the memory, and expand it (mechanism (v)
below) reaches the \(k^2\) entropy of the parametric case and the
exponential entropy of the nonparametric case
(\cref{sec:proof-envelope}).  Third, \(\Lambda^k\) does appear on an
unbounded state space, where the reachable states lie in a ball of radius
\(\Lambda^k\) times that of the inputs (\cref{lem:reachable-radius}), so
that \(D_k(S)\) can grow like \(\Lambda^k\); this is the product of
layerwise norms of norm-based bounds \citep{Neyshabur2015,Bartlett2017},
it enters only through the diameter, and bounded activations,
normalization layers \citep{Ioffe2015,Ba2016layernorm}, or a projection
onto a bounded set remove it.

\subsection{Structures below the layerwise bound, and mechanisms that attain it}

Depth becomes cheaper than the layerwise bound when composition does not
create many well-separated maps.  Whether it does is not decided by the
compactness of \(\calX\) or by the algebra of \(F\) alone, but by how
composition interacts with the metric.  The following theorem lists the
mechanisms; the formal conditions, named P1, P1', P2, E1, E2, and E3, are
stated and proved with examples in \cref{sec:growth-conditions-proof},
and the theorem summarizes them.

\begin{theorem}[Growth mechanisms]
\label{thm:mechanisms-main}
Let \(N_k(\eps):=N(B(k,F),d_\infty,\eps)\).
\begin{enumerate}
\item[(i)]
(Saturation by equicontinuity, P1; \cref{sec:app-saturation}.)  If
\(\calX\) is compact and the compositions of maps from \(F\) are
equicontinuous, in particular if the maps in \(F\) are non-expanding, then
\(\sup_kN_k(\eps)<\infty\) for every \(\eps>0\).
\item[(ii)]
(Saturation by contraction, P1'; \cref{sec:app-saturation}.)  If the maps
in \(F\) are uniformly contractive, with a nonempty invariant set \(A\)
and a bounded absorbing set, then for every \(\eps>0\) and every \(k\),
\(N_k(\eps)\le N(A,\eps/2)+\sum_{\ell<m(\eps)}N(F^\ell,d_\infty,\eps)\) with
\(m(\eps)=O(\log(1/\eps))\), a bound that does not depend on \(k\).
\item[(iii)]
(Polynomial growth by nilpotent control, P2; \cref{sec:app-polynomial}.)
If the maps in \(B(k,F)\) are images, under a Lipschitz action, of the
ball of radius \(O(k)\) in a group of polynomial growth of degree \(D\),
such as translations, shears, or upper-triangular matrices, then
\(N_k(\eps)\le C(1+k/\eps)^D\) and \(D_k(S)=O(k)\), with
\(D_k(S)\le\diam(\calX)\) on a compact state space.
\item[(iv)]
(Exponential growth by separation, E1 and E2; \cref{sec:app-exponential}.)
If \(F=\{f_1,\dots,f_r\}\) and distinct compositions of the same length are
separated by a fixed amount at a common point (E1), or \(F\) satisfies the
ping-pong coding condition, expanding on disjoint chambers and resetting
elsewhere, with marker--anchor separation \(\alpha\) (E2), then
\(N_k(\eps)\ge r^k\) for all \(k\) and all sufficiently small \(\eps\).
\item[(v)]
(Attainment of the layerwise bound by memory-preserving expansion, E3;
\cref{sec:app-memory}.)  There is a class \(F\) of \(\lambda\)-Lipschitz
maps whose layers write a symbol, shift the stored symbols, and expand
them, for which \(\log N_k(\eps)\asymp k^2\) when the written symbols range
over a finite-dimensional set, and \(\log N_k(\eps)\asymp\lambda^{pk}\)
when they range over a set of entropy \(\delta^{-p}\).
\end{enumerate}
\end{theorem}

The mechanisms explain one another.  Contraction and equicontinuity
saturate because Ascoli's theorem makes the set of all compositions totally
bounded: after a few layers no new distinguishable maps appear
(\cref{sec:app-saturation}).  Nilpotent control saturates nothing but
organizes the compositions by a group in which the number of elements at
distance \(k\) grows polynomially; these layers are realized exactly by
affine ReLU layers, so the mechanism is not restricted to abstract
examples.  Freeness of the compositions is not enough for a lower bound,
because contracting maps may collapse many compositions to nearly the same
map; what is needed is a probe point (\cref{lem:probes-packing}) at which
distinct compositions stay apart, and expanding maps with resets provide
one probe per composition.  Expansion alone still does not create growth
beyond \(r^k\), because later layers can erase what earlier layers
distinguished; growth beyond \(\sqrt k\) in the profile requires a memory,
and with a memory the entropy accumulates across layers up to the
layerwise bound.  Symbolic shifts with guarded steps are the basic example
of (iv) (\cref{ex:e2-pingpong-subshift}), and branching symbolic
computation is of this kind in \(d_\infty\), although its empirical entropy
is much smaller (\cref{sec:app-cot}).

\Cref{prop:profiles} in \cref{sec:proof-profiles} converts each mechanism,
with its diameter bound, into the profile of \(\mathsf V_k(S)\): constant
under (i) and (ii), \(\sqrt{\log k}\) under (iii) and for branching
computation on any sample, \(\sqrt k\) under (iv) and for the layerwise
bound of non-expanding parametric layers, \(k\) for the layerwise bound of
expanding parametric layers, which (v) attains, and exponential for
expanding nonparametric layers (listed by mechanism in
\cref{sec:proof-profiles}).  All these bounds are stated in \(d_\infty\)
and transferred to \(d_S\) through \(d_S\le d_\infty\); the transfer is
one-sided, and \cref{sec:app-cot} shows a family whose entropy grows like
\(k\) in \(d_\infty\) but only like \(\log k\) in \(d_S\), for every sample.
These five profiles are the variance side of the balance in the next
section.

\section{Balancing the bias against the estimation error}
\label{sec:optdepth}

\begin{figure}[t]
    \centering
    \includegraphics[width=\linewidth]{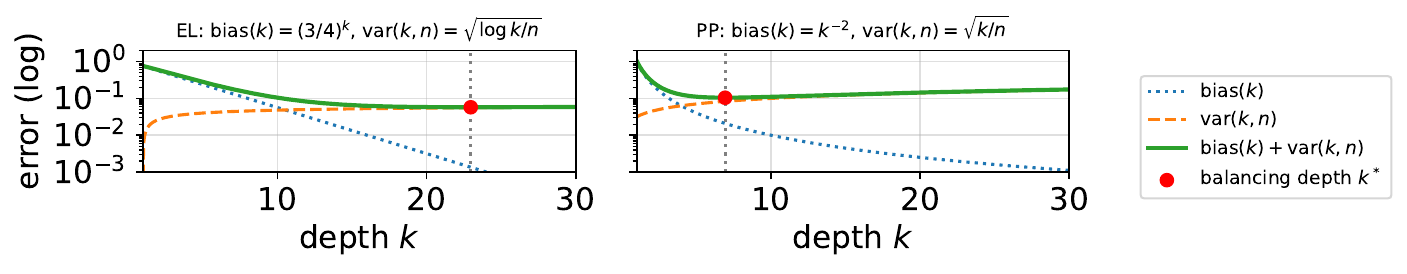}
    \vspace{-6pt}
    \caption{Balancing the bias (dotted) and the variance bound (dashed) in the depth \(k\); the solid curve is their sum and the red dot its minimizer, the balancing depth \(\kopt\); \(n=1000\), same axes in both panels.  Left, EL: bias \((3/4)^k\), as for a contractive iteration with contraction factor \(3/4\), and the logarithmic variance profile \(\sqrt{\log k/n}\); \(\kopt\) is a threshold, and deeper models are as good in the bound.  Right, PP: bias \(k^{-2}\), as for an order-\(2\) discretization or a Jackson rate with \(s/d=2\), and the polynomial profile \(\sqrt{k/n}\); depth is a budget, and the bound increases beyond \(\kopt\).  The other two combinations are derived in \cref{sec:proof.tradeoff}.}
    \label{fig:tradeoff-main}
\end{figure}

This section proves the third claim: the profile of the variance bound,
balanced against a law for the approximation error, gives a depth
\(\kopt(n)\) that grows with the sample size, and the kind of profile
decides whether the depth has to be balanced at all.  Write
\(\bias(k):=\errmodel(k)\) for the approximation error, which decreases with
the depth and depends on the target class, and
\(\gen(k,n)\lesssim\bias(k)+\var(k,n)\) for the excess-risk bound of
\cref{thm:hidden-decomp}(i) with the depth-independent terms omitted.
Since \(\bias(k)\) is nonincreasing and \(\var(k,n)\) is nondecreasing in
\(k\), the bound is minimized, up to a factor \(2\), at the depth
\(\kopt=\kopt(n)\) where the two terms are of the same order
(\cref{lem:balancing}).  Two laws for each term cover the mechanisms of
\cref{sec:growth-conditions} and the approximation results of the
examples.  The bias decays exponentially, \(\bias(k)=e^{-\Theta(\alpha k)}\),
when the target is itself an iteration that converges geometrically, as for
contractive iterations, or polynomially, \(\bias(k)=\Theta(k^{-\beta})\), as
for discretization errors and smoothness classes.  The variance profile is
logarithmic, \(\var(k,n)=O(\sqrt{\log k/n})\), which also covers the
saturated case, or polynomial, \(\var(k,n)=O(\sqrt{k^\gamma/n})\) with
\(\gamma=1\) for the \(\sqrt k\) profile and \(\gamma=2\) for the \(k\)
profile.  We name the combinations by two letters, the bias first: EL is
exponential bias with logarithmic variance, PP is polynomial bias with
polynomial variance (\cref{fig:tradeoff-main}); the mixed cases EP and PL
are treated in \cref{sec:proof.tradeoff}.

\begin{corollary}[Profiles and balancing depths]
\label{cor:balancing-main}
Under \cref{ass:loss-main,ass:sg-increment-main,ass:geometry-main}, the
variance bound \(\var(k,n)\) of \cref{thm:hidden-decomp} has, up to
constants, the profile \(n^{-1/2}\) under the saturation mechanisms of
\cref{thm:mechanisms-main}(i)--(ii), \(\sqrt{\log k/n}\) under nilpotent
control (iii), \(\sqrt{k/n}\) under the layerwise bound of
\cref{prop:envelope} for non-expanding parametric layers and under the
separation mechanisms (iv), and \(k/\sqrt n\) under the layerwise bound
for expanding parametric layers, which (v) attains
(\cref{prop:profiles} in \cref{sec:proof-profiles}).  Balanced against the
two bias laws, the balancing depth and the resulting rate are
\(\kopt=\frac{\log n}{2\alpha}(1+o(1))\) and
\(\gen(\kopt,n)\asymp\sqrt{\log\log n/n}\) in the EL case, and
\(\kopt\asymp n^{1/(2\beta+\gamma)}\) and
\(\gen(\kopt,n)\asymp n^{-\beta/(2\beta+\gamma)}\) in the PP case, with the
mixed cases EP and PL and the explicit constants in
\cref{sec:proof.tradeoff}.
\end{corollary}

In both cases the balancing depth grows with the sample size: larger
datasets support deeper networks.  An exponentially decaying bias asks for
a logarithmic depth, a polynomially decaying bias for a polynomial depth.

\paragraph{Depth-independent and depth-dependent estimation error.}
The two panels of \cref{fig:tradeoff-main} differ in kind, not only in
rate.  When the entropy integral saturates, \(\var(k,n)\) is bounded by a
quantity that does not depend on \(k\), so the bound is nonincreasing in the
depth: \(\kopt\) is a threshold rather than an optimum, every depth beyond
it is as good, and the model may be run to convergence.  This is the case
of contractive iterations, fixed-point solvers, iterative denoisers, and
append-only symbolic computation (\cref{sec:app-ode,sec:app-cot}).  When
the profile is logarithmic, the estimation bound grows so slowly that
\(\kopt\) is set by the bias.  When the profile is polynomial, which is
the case of the layerwise bound and hence of generic parametric layers,
the estimation bound grows with the depth and has to be balanced against
the bias, and deeper is better only as far as the bias improves.
Width is held fixed throughout; it enters the one-layer covering
numbers, with \(p\) of the order of the number of parameters per layer, the
output-layer class, the bias, and the implementation error
(\cref{rem:width}).  The three results are now in place; the next section
applies them to three architectures, and in two of them both sides of
the balance are derived rather than assumed.

\section{Using the framework}
\label{sec:discussion}

\subsection{A recipe for a given architecture}

To apply the results to an architecture one has to (1) choose the state
space, the layer class \(F\), and the output-layer class \(H\), and check
\cref{ass:sg-increment-main}, which holds for fixed linear or finite
output layers; (2) bound the covering numbers and the diameter of
\(B(k,F)\), by the layerwise bound or by one of the mechanisms of
\cref{thm:mechanisms-main}, which gives the variance profile; (3) bound the
approximation error through \eqref{eq:approx-transfer}; (4) control the
implementation error of the concrete network (\cref{prop:implementation}
in \cref{sec:examples-regime}); and (5) balance the two terms.  Step (2)
decides the cost of depth, and it depends on the geometry of the layers,
not on their parameter count.  We carry the recipe out for fully connected
ReLU networks, where the layer class is given and the framework reads off
the cost of depth, and for chain-of-thought computation, where the layers
are shared by design, the depth is a number of steps, and both the profile
and the bias are derived, so that the number of steps a sample of size
\(n\) justifies is a consequence rather than an input.  Unrolled solvers
and diffusion samplers are of the second kind and are worked out in
\cref{sec:app-ode}: contractive iterations are in the EL case with
\(\kopt\asymp\log n\), and fixed-horizon integrators in the PL case with
\(\kopt\asymp n^{1/(2p)}\).  Each example comes with a computed figure that
shows the same quantity, the number of compositions in \(B(k,F)\) that are
distinguishable at a fixed resolution, as a function of the depth;
\cref{sec:examples-regime} lists what is derived and what is assumed.

\subsection{Fully connected ReLU networks}
\label{sec:discussion-relu}

\begin{example}[Deep ReLU networks with clipped states]
\label{ex:relu-main}
The state space is \([0,1]^m\), a hidden layer is one ReLU block
\(x\mapsto\Pi_{[0,1]^m}(W\sigma(Ax+b)+c)\) with norm-bounded weights and
clipping, and the output layer is linear.  If the blocks are
\(\Lambda\)-Lipschitz with \(\Lambda<1\), the profile saturates (EL with
\(\kopt\asymp\log n\)); if they are non-expanding and commute, it is
logarithmic; if they are only Lipschitz, the layerwise bound gives the
polynomial profile (PP with \(\kopt\asymp n^{1/(2\beta+\gamma)}\)), and
for \(\Lambda>1\) this bound is attained by ping-pong blocks
(\cref{prop:relu-regimes} in \cref{sec:app-relu}).
\end{example}

\begin{figure}[t]
    \centering
    \includegraphics[width=\linewidth]{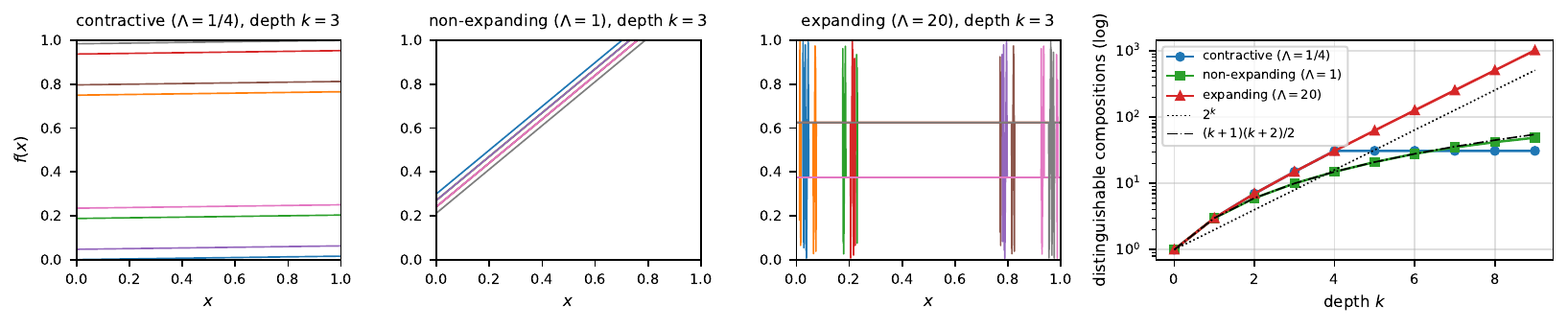}
    \vspace{-8pt}
    \caption{Three classes of two ReLU maps of \([0,1]\) with Lipschitz constants \(1/4\), \(1\), and \(20\).  Left three panels: all compositions of depth \(3\).  Right: distinguishable compositions of depth at most \(k\) at resolution \(10^{-2}\) (blue circles contractive, green squares non-expanding, red triangles expanding), with the references \((k+1)(k+2)/2\) (dash-dotted) and \(2^k\) (dotted).  See \cref{sec:discussion-relu,sec:relu-computed}.}
    \label{fig:relu-example}
\end{figure}

For layers that are only Lipschitz and norm-bounded, the layerwise bound
places the network in the PP case on a bounded state space, in agreement
with the VC-dimension and nonparametric analyses of deep ReLU networks
(\cref{sec:lit-overview}); on an unbounded state space only the diameter
can grow exponentially.  Structure lowers the profile, and
\cref{fig:relu-example} shows this on the smallest ReLU networks: three
classes of two one-hidden-layer ReLU maps of \([0,1]\) into itself that
differ only in their Lipschitz constant, namely the two contractions
\(x/4\) and \((x+3)/4\) (\(\Lambda=1/4\)), two clipped translations by
incommensurable positive steps, which commute (\(\Lambda=1\)), and two
ping-pong maps that expand a subinterval onto \([0,1]\) and reset the rest
to a constant (\(\Lambda=20\)).  The left panels show all eight
compositions of depth \(3\), and the right panel counts the compositions
of depth at most \(k\) that are distinguishable at resolution \(10^{-2}\):
the contractive count saturates after four layers, the commuting count
grows quadratically, and the expanding count grows like \(2^k\) (maps,
evaluation set, and probe points in \cref{sec:relu-computed}).  The same
layer type thus realizes every case of \cref{thm:mechanisms-main}, and the
case is decided by the Lipschitz constant and the algebra of the layers,
not by the width or the parameter count.

\subsection{Chain-of-thought computation}
\label{sec:discussion-cot}

\begin{example}[Programs on a scratchpad]
\label{ex:cot-main}
The state is a scratchpad, an infinite sequence \(x=(x_0,x_1,\dots)\) of
symbols from an alphabet with \(r\) active symbols and a padding symbol,
with the metric \(d(x,y)=\theta^{n(x,y)}\), where \(n(x,y)\) is the first
position at which \(x\) and \(y\) differ and \(\theta\in(0,1)\); a hidden
layer is one reasoning step, and the output layer reads the first symbols
linearly.  For append-only steps the profile saturates and the bias of a
long target program decays like \(\theta^k\), so \(\kopt\asymp\log n\)
(EL; \cref{prop:cot-append}).  For branching steps the uniform entropy
grows like \(k\), but on every sample of size \(n\) the empirical entropy
grows like \(\log k\), so the cost of \(k\) steps is \(\sqrt{\log k/n}\)
(\cref{prop:cot-branch} in \cref{sec:app-cot}).
\end{example}

\begin{figure}[t]
    \centering
    \includegraphics[width=\linewidth]{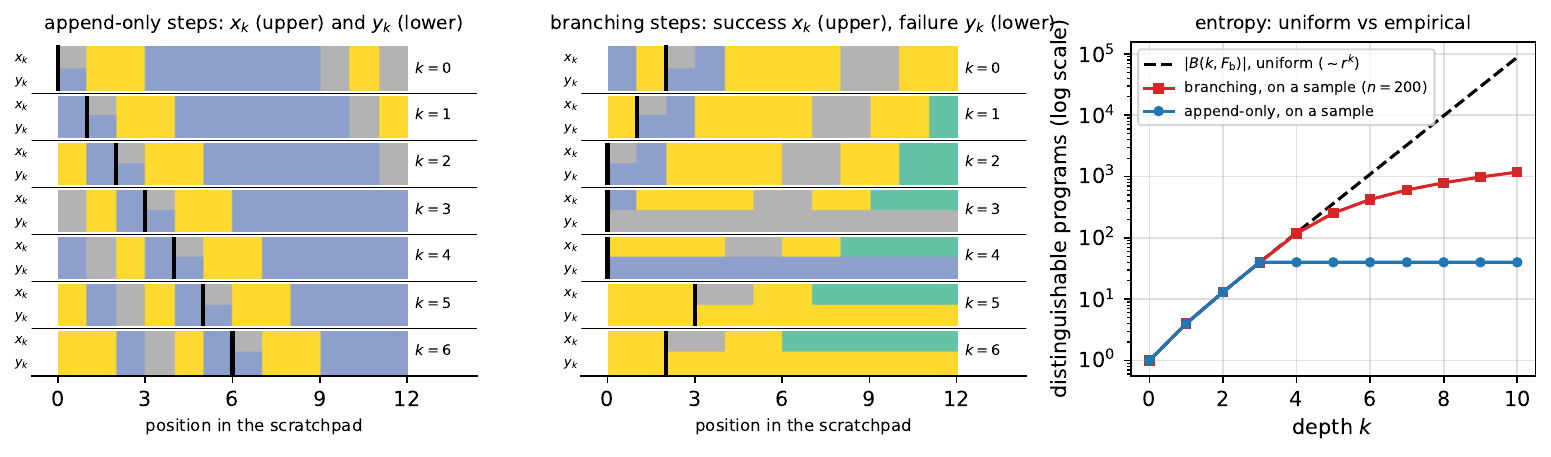}
    \caption{Chain-of-thought computation with \(r=3\) symbols.  Left and middle: scratchpads \(x_k\) (upper row of each pair) and \(y_k\) (lower row) of two inputs after \(k\) steps of one program, under append-only steps (left) and branching steps (middle); one colour per symbol, grey for padding, and a black bar at the first position where the two differ.  Right: distinguishable programs of length at most \(k\) on a sample of \(200\) inputs (red squares branching, blue circles append-only) and the number of programs (dashed).  See \cref{sec:discussion-cot,sec:cot-computed}.}
    \label{fig:cot-example}
\end{figure}

The two step rules differ in what a step does to the prefix.  An
\emph{append-only} step with symbol \(a\) writes \(a\) in front,
\(g_a(x)=(a,x_0,x_1,\dots)\), so each \(g_a\) is a contraction with factor
\(\theta\) and the entropy saturates.  A \emph{branching} step with symbol
\(a\) reads the current symbol: if \(x_0=a\) it consumes it,
\(f_a(x)=(x_1,x_2,\dots)\), and otherwise it resets the scratchpad to the
constant sequence \((a,a,\dots)\).  Distinct programs of \(k\) branching
steps are separated in \(d_\infty\), so their number grows like \(r^k\);
but on any sample of size \(n\), programs of the same length succeed on
disjoint sets of inputs, so at most \(n\) of them are distinguishable
from a constant map, and the empirical entropy grows only like \(\log k\).
\Cref{fig:cot-example} shows both rules with \(r=3\) and \(\theta=1/2\).
In the left and middle panels each pair of rows shows the scratchpads of
two inputs after \(k\) steps of the program \((1,2,3,1,2,2)\), one colour
per symbol, with a black bar at the first position at which they differ:
under append-only steps the bar moves right by one per step, and under
branching steps the lower input fails at its third step and is reset.
The right panel counts, for a sample of \(200\) inputs, the programs of
length at most \(k\) that are distinguishable at resolution \(\theta^3\)
on the sample: for branching steps the count follows \(3^k\) while that is
below \(200\) and then grows only linearly, and for append-only steps it
saturates at \(40\) (details in \cref{sec:cot-computed}).

\section{Conclusion}

We analyzed depth in a model in which the hidden layers are fixed and only
their number varies.  The statistical cost of depth is the metric entropy
of the compositions \(B(k,F)\): it bounds the estimation error with
constants that do not depend on the depth, it is paid whenever the output
layer can see the hidden states, and its growth with \(k\) is decided by
the geometry of composition.  For Lipschitz layers on a bounded state space
it grows at most polynomially, and contraction, equicontinuity, or
nilpotent structure make it constant or logarithmic.  Balanced against the
approximation error, this gives the depth that a sample of size \(n\)
justifies, and it says whether that depth is a threshold beyond which
further layers do not add to the estimation error or an optimum beyond
which they do.  The examples show that one layer type realizes every case
according to its Lipschitz constant, and that for iterative models both
sides of the balance follow from a single step.  Deeper is better than
shallower, in this family, when the target admits fast depth-driven
approximation and the compositions of hidden layers do not create many
distinguishable maps that the output layer and the sample can see.

\subsubsection*{Acknowledgments}
\begin{small}
This work was supported by
JSPS KAKENHI 24K21316, 25H01453,
JST BOOST JPMJBY24E2,
JST CREST JPMJCR2015, JPMJCR25I5.
\end{small}

\bibliographystyle{abbrvnat}
\bibliography{libraryS,extra}

\newpage
\appendix

\section*{Guide to the appendix}
The appendix follows the order of the main text.
\Cref{sec:app-remarks,sec:lit-overview} collect conventions and related
work.  \Cref{sec:proof-bv,sec:proof-hidden,sec:sudakov-proof} contain the
statements and proofs of \cref{sec:bounds}: the bias--variance
decomposition, the hidden--output decomposition with the output-layer
classes that satisfy \cref{ass:sg-increment-main}, and the Sudakov-type
converse.  \Cref{sec:growth-conditions-proof,sec:proof-envelope,sec:proof-profiles}
contain the growth mechanisms of \cref{sec:growth-conditions} with their
examples, the layerwise layerwise bound, and the variance profiles.
\Cref{sec:proof.tradeoff} contains the four regimes of \cref{sec:optdepth}
and their balancing depths.  The remaining appendices are the worked
examples of \cref{sec:discussion}, each written in the same order
(setting, variance, bias, balancing, computed illustration,
implementation, literature): teacher--student models and neural operators
together with the implementation layer (\cref{sec:examples-regime}), deep
ReLU networks (\cref{sec:app-relu}), chain-of-thought computation
(\cref{sec:app-cot}), and unrolled solvers and samplers (\cref{sec:app-ode}).
Every statement is formalized in the Lean~4 repository at
\url{https://shosonoda.github.io/lean-deepgen/} under the label shown in its \LaTeX{} source.

\section{Conventions and basic facts}
\label{sec:app-remarks}

This appendix fixes the conventions used in all proofs: covering and
packing numbers and their elementary properties, the uniform and empirical
metrics on hidden maps, the semigroup generated by a hidden-layer class,
the symbolic state space used in several examples, and two remarks on the
learning problem of \cref{sec:setting}.  Nothing here is specific to a
mechanism or an example; later appendices refer back to this one.

\subsection{Covering and packing numbers}
\label{sec:app-covering}

For a metric space \((\calM,\rho)\),
  write
  \[
  N(A,\rho,\eps) := \min\left\{ |C| \;\middle|\; C \subset \calM \text{ s.t. } A \subset \bigcup_{y \in C} B_\rho(y,\eps)\right\}
  \]
  for the \(\eps\)-covering number of \(A\subset \calM\), and
  \[
  M(A,\rho,\eps) := \max\left\{|S| \mid S\subset A, \rho(x,y)\ge \eps \ \forall x\neq y\in S\right\}
  \] for the \(\eps\)-packing number.
  Thus \(N\) is the external covering number by closed balls (centres
  anywhere in \(\calM\)); it is monotone under inclusion, \(A\subset A'\)
  implies \(N(A,\rho,\eps)\le N(A',\rho,\eps)\).  The internal covering
  number \(N^{\rm int}(A,\rho,\eps)\), whose centres are required to lie in
  \(A\), satisfies \(N(A,\rho,\eps)\le N^{\rm int}(A,\rho,\eps)\le
  N(A,\rho,\eps/2)\) but is \emph{not} monotone under inclusion.  Upper
  bounds below are stated for \(N\); lower bounds are obtained through
  packing numbers, which are monotone under inclusion, and apply to both
  conventions.

The following facts are used throughout.

\begin{lemma}[Packing and covering numbers]
For any subset \(A\) of a metric space and $\eps>0$,
\[
M(A,2\eps) \le N(A,\eps) \le M(A,\eps).
\]
\end{lemma}

\begin{lemma}[Lipschitz images]
Let $(X,d_X), (Y, d_Y)$ be metric spaces. Suppose $\phi:X \to Y$ is $L$-Lipschitz, then for every subset $S \subset X$,
\[
N( \phi(S), d_Y, L \eps ) \le N( S, d_X, \eps ), \quad
M( \phi(S), d_Y, L \eps ) \le M( S, d_X, \eps ).
\]
\end{lemma}

\begin{lemma}[Subadditivity under unions]%
    For any metric space $(M,d)$, subsets $F,G \subset M$, and $\eps>0$,
    we have \[N(F \cup G,\eps) \le N(F,\eps) + N(G,\eps).\]
\end{lemma}
\begin{proof}
    Let $A,B$ be $\eps$-coverings of $F,G$ respectively. Then, $A \cup B$ is an $\eps$-covering of $F \cup G$ because 
    for any $h \in F \cup G$ there exists $c \in A \cup B$ satisfying $d(h,c) \le \eps$.
    Thus, $N( F \cup G, \eps ) \le |A \cup B| \le |A| + |B| = N(F,\eps) + N(G,\eps)$.
\end{proof}

\begin{lemma}[Submultiplicativity under composition]%
    For any bi-Lipschitz metric semigroup $(M,d)$ with $\sup_{f \in M} d(fx,fy) \le \lambda d(x,y)$ (left Lipschitz) and $\sup_{f \in M} d(xf,yf) \le \rho d(x,y)$ (right Lipschitz),
    for any subsets $F,G \subset M$, and $\eps,\delta>0$,
    we have \[N(FG,\eps+\delta) \le N(F,\eps/\rho)N(G,\delta/\lambda).\]
\end{lemma}
\begin{proof}
    Let $A,B$ be $\alpha,\beta$-coverings of $F,G$ respectively. Then, $AB$ is an $(\rho\alpha+\lambda\beta)$-covering of $FG$ because 
    for any $fg \in FG$ there exists $ab\in AB$ satisfying $d(fg,ab) \le d(fg,ag) + d(ag,ab) \le \rho d(f,a) + \lambda d(g,b) = \rho\alpha + \lambda\beta$.
    Thus, $N( FG, \rho\alpha+\lambda\beta ) \le |AB| \le |A||B| = N(F,\alpha)N(G,\beta)$.
    Letting $\alpha=\eps/\rho, \beta=\delta/\lambda$ yields the assertion.
\end{proof}

\begin{lemma}[Right composition is 1-Lipschitz]
$d_\infty(a \circ f, b \circ f) \le d_\infty(a,b)$
\end{lemma}
\begin{proof}
    $d_\infty(a \circ f, b \circ f)
    = \sup_{x \in \xdom} d( a(f(x)), b(f(x)) )
    \le \sup_{y \in \xdom} d( a(y), b(y) )
    = d_\infty( a, b )$
\end{proof}

\begin{lemma}[Left composition is Lipschitz]
$d_\infty(f \circ a, f \circ b) \le \lip(f) d_\infty(a,b)$
\end{lemma}
\begin{proof}
    $d_\infty(f \circ a, f \circ b)
    = \sup_{x \in \xdom} d( f(a(x)), f(b(x)) )
    \le \sup_{x \in \xdom} \lip(f) d( a(x), b(x) )
    = \lip(f) d_\infty( a, b )$
\end{proof}

\begin{lemma}[Packing lower bounds from finite probes] \label{lem:probes-packing}
Let \(P=\{x_1,\dots,x_m\}\subset \xdom\) be finite and define
$d_P\big((y_j),(z_j)\big):=\max_{1\le j\le m} d(y_j,z_j)
\text{ on }\xdom^m$, and
$\eval_P:C(\xdom,\xdom)\to \xdom^m, \eval_P(f)=(f(x_1),\dots,f(x_m))$.
Then \(\eval_P\) is 1-Lipschitz:
\(d_P \big(\eval_P(f),\eval_P(g)\big)\le d_\infty(f,g)\). If
\(\eval_P(S)\subset \xdom^m\) contains \(M\) points that are pairwise
\(\delta\)-separated (in \(d_P\)), then for every
\(\varepsilon<\delta/2\), \[
N(S,d_\infty,\varepsilon) \ge M.
\]
\end{lemma}

\begin{proof}
The 1-Lipschitz claim is immediate: \[
d\big(f(x_j),g(x_j)\big)\le \sup_{x\in \xdom} d\big(f(x),g(x)\big)=d_\infty(f,g).
\] If \(\eval_P(s_1),\dots,\eval_P(s_M)\) are \(\delta\)-separated, then
\(d_\infty(s_i,s_j)\ge d_P(\eval_P(s_i),\eval_P(s_j))\ge\delta\). Any ball of
radius \(\varepsilon<\delta/2\) in the \(d_\infty\)-metric can contain
at most one of the \(s_i\)'s, so at least \(M\) balls are needed.
\end{proof}

\subsection{Uniform and empirical metrics on hidden maps}
\label{sec:app-metrics}

Throughout, \((\xdom,d)\) is a metric space, not necessarily compact, and
\(C(\xdom,\xdom)\) is the set of continuous self-maps of \(\xdom\).

\paragraph{Uniform metric.}
On \(C(\xdom,\xdom)\) we use the uniform metric \[
  d_\infty(f,g) := \sup_{x\in \xdom} d\left(f(x),g(x)\right).
  \]
  When \(\xdom\) is non-compact, this quantity may be infinite for some
  pairs of maps. Whenever a finite ambient metric is needed, we use one
  of the following two devices:
  \begin{itemize}
  \item
    ($d_b$)
    replace \(d\) by its bounded version
    \(d_b(x,y):=\min\{1,d(x,y)\}\);
  \item
    ($C_b$)
    fix a compact set \(K_0\subset \xdom\) and assume all
    maps under discussion (and all words built from them) take values in
    \(K_0\).
  \end{itemize}
  These three settings---$C(\xdom,\xdom)$ with compact $\xdom$ and $d_\infty$, 
  $C(\xdom,\xdom)$ with non-compact $\xdom$ and $d_b$,
  and $C_b(\xdom,\xdom)$ with non-compact $\xdom$ and $d_\infty$---give finite uniform metrics.

\paragraph{Ascoli's theorem.}
We use Ascoli's theorem in the following form \citep[Appendix~A5]{Rudin.FA.new}: a family of continuous self-maps of a compact metric space is totally bounded in \(d_\infty\) if and only if it is equicontinuous.  On a non-compact space the theorem gives only compactness in the compact-open topology, which is why the bounded-metric and bounded-range devices above are needed.

\paragraph{Lipschitz and co-Lipschitz constants.}
For \(f\in C(\xdom,\xdom)\) and
  \(S\subset \xdom\) nonempty, \[
  \lip_S(f):=\sup_{x\ne y\in S}\frac{d(f(x),f(y))}{d(x,y)},\qquad
  \colip_S(f):=\inf_{x\ne y\in S}\frac{d(f(x),f(y))}{d(x,y)}.
  \] We say ``\(f\) is uniformly expanding on \(S\)'' if
  \(\colip_S(f) > 1\).

\subsection{Semigroups and word balls}
\label{sec:app-semigroup}

For \(F\subset C(\xdom,\xdom)\), write
  \(\langle F\rangle\) for the semigroup generated by \(F\) under
  composition. For \(k\in\NN\), let \(B(k,F)\) be the set of maps
  obtainable as a composition of at most \(k\) maps from \(F\).  Since
  \(B(0,F)=\{\id\}\), the identity belongs to
  \(\langle F\rangle=\bigcup_{k\ge0}B(k,F)\), which is therefore a monoid.

\paragraph{Word length.} For
  \(h\in\langle F\rangle\), denote by \(\ell_F(h)\) the least \(\ell\)
  such that \(h\) is a composition of \(\ell\) maps from \(F\). In all
  results below, when we assert that certain auxiliary maps \(h\) belong
  to \(\langle F\rangle\), we also assume that \(\ell_F(h)\) is bounded
  by a constant independent of \(k\).

\paragraph{Homeomorphism and isometry groups.}
$\homeo(\xdom) \subset C(\xdom,\xdom)$ denotes the homeomorphism group of $\xdom$, that is, the set of all \emph{bijective} bi-continuous self-maps with the function composition as group operation. \(\mathrm{Isom}(\xdom) \subset C(\xdom,\xdom)\) denotes the isometry group of \((\xdom,d)\), that is, the set of all \emph{bijective} distance-preserving self-maps. We only use these groups as sources of actions; the metric estimates below are stated explicitly in \(d_\infty\).

\subsection{The full shift and its ultrametric}
\label{sec:app-shift}

On the \emph{one-sided full shift} \(\Sigma_m^+=[m]^\NN\) (with
\([m]=\{1,\dots,m\}\) discrete), fix \(\theta\in(0,1)\) and define the
ultrametric \[
d_\theta(x,y)=
\begin{cases}
0,& x=y,\\
\theta^{n(x,y)-1},& x\neq y,
\end{cases}
\quad\text{where}\quad
n(x,y):=\min\{i\ge1 \mid x_i\neq y_i\}.
\]
This is an ultrametric (satisfies \emph{strong triangle inequality}
\(d(x,z)\le\max\{d(x,y),d(y,z)\}\)), it induces the product topology, and
it makes \(\Sigma_m^+\) a compact, totally disconnected, perfect
(Cantor-type) space. Distances take only the values
\({1,\theta,\theta^2,\ldots}\); in particular, sequences that differ in
the first symbol are at distance \(1\).

For a finite word \(u=u_1\cdots u_\ell\), the \emph{(prefix) cylinder}
of depth \(\ell\) is given by
\[
\ssq{u}:=\{w\in\Sigma_m^+ \mid w_1=u_1,\dots,w_\ell=u_\ell\},
\] a clopen set; cylinders form a basis and the depth-\(\ell\) cylinders
partition \(\Sigma_m^+\) into \(m^\ell\) pieces. In the metric
\(d_\theta\), \(\diam(\ssq{u})=\theta^\ell\), and any two
distinct depth-\(\ell\) cylinders are separated by at least
\(\theta^{\ell-1}\). Moreover, balls coincide with cylinders: for
\(w\in\Sigma_m^+\) and \(\ell\ge0\),
\(\overline{B}_{d_\theta}(w,\theta^\ell)=\ssq{w_1\cdots w_\ell}\).
The left shift \(\sigma(w)_i=w_{i+1}\) satisfies
\(\sigma(\ssq{a u})=\ssq{u}\) and \(\sigma^{-1}(\ssq{u})=\bigsqcup_{a\in[m]}\ssq{a u}\).

\subsection{Remarks on the learning problem}
\label{sec:app-setting-remarks}

\begin{remark}[Approximation term versus function approximation]
\label{rem:approx-vs-risk-app}
\(\errmodel(k)\) is a gap between oracle risks, not the distance between a
target function and its best approximant.  The two notions are related
through the Lipschitz loss: since
\(\errmodel(k)=\sup_{c\in\Con}\bigl(\inf_{f\in\calH_k}L[f]-L[c]\bigr)\)
and \(L[f]-L[c]\le\beta_\ell\|f-c\|_\infty\), \eqref{eq:approx-transfer}
holds.  Hence uniform approximation rates from approximation theory
transfer to the bias term with the constant \(\beta_\ell\).  We call
\(\errmodel(k)\) the approximation term when it is nonnegative, as is the
case when \(\Con\) is a benchmark at least as good as the depth-\(k\)
hypothesis class in oracle risk; in complete generality it is an oracle gap
relative to \(\Con\), and if a nonnegative approximation error is desired
one may replace it by its positive part, at the cost of a weaker but still
valid bound.  The risk-based definition is used because it allows the same
hypothesis class to be compared against different benchmarks and requires
no norm structure on \(\Con\).
\end{remark}

\begin{remark}[Loss constants]
\label{rem:loss-app}
For a fixed loss, \(b\) and \(\beta_\ell\) are fixed and do not depend on the
depth; examples are the clipped squared loss, hinge and ramp losses, and the
logistic loss on a bounded logit range.  Cross-entropy expressed in class
probabilities is unbounded and non-Lipschitz near the boundary of the
simplex; expressed in logits its derivative is bounded but the loss itself is
unbounded unless the logits are restricted to a bounded range.  Such losses
therefore enter \cref{thm:bv} only after probability clipping or a logit range
restriction.  If the clipping threshold \(\tau_{k,n}\) is allowed to vary
with \(k\) or \(n\), then \(b\asymp\log(1/\tau_{k,n})\) and
\(\beta_\ell\asymp 1/\tau_{k,n}\) must be carried through the rates in
\cref{sec:optdepth}; with a fixed threshold the exponents there are unchanged.
\end{remark}

\section{Literature Overview}\label{sec:lit-overview}

One motivation of this paper is to organize the generalization theory of
deep learning that has been developed, for the most part, on the fully
connected ReLU network.  That theory contains several depth dependences,
polynomial, logarithmic, and depth-independent, obtained by different
techniques and under different assumptions; we review them below and then
state, at the end of this appendix, how each of them appears as a case of
the present framework.  \Cref{sec:app-relu} works out the ReLU network as
an example of the framework in its own right.

\textbf{Depth separation and expressivity.} A classical line of work
shows that modest increases in depth can yield exponential
representational advantages. \citet{Eldan2016} proved a
three-vs-two-layer separation for a simple radial function, requiring
exponential width for any depth-2 approximant, while \citet{Telgarsky2016}
established families exhibiting exponential gaps between networks of
depth \(O(k^3)\) and \(O(k)\) with semi-algebraic gates (including ReLU)
and provided constructive hard instances; subsequent work extended
separations beyond radial constructions. These results clarify when
\emph{expressivity} favors depth, but do not by themselves pin down
estimation behavior. %

\textbf{Generalization via capacity control.} Combinatorial analyses
give nearly tight VC/pseudodimension bounds for piecewise-linear
networks, scaling roughly linearly in depth for fixed width/weights,
providing a baseline picture of depth in classical uniform-convergence
frameworks \citep{Bartlett2019}. Norm- and margin-based approaches bound estimation error
through products of layer norms (path/spectral) \citep{Neyshabur2015,Bartlett2017}, sometimes yielding
size-independent or gently depth-dependent bounds under additional
structure (e.g., margin normalization).
These strands highlight multiple possible depth dependencies---linear, polynomial, or even
milder---depending on how complexity is measured.

\textbf{Rademacher/covering and size-independent bounds.} A
complementary thread controls depth via data-dependent complexities
(Rademacher, covering). 
\citet{Golowich2018,Golowich2020} obtained bounds
that (under norm constraints) improve the depth dependence and can be
independent of width and depth in certain regimes; later refinements
further reduced explicit depth factors. These works show how the
estimation side may be decoupled from naively counted parameters
and instead tied to geometric quantities of the hypothesis class.

\textbf{Compression and PAC-Bayes.} A productive viewpoint explains
generalization via compressibility: if a trained network admits a
succinct reparametrization, one can transfer that compression into
generalization guarantees. 
\citet{Arora2018compression} formalized this link and
demonstrated strong bounds in practice; follow-ups convert compression
bounds to the original (non-compressed) networks and sharpen them via
PAC-Bayes with subspace quantization, yielding state-of-the-art
nonvacuous estimates \citep{Suzuki2020compression,Lotfi2022compression}.
While depth typically enters these bounds through
compressibility or margin quantities, the methodology is agnostic to
architecture details.

\textbf{Nonparametric regression with deep networks.} Another large body
of work analyzes approximation--estimation trade-offs of ReLU networks
on smoothness classes. \citet{Schmidt-Hieber2020} showed near-optimal rates in
nonparametric regression, with depth playing an essential role; \citet{suzuki2018besov-adaptivity}
established optimal adaptivity on (mixed) Besov spaces and improvements
over linear/kernel baselines; \citet{Nakada2020imaizumi} tied generalization to
intrinsic (Minkowski) dimension; and \citet{Imaizumi2019,Imaizumi2022fukumizu} identified
regimes with singularities where DNNs are minimax-superior to
traditional estimators. Recent approximation results (e.g., optimal
Sobolev/Besov rates) further sharpen the expressivity side \citep{Yarotsky2017,Yarotsky2018,Yarotsky2020,Siegel2022,Siegel2023,Yang2024deeperwider}.
These analyses, however, are typically architecture-specific (ReLU
feedforward) and hinge on smoothness assumptions.

\textbf{Iterative/hierarchical models and continuous depth.} Many modern
systems are naturally modeled as \emph{compositions} or
\emph{flows}---precisely the setting of our state-transition
abstraction. Neural ODEs \citep{Chen2018a} as well as Neural Operators \citep{Kovachki2021NOreview} treat depth as continuous time evolution;
diffusion/score-based models \citep{shen2025diffusion-survey-tmlr}
implement long iterative refinement; and
chain-of-thought \citep{Wei2022cot} prompting in LLMs explicitly unfolds multi-step
reasoning. These families motivate studying depth-dependent
generalization at the level of abstract state transitions rather than
fixed architectures.

\textbf{Deep ReLU networks in the coordinates of this paper.}
For H\"older, Sobolev, and Besov targets on \([0,1]^d\), the best
\(m\)-parameter approximation has the Jackson rate \(O(m^{-s/d})\) with
matching lower bounds \citep{DeVore1989}; deep ReLU networks reach and,
by bit-extraction, slightly surpass this rate, but the decay in the depth
remains polynomial \citep{Yarotsky2017,Yarotsky2018,Yarotsky2020,Siegel2023,Yang2024deeperwider}.
On the estimation side, VC-dimension and compression bounds grow
polynomially in the depth \citep{Bartlett2019,Arora2018compression,Suzuki2020compression,Lotfi2022compression}.
Hence deep ReLU networks on smoothness classes are in the PP case of
\cref{sec:optdepth}, and the minimax analysis of
\citet{suzuki2018besov-adaptivity} is an instance.  For the hierarchical
compositions of H\"older functions of \citet{Schmidt-Hieber2020}, the
covering numbers of deep ReLU classes grow only logarithmically in the
depth while the approximation error has an exponentially decaying
compositional part, which is close to the PL case.  In both cases the
framework classifies known rates rather than deriving new ones; its
contribution is the quantity, the entropy of \(B(k,F)\), that explains why
the two cases differ.

\textbf{How the ReLU results appear in the framework.}
Read in the coordinates of this paper, the depth dependences above are
instances of the mechanisms of \cref{sec:growth-conditions}, with the
ReLU layer as the hidden-layer class (\cref{sec:app-relu}).  The
polynomial depth dependence of VC-dimension and covering-number bounds
\citep{Bartlett2019,Schmidt-Hieber2020} is the layerwise bound of
\cref{prop:envelope} for a parametric layer class, which reproduces its
order up to a factor \(\sqrt k\) and logarithms.  The product of layer
norms in norm-based bounds \citep{Neyshabur2015,Bartlett2017} is the
diameter of the reachable states on an unbounded state space
(\cref{lem:reachable-radius}); on a bounded state space it disappears.
The depth-independent bounds obtained under norm constraints
\citep{Golowich2018} correspond to non-expanding layers on a bounded
state space, which is the saturation mechanism P1
(\cref{prop:relu-regimes}(ii)).  The exponential-in-depth worst cases of
uniform-convergence bounds correspond to expanding layers with separated
compositions, mechanism E2, and are realized by expand-and-reset ReLU
layers (\cref{prop:relu-regimes}(iii)).  On the approximation side, the
minimax analyses on smoothness classes and hierarchical compositions
\citep{suzuki2018besov-adaptivity,Schmidt-Hieber2020} supply the bias
laws of the PP and PL regimes.  The framework does not reproduce the constants of these proofs; it
reproduces their depth dependence from one quantity, the entropy of
\(B(k,F)\), and the same computation applies to architectures that the
ReLU theory does not cover.

\textbf{This Study.}
Relative to these threads, this study is
\emph{implementation-agnostic}: instead of parameterizing a specific architecture, we analyze \emph{state-hidden-layer semigroups on metric spaces}, derive depth dependence of the \emph{variance} via
covering/Rademacher complexity of word balls, give conditions for
\emph{polynomial/logarithmic} growth, and couple them with
\emph{exponential/polynomial} bias decay to compare depth scalings across
four typical regimes. This yields a unified lens for when and
why depth can be statistically preferable to shallower members of the same
transition model---particularly in iterative/hierarchical settings suggested
above.  The distinction from the architecture-specific threads is one of
scope rather than of rates: the sharp depth-dependent results above are
proved for one implementation, most often a fully connected ReLU network on a
Sobolev, H\"older, or Besov class, and a change in the layer structure, the
state space, or the parameter sharing generally requires a new proof.  Here
the one-layer class \(F\) carries all implementation-specific quantities,
including width through its covering numbers, while composition depth is
isolated in the word ball \(B(k,F)\); existing rates for particular networks
then become examples that can be inserted into the framework
(\cref{sec:examples-regime}).

\section{The bias--variance decomposition}
\label{sec:proof-bv}

This appendix states the bias--variance decomposition of \cref{sec:bounds}
precisely, proves a general version of it for an arbitrary abstract class,
and records the specialization to \(\calH_k\).  The decomposition is the
first step of the reduction: it isolates the statistical term
\(\erad_S(\calH_k)\) that the rest of the paper analyzes.

\subsection{Statement}

\begin{theorem}[Implementation-free bias--variance decomposition]
\label{thm:bv}
Fix $k\ge 0$ and \(\eta\ge0\).  Assume that $\calH_k\subset L^\infty(\xdom)$
is separable in the uniform norm and pointwise bounded, and that the loss
\(\ell:\RR\times\calY\to[0,b]\), \(b>0\), is measurable and
\(\beta_\ell\)-Lipschitz in its first argument.  Let \(\fhat\in\calH_k\) be
an \(\eta\)-empirical minimizer and \(\hhat:=\embed(\fhat)\).  Then for every
$\delta\in(0,1)$, with probability at least $1-\delta$ over $\calD\sim P^n$,
\begin{equation}
L[\hat h]-\inf_{c\in\Con}L[c]
\le \beta_\ell \errimp(k)+\errmodel(k)+\eta+4\beta_\ell \erad_S(\calH_k)+6b\sqrt{\frac{2\log(4/\delta)}{n}} .
\label{eq:bv-excess}
\end{equation}
The same bound holds with $\erad_S(\calH_k)$ replaced by $\rad_n(\calH_k)$;
a generalization-gap version, valid for every \(f\in\calH_k\), is given in
\cref{sec:proof-bv}.
\end{theorem}

\subsection{A general version and its proof}

We prove a slightly more general version of \cref{thm:bv}.  In this appendix,
the abstract class is denoted by \(\Habs\) rather than \(\calH_k\), because the
argument does not use the state-transition representation \(H\circ B(k,F)\).
The specialization \(\Habs=\calH_k\) gives the theorem in the main text.

\begin{theorem}[General bias--variance decomposition]
\label{thm:bv-general}
Let \(\Habs\) be a class of measurable functions \(\xdom\to\RR\) that is
separable in the uniform norm and pointwise bounded
(\(\sup_{f\in\Habs}|f(x)|<\infty\) for every \(x\)), let \(\Con\) be a
nonempty class of measurable benchmark functions \(\xdom\to\RR\), and let
\(\ell:\RR\times\calY\to[0,b]\), \(b>0\), be measurable and
\(\beta_\ell\)-Lipschitz in its first argument.  Define
\[
L[f]:=\EE[\ell(f(X),Y)],
\qquad
\Lhat[f]:=\frac1n\sum_{i=1}^n \ell(f(X_i),Y_i),
\]
and set
\[
L_{\Con}:=\inf_{c\in\Con}L[c],
\qquad
L_{\Habs}:=\inf_{f\in\Habs}L[f],
\qquad
\errmodel:=L_{\Habs}-L_{\Con}.
\]
Let \(\embed:\Habs\to\Himp\) be a implementation map into an implementation
class and let \(d_T\) be a pseudo-metric on \(\Habs\cup\Himp\).  Assume that
\[
\errimp:=\sup_{f\in\Habs}d_T(\embed(f),f)<\infty
\]
and that, for every \(f\in\Habs\),
\begin{align*}
|L[\embed(f)]-L[f]|&\le \beta_L d_T(\embed(f),f),\\
|\Lhat[\embed(f)]-\Lhat[f]|&\le \beta_{\hat L}d_T(\embed(f),f).
\end{align*}
For \(\eta\ge0\), let \(\fhat\in\Habs\) be an \(\eta\)-empirical minimizer:
\[
\Lhat[\fhat]\le \inf_{f\in\Habs}\Lhat[f]+\eta,
\]
and output \(\hhat:=\embed(\fhat)\).  Then, for every \(\delta\in(0,1)\),
with probability at least \(1-\delta\) over \(\calD\sim P^n\),
\begin{align}
L[\hhat]-L_{\Con}
&\le
\beta_L\errimp+\errmodel+\eta
+4\beta_\ell\erad_S(\Habs)
+C b\sqrt{\frac{\log(1/\delta)}{n}},
\label{eq:bv-general-excess}
\\
L[\hhat]-\Lhat[\hhat]
&\le
(\beta_L+\beta_{\hat L})\errimp
+2\beta_\ell\erad_S(\Habs)
+C b\sqrt{\frac{\log(1/\delta)}{n}},
\label{eq:bv-general-gap}
\end{align}
where \(C\) is a universal constant; explicitly, one may take the deviation
term to be \(6b\sqrt{2\log(4/\delta)/n}\) in \eqref{eq:bv-general-excess}
and \(3b\sqrt{2\log(4/\delta)/n}\) in \eqref{eq:bv-general-gap}.  On the
same event, \eqref{eq:bv-general-gap} holds for \(\hhat=\embed(f)\) with
every \(f\in\Habs\), not only for \(\eta\)-empirical minimizers.  The same
bounds hold with \(\erad_S(\Habs)\) replaced by \(\rad_n(\Habs)\).
\end{theorem}

\begin{proof}
No topological structure is needed on \(\Con\).  Since the loss is bounded
and nonnegative, \(L[c]\in[0,b]\) for every \(c\in\Con\), so
\(L_{\Con}=\inf_{c\in\Con}L[c]\) is a well-defined real number whenever
\(\Con\) is nonempty.

The only probabilistic ingredient is the standard uniform deviation bound for
bounded Lipschitz loss classes.  Since \(\Habs\) is separable in the uniform
norm and pointwise bounded, the relevant suprema are finite and measurable,
and symmetrization, bounded differences, and the contraction inequality imply
that, with probability at least \(1-\delta\),
\begin{equation}
\Delta_{\Habs}:=
\sup_{f\in\Habs}|L[f]-\Lhat[f]|
\le
2\beta_\ell\erad_S(\Habs)
+C b\sqrt{\frac{\log(1/\delta)}{n}}.
\label{eq:uniform-dev-abstract}
\end{equation}
Explicitly, applying the one-sided symmetrization and bounded-differences
argument to \(\ell\circ\Habs\) and to \(-\ell\circ\Habs\), each with failure
probability \(\delta/2\), and the contraction inequality without absolute
values, gives \eqref{eq:uniform-dev-abstract} with
\(Cb\sqrt{\log(1/\delta)/n}\) replaced by \(3b\sqrt{2\log(4/\delta)/n}\).
The population Rademacher version follows from the same argument after taking
expectation over \(S\).

On the event \eqref{eq:uniform-dev-abstract}, the implementation error is
controlled directly by the assumptions:
\[
L[\hhat]-L[\fhat]\le \beta_L\errimp,
\qquad
\Lhat[\fhat]-\Lhat[\hhat]\le \beta_{\hat L}\errimp.
\]
For the excess risk, no compactness or exact population minimizer is needed.
Using only the infimum \(L_{\Habs}\) and the \(\eta\)-minimality of \(\fhat\),
\begin{align*}
L[\fhat]-L_{\Habs}
&\le
L[\fhat]-\Lhat[\fhat]
+\Lhat[\fhat]-\inf_{f\in\Habs}\Lhat[f]
+\inf_{f\in\Habs}\Lhat[f]-L_{\Habs}
\\
&\le
\Delta_{\Habs}+\eta+\Delta_{\Habs}
=2\Delta_{\Habs}+\eta .
\end{align*}
Therefore,
\[
L[\hhat]-L_{\Con}
\le
\beta_L\errimp
+\bigl(L[\fhat]-L_{\Habs}\bigr)
+\errmodel
\le
\beta_L\errimp+\errmodel+2\Delta_{\Habs}+\eta.
\]
Substituting \eqref{eq:uniform-dev-abstract} gives
\eqref{eq:bv-general-excess}.

The generalization gap bound does not require empirical optimality and
holds for \(\hhat=\embed(\fhat)\) with an arbitrary \(\fhat\in\Habs\):
\begin{align*}
L[\hhat]-\Lhat[\hhat]
&=
\bigl(L[\hhat]-L[\fhat]\bigr)
+\bigl(L[\fhat]-\Lhat[\fhat]\bigr)
+\bigl(\Lhat[\fhat]-\Lhat[\hhat]\bigr)
\\
&\le
(\beta_L+\beta_{\hat L})\errimp+\Delta_{\Habs}.
\end{align*}
Again substituting \eqref{eq:uniform-dev-abstract} proves
\eqref{eq:bv-general-gap}.
\end{proof}

\paragraph{Remarks.}
(1) Specialization to \cref{thm:bv}.
Take \(\Habs=\calH_k\), \(d_T(f,g)=\|f-g\|_\infty\),
\(\errimp=\errimp(k)\), and
\(\errmodel=\errmodel(k)\).  The Lipschitz condition on the loss gives
\(\beta_L=\beta_{\hat L}=\beta_\ell\).  If the empirical minimizer in the
main text exists, set \(\eta=0\).  Without exact attainment, the same proof
applies to any \(\eta\)-empirical minimizer and adds only the displayed
\(\eta\) term to the excess-risk bound.

\section{The hidden--output decomposition}
\label{sec:proof-hidden}

This appendix proves \cref{thm:hidden-decomp}(ii), the second step of the
reduction, and shows that the common output-layer classes satisfy
\cref{ass:sg-increment-main} with constants that do not depend on the
depth.  The theorem is restated in local notation, and the proof applies
Dudley's entropy integral to the anchored process \(Z_f-Z_{f_0}\); the
increment condition \eqref{eq:sg-increment} below is exactly
\cref{ass:sg-increment-main}.

\subsection{Statement and proof}

\begin{theorem}[\cref{thm:hidden-decomp}(ii), restated]
\label{thm:mixed-sg}
Let $(X,d)$ be a metric space, let $S=(X_1,\dots,X_n)\subset X$ be fixed, and define
\[
d_S(f,g):=\Bigl(\frac1n\sum_{i=1}^n d\bigl(f(X_i),g(X_i)\bigr)^2\Bigr)^{1/2}
\qquad (f,g:X\to X).
\]
Let $F\subset C(X,X)$ be totally bounded in $d_S$, and fix an anchor $f_0\in F$.
Write
\[
\diam_S(F):=\sup_{f,g\in F}d_S(f,g).
\]
Let $H\subset \mathbb{R}^X$ be a class of real-valued functions.  For each $f\in F$ define
\[
Z_f(\sigma):=\sup_{h\in H}\frac1n\sum_{i=1}^n \sigma_i  h \left(f(X_i)\right),
\]
where $\sigma_1,\dots,\sigma_n$ are i.i.d. Rademacher variables, and assume
that each $Z_f(\sigma)$ is finite.

Assume that there exist constants $L>0$ and $A_H>0$ such that, for all $f,g\in F$ and all $t>0$,
\begin{align}
\label{eq:sg-increment}
\PP_\sigma \left( |Z_f-Z_g|>t \right)
\le
2\exp \left(
-\frac{n t^2}{2A_H^2L^2  d_S(f,g)^2}
\right).
\end{align}
Then
\[
\erad_S(H\circ F)
\le
\erad_S(H\circ f_0)
+
\frac{12A_HL}{\sqrt n}
\int_0^{\diam_S(F)}
\sqrt{\log N(F,d_S,\varepsilon)} d\varepsilon,
\]
where $\erad_S(H\circ f_0)=\EE_\sigma Z_{f_0}$ and the right-hand side is
read as $+\infty$ (so that the bound is trivial) if the entropy integrand is
not integrable on $[0,\diam_S(F)]$.  If $\mathrm{id}_X\in F$, the choice
$f_0=\mathrm{id}_X$ gives the first term $\erad_S(H)$.
In particular, for $F=B(k,\mathcal F)$ one obtains the same bound, with first
term $\erad_S(H)$, for the depth-$k$ class $H_k:=H\circ B(k,\mathcal F)$.
\end{theorem}

\begin{proof}
Anchor the process at $f_0$ and write
\[
Y_f:=Z_f-Z_{f_0}, \qquad f\in F.
\]
Then $Y_{f_0}=0$, and by \eqref{eq:sg-increment},
\[
\PP_\sigma \left(|Y_f-Y_g|>t\right)
=
\PP_\sigma \left(|Z_f-Z_g|>t\right)
\le
2\exp \left(
-\frac{n t^2}{2A_H^2L^2  d_S(f,g)^2}
\right).
\]
Thus $(Y_f)_{f\in F}$ is a process with sub-Gaussian increments with respect to the metric
\[
\rho(f,g):=\frac{A_HL}{\sqrt n}  d_S(f,g).
\]
By Dudley's entropy integral bound for processes with sub-Gaussian
increments (this is the only probabilistic ingredient; it applies to any
process on a finite probability space with sub-Gaussian increments, anchored
at an arbitrary index point, with the constant $12$, and the Rademacher
structure of $Z_f$ enters only through \eqref{eq:sg-increment}),
\[
\EE_\sigma\Bigl[\sup_{f\in F} Y_f\Bigr]
\le
12 \int_0^{\Delta_\rho(F)}
\sqrt{\log N(F,\rho,u)} du,
\]
where
\[
\Delta_\rho(F)
=
\sup_{f\in F}\rho(f,f_0)
\le
\frac{A_HL}{\sqrt n}\diam_S(F).
\]
Changing variables $u=\frac{A_HL}{\sqrt n}\varepsilon$ and enlarging the upper limit gives
\[
\EE_\sigma\Bigl[\sup_{f\in F} Y_f\Bigr]
\le
\frac{12A_HL}{\sqrt n}
\int_0^{\diam_S(F)}
\sqrt{\log N(F,d_S,\varepsilon)} d\varepsilon.
\]
Finally,
\[
\erad_S(H\circ F)
=
\EE_\sigma\Bigl[\sup_{f\in F} Z_f\Bigr]
=
\EE_\sigma\Bigl[Z_{f_0}+\sup_{f\in F}(Z_f-Z_{f_0})\Bigr]
=
\erad_S(H\circ f_0)+\EE_\sigma\Bigl[\sup_{f\in F}Y_f\Bigr],
\]
and the claim follows (if the entropy integrand is not integrable there is
nothing to prove).
\end{proof}

\subsection{Output-layer classes satisfying \cref{ass:sg-increment-main}}
\label{sec:readout-examples}

Assumption~\ref{ass:sg-increment-main} is used only for the upper-bound
decomposition: it prevents the supremum over output layers from amplifying small
hidden-state perturbations beyond the \(d_S(f,g)/\sqrt n\) scale.  It is not a
output-separation assumption; the latter appears only in the Sudakov-type
lower bound.  The condition is satisfied by standard output-layer mechanisms once
the last scalarization is bounded and Lipschitz.

\begin{proposition}%
\label{prop:hilbert-sg}
Let $\calH$ be a real inner product space (completeness is not needed), let $\Phi:X\to\calH$ be $L$-Lipschitz, and define
\[
H_\Phi:=\Bigl\{ x\mapsto \langle w,\Phi(x)\rangle_{\calH}: \|w\|_{\calH}\le 1 \Bigr\}.
\]
Then the assumption \eqref{eq:sg-increment} holds with $H=H_\Phi$ and $A_H=1$.
Consequently,
\[
\erad_S(H_\Phi\circ F)
\le
\erad_S(H_\Phi)
+
\frac{12L}{\sqrt n}
\int_0^{\diam_S(F)}
\sqrt{\log N(F,d_S,\varepsilon)} d\varepsilon.
\]
\end{proposition}

\begin{proof}
For $H_\Phi$, the supremum over the unit ball of $\calH$ can be evaluated explicitly:
\[
Z_f
=
\sup_{\|w\|_{\calH}\le 1}
\frac1n\sum_{i=1}^n \sigma_i \langle w,\Phi(f(X_i))\rangle_{\calH}
=
\Bigl\|
\frac1n\sum_{i=1}^n \sigma_i \Phi(f(X_i))
\Bigr\|_{\calH}.
\]
Hence, for any $f,g\in F$,
\[
|Z_f-Z_g|
\le
\Bigl\|
\frac1n\sum_{i=1}^n \sigma_i\bigl(\Phi(f(X_i))-\Phi(g(X_i))\bigr)
\Bigr\|_{\calH}.
\]
Set
\[
v_i:=\Phi(f(X_i))-\Phi(g(X_i))\in\calH.
\]
The Rademacher tail bound in inner product spaces \citep{Pinelis1994}, whose constant $2$ is optimal, gives, for every $t>0$,
\[
\PP_\sigma \left(
\Bigl\|
\frac1n\sum_{i=1}^n \sigma_i v_i
\Bigr\|_{\calH}
>t
\right)
\le
2\exp \left(
-\frac{n^2 t^2}{2\sum_{i=1}^n \|v_i\|_{\calH}^2}
\right).
\]
Because $\Phi$ is $L$-Lipschitz,
\[
\sum_{i=1}^n \|v_i\|_{\calH}^2
\le
L^2 \sum_{i=1}^n d\bigl(f(X_i),g(X_i)\bigr)^2
=
nL^2 d_S(f,g)^2.
\]
Combining the last two displays yields
\[
\PP_\sigma \left(|Z_f-Z_g|>t\right)
\le
2\exp \left(
-\frac{n t^2}{2L^2 d_S(f,g)^2}
\right),
\]
which is exactly \eqref{eq:sg-increment} with $A_H=1$.
The final bound is now an immediate application of Theorem~\ref{thm:mixed-sg}.
\end{proof}

In neural-network terminology, an MLP with a linear output layer fits
\Cref{prop:hilbert-sg} with the hidden layers placed in the state transition
class \(B(k,F)\) and a fixed feature map \(\Phi\) (for instance the identity on
\(\RR^m\), or a fixed embedding of the terminal state) followed by a
norm-bounded linear functional as the output layer.  With this convention
\(L=\lip(\Phi)\) and \(A_H=1\) do not depend on the depth, which is the
depth-uniform form of Assumption~\ref{ass:sg-increment-main} required in
\cref{sec:bounds}.  One may alternatively absorb some lower layers
into the feature map, but then \(\Phi=\Phi_k\) depends on the depth and
\(L\) must be replaced by \(\lip(\Phi_k)\), which for a \(k\)-layer network
may grow exponentially in \(k\); the bound remains valid but its depth
dependence is then no longer carried by the word-ball entropy alone.

\begin{proposition}[Finite Lipschitz scalar output layers]
\label{prop:finite-lipschitz-sg}
Let \(H=\{h_1,\dots,h_m\}\) be a finite class of real-valued functions on
\(X\).  Suppose every \(h_j\) is \(L\)-Lipschitz with respect to \(d\).  Then
the assumption \eqref{eq:sg-increment} holds with
\[
A_H=\left(1+\frac{\log m}{\log 2}\right)^{1/2}.
\]
\end{proposition}

\begin{proof}
For \(j=1,\dots,m\), define
\[
S_j(f,g):=\frac1n\sum_{i=1}^n \sigma_i
\bigl(h_j(f(X_i))-h_j(g(X_i))\bigr).
\]
Since the difference of two suprema is bounded by the supremum of the
differences,
\[
|Z_f-Z_g|\le \max_{1\le j\le m}|S_j(f,g)|.
\]
For each fixed \(j\), Hoeffding's inequality and the Lipschitz property give
\[
\PP_\sigma(|S_j(f,g)|>t)
\le
2\exp\left(-\frac{n t^2}{2L^2d_S(f,g)^2}\right).
\]
Writing \(x:=n t^2/(2L^2d_S(f,g)^2)\), the union bound yields
\[
\PP_\sigma(|Z_f-Z_g|>t)\le \min\{1,2m e^{-x}\}.
\]
With \(A_H^2=1+\log m/\log 2\), the right-hand side is bounded by
\(2\exp(-x/A_H^2)\) for all real \(x\) (so no restriction on \(t\), \(L\),
\(d_S(f,g)\), or \(m\ge1\) is needed), which is \eqref{eq:sg-increment}.
\end{proof}

Softmax output layers can be handled by including the logits/softmax map in the
state transition and taking the final scalar quantity used for prediction or
loss as the output layer.  Coordinate probabilities are linear output layers on the
simplex, while margins and clipped ramp-type losses are bounded Lipschitz
finite-coordinate scalarizations, so \Cref{prop:finite-lipschitz-sg} applies.
Unconstrained post-softmax output-layer classes and unbounded cross-entropy near
the boundary of the simplex require additional norm, range, or clipping
assumptions.

\paragraph{Remarks.}
(1) \Cref{thm:hidden-decomp} keeps the output-layer complexity and the hidden entropy separate; a cruder alternative, which chains over the whole class \(H\circ F\) with the sum of the two entropies and needs no increment condition, is not used in this paper.
(2) The increment condition is used only for the upper bound; the
separation condition of \cref{sec:sudakov-proof} is a different and
stronger requirement on the output layer.

\section{The Sudakov-type converse}
\label{sec:sudakov-proof}

This appendix proves the conditional lower bound of \cref{sec:bounds}, the
third step of the reduction, and describes the output-layer classes for
which its separation assumption holds.  The assumption asks that, for each
hidden map, some output layer realizes the empirical geometry of
\(B(k,F)\) inside the real-valued class; under it, the packing numbers of
\(B(k,F)\) bound the Rademacher complexity from below, and the depth
dependence of \cref{thm:hidden-decomp} is matched.  The empirical
Rademacher complexity is assumed finite throughout, that is, the averages
\(n^{-1}\sum_i\sigma_ig(X_i)\), \(g\in\calH_k\), are bounded above for every
sign pattern.

\subsection{Statement and proof}

\begin{assumption}[Output-layer realization of hidden geometry]
\label{ass:readout-realization-main}
There exist constants $\kappa,R_{\rm out}>0$ and, for each $g\in B_k$, an
output layer $h_g\in H$ such that $\Psi_k(g):=h_g\circ g$ satisfies
\(\|\Psi_k(g)-\Psi_k(g')\|_S \ge \kappa\, d_S(g,g')\) and
\(\|\Psi_k(g)\|_{S,\infty} \le R_{\rm out}\) for all \(g,g'\in B_k\).
\end{assumption}

\begin{theorem}[Conditional Sudakov-type lower bound]
\label{thm:sudakov-type}
Under Assumption~\ref{ass:readout-realization-main}, if
\(\erad_S(\calH_k)<\infty\), then for a universal constant $c>0$
\begin{equation}
\erad_S(\calH_k)
\ge
c\sup_{\eps>0}
\min\left\{
\kappa\eps\sqrt{\frac{\log M(B_k,d_S,2\eps)}{n}},
\frac{\kappa^2\eps^2}{R_{\rm out}}
\right\},
\label{eq:sudakov-main}
\end{equation}
where the first entry is \(+\infty\) if the packing number is.  The
word-ball structure is not used: the statement holds for any
\(B\subset C(\xdom,\xdom)\).
\end{theorem}

\begin{theorem}[\cref{thm:sudakov-type}, restated]
\label{thm:sudakov}
Fix \(k\in\NN\), a sample \(S=(X_1,\dots,X_n)\subset X\), and write
\[
B_k:=B(k,F),
\qquad
H_k:=H\circ B_k .
\]
(The word-ball structure is not used below: the statement and proof apply
verbatim to an arbitrary subset \(B\subset C(X,X)\) in place of \(B_k\), with
\(H\circ B\) in place of \(H_k\).)
For real-valued functions on the sample, write
\[
\|u\|_{S,\infty}:=\max_{i\in[n]} |u(X_i)|.
\]
Assume that there exist constants \(\kappa,R_{\rm out}>0\) and, for each \(f\in B_k\), an output layer \(h_f\in H\) such that
\begin{align}
\|h_f\circ f-h_g\circ g\|_S & \ge \kappa  d_S(f,g)
\qquad (f,g\in B_k), \label{eq:lower-lip-readout}\\
\|h_f\circ f\|_{S,\infty} &\le R_{\rm out}
\qquad (f\in B_k). \label{eq:bounded-readout}
\end{align}
Equivalently, the map
\[
\Psi_k:B_k\to H_k,
\qquad
\Psi_k(f):=h_f\circ f,
\]
is \(\kappa\)-co-Lipschitz from \((B_k,d_S)\) to \((H_k,\|\cdot\|_S)\), and its image is uniformly bounded on the sample.
Assume also that \(\erad_S(H_k)<\infty\), i.e.\ for every sign pattern
\(\sigma\) the averages \(n^{-1}\sum_i\sigma_ig(X_i)\), \(g\in H_k\), are
bounded above (for instance, \(H_k\) is uniformly bounded on the sample);
without this the left-hand side below is not a finite supremum.

Then there exists a universal constant \(c>0\) such that, for every \(\varepsilon>0\),
\begin{equation}
\erad_S(H_k)
 \ge 
c\sup_{\varepsilon>0}
\min \left\{
\kappa\varepsilon\sqrt{\frac{\log M(B_k,d_S,2\varepsilon)}{n}},
\frac{\kappa^2\varepsilon^2}{R_{\rm out}}
\right\}.
\label{eq:sudakov-sup}
\end{equation}
By packing-covering duality, the same conclusion holds with \(M\) replaced by \(N\), up to absolute constants.  When \(M(B_k,d_S,2\varepsilon)=\infty\), the first entry of the minimum is read as \(+\infty\), so the bound is its second entry.
\end{theorem}

\begin{proof}
Fix \(\varepsilon>0\), and let \(f_1,\dots,f_M\in B_k\) be a \(2\varepsilon\)-packing of \(B_k\) with respect to \(d_S\), where
\[
M=M(B_k,d_S,2\varepsilon)
\]
(if this packing number is infinite, take an arbitrary finite
\(2\varepsilon\)-packing and let its cardinality tend to infinity at the end).
Set
\[
g_j:=\Psi_k(f_j)=h_{f_j}\circ f_j\in H_k
\qquad (j=1,\dots,M).
\]
By \eqref{eq:lower-lip-readout},
\[
\|g_j-g_\ell\|_S
=
\|\Psi_k(f_j)-\Psi_k(f_\ell)\|_S
\ge
\kappa  d_S(f_j,f_\ell)
\ge
2\kappa\varepsilon
\qquad (j\neq \ell).
\]
Hence \(\{g_1,\dots,g_M\}\subset H_k\) is a \(2\kappa\varepsilon\)-packing in the empirical norm \(\|\cdot\|_S\). Moreover, by \eqref{eq:bounded-readout},
\[
\|g_j\|_{S,\infty}\le R_{\rm out}
\qquad (j=1,\dots,M).
\]

Applying the Bernoulli--Sudakov minoration to the bounded class
\(\{g_1,\dots,g_M\}\subset H_k\), we obtain
\[
\erad_S(H_k)
\ge
c\min \left\{
\kappa\varepsilon\sqrt{\frac{\log M}{n}},
\frac{\kappa^2\varepsilon^2}{R_{\rm out}}
\right\},
\]
for a universal constant \(c>0\).
Taking the supremum over \(\varepsilon\) yields \eqref{eq:sudakov-sup}.
\end{proof}

\paragraph{Remarks.}
(1) The strengthened assumption \eqref{eq:lower-lip-readout}--\eqref{eq:bounded-readout} is a genuine expressivity condition on the output-layer class \(H\): for each hidden map \(f\in B_k\), one is allowed to choose an output layer \(h_f\in H\) so that the discrete geometry of the word ball \((B_k,d_S)\) is realized inside the real-valued class \(H_k\). It is a sufficient condition for the \(\sqrt{k/n}\)- or \(\sqrt{\log k/n}\)-type lower bounds in the exponential or polynomial growth regimes, respectively, up to the depth-independent output-layer term in \cref{thm:hidden-decomp}(ii).
(2) Only the disjointness of the hidden maps on the sample enters: the
word-ball structure of \(B_k\) is not used, and the statement holds for an
arbitrary subset of \(C(\xdom,\xdom)\).

\subsection{Matching rates}

\begin{corollary}[Matching depth dependence under output-visible separation]
\label{cor:matching}
Suppose Assumption~\ref{ass:readout-realization-main} holds for every
\(k\) with constants independent of \(k\), and fix \(\eps_0>0\).  If
\(M(B_k,d_S,2\eps_0)\ge e^{\alpha k}\), then
\(\erad_S(\calH_k)\ge c\,\kappa\eps_0\sqrt{\alpha k/n}\) for
\(n\ge R_{\rm out}^2\alpha k/(\kappa^2\eps_0^2)\); if \(k\ge1\) and
\(M(B_k,d_S,2\eps_0)\ge k^{\beta}\), then
\(\erad_S(\calH_k)\ge c\,\kappa\eps_0\sqrt{\beta\log k/n}\) for
\(n\ge R_{\rm out}^2\beta\log k/(\kappa^2\eps_0^2)\).  Hence, when the
output-layer constants and the diameter in \cref{thm:hidden-decomp} are
depth-independent and the covering numbers obey the matching upper bounds
of \cref{prop:profiles}, the hidden complexity term has the same depth
dependence, \(\sqrt k\) or \(\sqrt{\log k}\), in the upper and in the lower
bound, up to constants and the depth-independent term \(\erad_S(H)\).
\end{corollary}

\begin{corollary}[Rates under fixed-scale packing lower bounds]
\label{cor:sudakov-rates}
If there exist \(\varepsilon_0>0\) and \(\alpha>0\) such that
\[
M(B_k,d_S,2\varepsilon_0)\ge e^{\alpha k},
\]
then
\begin{equation}
\erad_S(H_k)
 \ge 
c\min \left\{
\kappa\varepsilon_0\sqrt{\frac{\alpha k}{n}},
\frac{\kappa^2\varepsilon_0^2}{R_{\rm out}}
\right\}.
\label{eq:sudakov-exp}
\end{equation}
Likewise, if \(k\ge1\) and there exist \(\varepsilon_0>0\) and \(\beta>0\) such that
\[
M(B_k,d_S,2\varepsilon_0)\ge k^\beta,
\]
then
\begin{equation}
\erad_S(H_k)
 \ge 
c\min \left\{
\kappa\varepsilon_0\sqrt{\frac{\beta\log k}{n}},
\frac{\kappa^2\varepsilon_0^2}{R_{\rm out}}
\right\}.
\label{eq:sudakov-poly}
\end{equation}
\end{corollary}

\begin{proof}
    \eqref{eq:sudakov-exp}--\eqref{eq:sudakov-poly} follow by substitution.
    The first branch of the minimum in \eqref{eq:sudakov-exp} is the smaller
    one exactly when \(n\ge R_{\rm out}^2\alpha k/(\kappa^2\varepsilon_0^2)\),
    and in \eqref{eq:sudakov-poly} exactly when
    \(n\ge R_{\rm out}^2\beta\log k/(\kappa^2\varepsilon_0^2)\); this gives
    \cref{cor:matching}.  If \(\kappa,R_{\rm out}\) and \(\varepsilon_0\) do
    not depend on \(k\), the resulting lower bounds have the depth
    dependence \(\sqrt k\) and \(\sqrt{\log k}\), matching cases (iii) and
    (ii) of \cref{prop:profiles}.
\end{proof}

\subsection{Output-layer classes that separate hidden states}

The class
\[
H_R(\Phi)
:=
\bigl\{
h_w(x)=\langle w,\Phi(x)\rangle_{\calH}
\;:\;
\|w\|_{\calH}\le R
\bigr\}
\]
is natural, but the mere Lipschitz continuity of \(\Phi\) is in general sufficient only for the upper-bound side (\cref{thm:hidden-decomp}(ii)), not for the lower-bound side.
For the lower bound, one needs either a globally co-Lipschitz scalar observable (\cref{prop:global_scalar_observable}), or a finite-set
interpolation property on the reachable sample states (\cref{prop:linear-interpolation} and its
corollaries).

\begin{proposition}[Global scalar observable]%
\label{prop:global_scalar_observable}
Fix a sample \(S=(X_i)_{i=1}^n\) and let
\[
U_{k,S}:=\{f(X_i)\in X \;:\; f\in B_k,\ i\in[n]\}.
\]
Let \(\Phi:X\to\calH\) and
\[
H_{R_{\rm H}}(\Phi)
:=
\bigl\{
h_w(x)=\langle w,\Phi(x)\rangle_{\calH}
\;:\;
\|w\|_{\calH}\le R_{\rm H}
\bigr\}.
\]
Assume that there exists \(u\in\calH\) with \(\|u\|_{\calH}\le R_{\rm H}\) and constants \(\kappa,R_\Phi>0\) such that
\begin{align}
|\langle u,\Phi(x)-\Phi(y)\rangle_{\calH}| &\ge \kappa\, d(x,y)
\qquad (x,y\in U_{k,S}), \label{eq:scalar-colip}\\
\sup_{x\in X}\|\Phi(x)\|_{\calH} &\le R_\Phi. \label{eq:scalar-bdd}
\end{align}
Then the lower bound holds with the single choice
\[
\Psi_k(f):=h_u\circ f,
\qquad
h_u(x):=\langle u,\Phi(x)\rangle_{\calH},
\]
namely
\[
\|\Psi_k(f)-\Psi_k(g)\|_S\ge \kappa\, d_S(f,g)
\qquad (f,g\in B_k),
\]
and
\[
\|\Psi_k(f)\|_{S,\infty}\le R_{\rm H}R_\Phi
\qquad (f\in B_k).
\]
\end{proposition}

\begin{proof}
For \(f,g\in B_k\),
\[
\|\Psi_k(f)-\Psi_k(g)\|_S^2
=
\frac1n\sum_{i=1}^n
|\langle u,\Phi(f(X_i))-\Phi(g(X_i))\rangle|^2
\ge
\frac1n\sum_{i=1}^n
\kappa^2 d(f(X_i),g(X_i))^2
=
\kappa^2 d_S(f,g)^2,
\]
by \eqref{eq:scalar-colip}. Also,
\[
|\Psi_k(f)(x)|
=
|\langle u,\Phi(f(x))\rangle|
\le
\|u\|\,\|\Phi(f(x))\|
\le R_{\rm H}R_\Phi,
\]
by \eqref{eq:scalar-bdd}.
\end{proof}

Proposition~\ref{prop:global_scalar_observable} is the cleanest way in which a linear Hilbert-space output layer
satisfies the \emph{full} lower-Lipschitz assumption. Geometrically, it says that one scalar coordinate
of the feature map already observes the hidden-state geometry without collapse.

\begin{proposition}[Map-dependent finite-set interpolation criterion for linear heads]
\label{prop:linear-interpolation}
Fix \(f_1,\dots,f_M\in B_k\), and write
\[
z_{j,i}:=f_j(X_i)
\qquad
(j\in[M],\ i\in[n]).
\]
Let \(\Phi:X\to\calH\), and define
\[
H_{R_{\rm H}}(\Phi)
:=
\bigl\{
h_w(x)=\langle w,\Phi(x)\rangle_{\calH}
\;:\;
\|w\|_{\calH}\le R_{\rm H}
\bigr\}.
\]
For each \(j\), assume that the evaluation operator
\[
T_j:\calH\to \mathbb R^{n},
\qquad
T_j w := \bigl(\langle w,\Phi(z_{j,i})\rangle_{\calH}\bigr)_{i\in[n]},
\]
admits a right inverse \(R_j:\mathbb R^{n}\to\calH\) such that
\[
T_j\circ R_j=\mathrm{Id}_{\mathbb R^{n}},
\qquad
\|R_j\|_{\ell_2^n\to\calH}\le \Lambda_j
\]
with \(\Lambda_j\ge0\) (for \(n\ge2\) this forces \(z_{j,1},\dots,z_{j,n}\)
to be distinct, so no separate distinctness assumption is needed).
Then, for every family of target code vectors
\[
u^{(1)},\dots,u^{(M)}\in\mathbb R^n,
\]
there exist output layers \(h_j\in H_{R_{\rm H}}(\Phi)\), \(j\in[M]\), provided
\[
R_{\rm H}\ge \max_{j\in[M]}\Lambda_j\|u^{(j)}\|_2,
\]
such that
\[
h_j(f_j(X_i))=u_i^{(j)}
\qquad
(j\in[M],\ i\in[n]).
\]
Consequently, if
\[
\Bigl(
\frac1n\sum_{i=1}^n |u_i^{(j)}-u_i^{(\ell)}|^2
\Bigr)^{1/2}
\ge \rho
\qquad
(j\neq \ell),
\]
then
\[
\|h_j\circ f_j-h_\ell\circ f_\ell\|_S\ge \rho
\qquad
(j\neq \ell).
\]
If additionally \(\max_{j,i}|u_i^{(j)}|\le R_{\rm out}\), then
\[
\|h_j\circ f_j\|_{S,\infty}\le R_{\rm out}
\qquad (j\in[M]).
\]
In particular, for \(R_{\rm out}\)-bounded codes it is enough to take
\[
R_{\rm H}\ge \sqrt n\,R_{\rm out}\max_{j\in[M]}\Lambda_j,
\]
which is independent of the packing cardinality \(M\).
\end{proposition}

\begin{proof}
For each fixed \(j\), set \(w_j:=R_ju^{(j)}\) and \(h_j:=h_{w_j}\). Then
\(T_jw_j=u^{(j)}\), hence
\[
h_{w_j}(f_j(X_i))
=
\langle w_j,\Phi(f_j(X_i))\rangle
=
u_i^{(j)}.
\]
Moreover,
\[
\|w_j\|_{\calH}
\le
\Lambda_j\|u^{(j)}\|_2,
\]
so \(h_{w_j}\in H_{R_{\rm H}}(\Phi)\) under the stated bound on \(R_{\rm H}\).
The separation and sample-boundedness claims follow by evaluating the
constructed functions on the sample:
\[
(h_j\circ f_j)(X_i)=u_i^{(j)}.
\]
\end{proof}

\begin{corollary}[Finite-dimensional feature map]
\label{cor:finite-dim-feature}
Assume \(\calH=\mathbb R^m\) and \(n\le m\). If, for every \(j\), the feature matrix
\[
\Phi_j
:=
\bigl(\Phi(z_{j,i})\bigr)_{i\in[n]}
\in\mathbb R^{m\times n}
\]
has full column rank (which requires the \(z_{j,i}\) to be distinct), then Proposition~\ref{prop:linear-interpolation} applies with
\[
\Lambda_j=\sigma_{\min}(\Phi_j)^{-1}.
\]
Hence a sufficiently wide linear last layer over the feature map \(\Phi\) can realize any prescribed finite coding
on each individual reachable sample set \(\{z_{j,i}\}_{i\in[n]}\).
\end{corollary}

\begin{proof}
Here \(T_jw=\Phi_j^\top w\). Since \(\Phi_j\) has full column rank, \(T_j\) is surjective, and one may take the
Moore--Penrose right inverse. Its operator norm is \(\sigma_{\min}(\Phi_j)^{-1}\).
\end{proof}

\begin{corollary}[RKHS / kernel output layer]
\label{cor:rkhs-readout}
Let \(K\) be a strictly positive definite kernel on \(X\), with canonical feature map
\(\Phi:X\to\calH_K\). Then Proposition~\ref{prop:linear-interpolation} applies to
\[
H_{R_{\rm H}}(K)
:=
\bigl\{
h_w(x)=\langle w,\Phi(x)\rangle_{\calH_K}
\;:\;
\|w\|_{\calH_K}\le R_{\rm H}
\bigr\},
\]
and, provided the points \(z_{j,1},\dots,z_{j,n}\) are distinct (so that the
Gram matrix below is positive definite), one may take
\[
\Lambda_j=\lambda_{\min}(G_j)^{-1/2},
\qquad
G_j:=\bigl(K(z_{j,i},z_{j,i'})\bigr)_{i,i'\in[n]}.
\]
Therefore, whenever the individual Gram matrices \(G_j\) are uniformly
well-conditioned, the RKHS linear output layer has the map-dependent finite-coding
property needed for the packing-wise Sudakov lower bound.  The output-layer norm
radius scales with \(\max_j\|u^{(j)}\|_2\), not with
\((\sum_j\|u^{(j)}\|_2^2)^{1/2}\).
\end{corollary}

\begin{proof}
Fix \(j\). For \(u\in\mathbb R^{n}\), let \(c=G_j^{-1}u\) and set
\[
w:=\sum_{i=1}^n c_i\,\Phi(z_{j,i}).
\]
Then
\[
h_w(z_{j,i})=u_i
\qquad
(i\in[n]),
\]
and
\[
\|w\|_{\calH_K}^2
=
u^\top G_j^{-1}u
\le
\lambda_{\min}(G_j)^{-1}\|u\|_2^2.
\]
Thus \(\|R_j\|_{\ell_2^n\to\calH_K}\le \lambda_{\min}(G_j)^{-1/2}\).
\end{proof}

Corollary~\ref{cor:rkhs-readout} gives a realistic interpretation of the lower-bound assumption:
a linear output layer over a feature map \(\Phi\) works if each reachable sample set is sufficiently
``feature-separated'' so that the corresponding evaluation map is well-conditioned. In finite dimension, this is a rank/singular-value
condition on each last-layer feature matrix; in an RKHS, it is a lower bound on the smallest eigenvalue of each
individual Gram matrix.

\subsection{When the converse is silent}
\label{sec:sudakov-collapse}

The converse is a conditional diagnostic and is not used in the upper
bounds.  If the architecture collapses or over-smooths its hidden
representations, Assumption~\ref{ass:readout-realization-main} fails and
the lower bound becomes weak or vacuous: a packing lower bound can only
certify the growth of the quotient of the reachable states by the relation
\(x\sim y\) if \(h(x)=h(y)\) for all \(h\in H\), which may be much smaller
than the packing in the state metric.  Near collapse the same effect appears
through the constants, since \(\kappa\) becomes small and, for linear or
RKHS output layers, the norm needed to realize a separating code grows with
the conditioning of the reachable feature matrices.  The upper-bound
analysis still applies in these cases, possibly conservatively.

\section{Growth mechanisms: conditions, proofs, and examples}
\label{sec:growth-conditions-proof}

This appendix states and proves the growth conditions summarized in
\cref{sec:growth-conditions}, one subsection per mechanism: saturation
(P1, P1'), polynomial growth (P2), exponential growth under separation
(E1, E1', E2), and memory-preserving expansion (E3).  Each subsection
follows the same order: the condition and its proof, one worked example
that verifies it, and closing remarks.  The conventions and the basic
covering-number facts are those of \cref{sec:app-remarks}; in this appendix
\(B(k,F)\) is called the word ball of radius \(k\) of the semigroup
generated by \(F\).

\subsection{Saturation under contraction and equicontinuity (P1, P1')}
\label{sec:app-saturation}

The two conditions below make the covering number of \(B(k,F)\) bounded
in \(k\) at every resolution.  The first uses compactness of the state
space and equicontinuity of the compositions; the second uses uniform
contraction toward an invariant set and applies on unbounded state spaces.
Both are used in \cref{prop:profiles}(i) and in the examples of
\cref{sec:examples-regime,sec:app-ode,sec:app-cot,sec:app-relu}.

\begin{namedcondition}{P1}[Equicontinuous semigroups on compact domains saturate]\label{cond:p1}
Assume $\xdom$ is compact.
Suppose (at least) one of the following assumptions is satisfied:
\begin{enumerate}
    \item[1.] Semigroup $\langle F \rangle$ is (pre)compact,
    \item[2a.] Semigroup $\langle F \rangle$ is equicontinuous,
    \item[2b.] Semigroup $\langle F \rangle$ is uniformly Lipschitz: $\lip \langle F \rangle < \infty$, or
    \item[2c.] Generators $F$ are non-expanding: $\lip F \le 1$.
\end{enumerate}
Then, closure semigroup $G:=\overline{\langle F\rangle}^{d_\infty}\subset C(\xdom,\xdom)$
is compact and equicontinuous, and 
for all $\eps>0$ and all $k$
\[
N\left(B(k,F),\eps\right) \le N\left(G,\eps\right) \quad (< \infty),
\]
hence no dependence on $k$.  Under assumption~1 the compactness of
\(\xdom\) is not needed: total boundedness of \(\langle F\rangle\) alone
gives \(N(B(k,F),\eps)\le N(\langle F\rangle,\eps)<\infty\) for all \(k\).
\end{namedcondition}

\begin{proof}
If $\xdom$ is compact, then $C(\xdom,\xdom)$ is complete in $d_\infty$.
Thus the closure of any precompact subset is compact. By the compact
self-map version of Ascoli's theorem (see the remark on the uniform metric above), a family
in \(C(\xdom,\xdom)\) is precompact in \(d_\infty\) if and only if it is
equicontinuous. Assumption 1 gives compactness of \(G\) directly, and
Assumption 2a gives it by Arzelà--Ascoli. Assumptions 2b and 2c imply
2a, since a uniform Lipschitz bound is a common modulus of continuity
and non-expanding generators generate a non-expanding semigroup.
Trivial inclusion $B(k,F)\subseteq G$ yields the bound, and compactness of $G$ gives finiteness of $N(G,\eps)$.
\end{proof}

\begin{namedcondition}{P1'}[Contraction to an invariant set]\label{cond:p1-ucont}
Assume 
\begin{enumerate}
    \item (uniform contraction) \(\sup_{f\in F}\lip(f)\le c\) for some \(0<c<1\),
    \item (invariant set) there exists a nonempty \(F\)-invariant set \(A \subset \xdom\) (i.e., \(f(A)\subset A\) for all \(f\in F\)), and
    \item (bounded absorbing set) there exist $L \in \NN$ and a bounded set $K \subset \xdom$ such that $f(\xdom) \subset K$ for all depth-$L$ maps $f \in F^L$.
\end{enumerate}
Then for every \(\eps>0\), with
\[m(\eps) := L + \left\lceil \log_{1/c}\left(\tfrac{2 \diam(K)}{\eps}\right)\right\rceil,\]
for \emph{every} \(k\ge0\),
\[
N\left(B(k,F),d_\infty,\eps\right)
\le
N\left(A,d,\tfrac{\eps}{2}\right)
+
\sum_{\ell=0}^{m(\eps)-1} N\left(F^\ell,d_\infty,\eps\right),
\]
an inequality in \(\NN\cup\{\infty\}\) whose right-hand side is finite as
soon as \(A\) is totally bounded (e.g.\ compact) and each
\(N(F^\ell,d_\infty,\eps)\), \(\ell<m(\eps)\), is finite (for instance when
\(F\) is totally bounded in \(d_\infty\), by sub-multiplicativity).
In particular, the right-hand side is \emph{independent of \(k\)},
so growth in \(k\) saturates at any fixed \(\eps\)
to the entropy of the invariant set \(A\).
\end{namedcondition}

\begin{proof}
Pick any \(a_0\in A\).
For any word \(w\in F^\ell\),
we have \(\lip(w)\le c^\ell\) and \(w(a_0)\in A\) (invariance).
If \(\ell\ge m(\eps)\), then
\[\sup_{x\in \xdom} d\left(w(x),w(a_0)\right) \le c^{\ell-L} \diam(K), \quad \text{(bounded absorbing set)}\]
and the right-hand side is \(\le \eps/2\), i.e.
\[d_\infty\left(w, \mathrm{const}_{w(a_0)}\right) \le \eps/2.\]
Hence all sufficiently deep words,
i.e. \(\bigcup_{\ell\ge m(\eps)} F^\ell\),
are covered by the \(\eps/2\)-thickening of the set of constant maps landing in $A$: 
\({\mathrm{const}_q: q\in A}\).
These constants at resolution \(\eps/2\) are parameterized by an
\(\eps/2\)-net of \(A\), giving the bound \(N(A,\eps/2)\).
For the finitely many shallow layers \(\ell < m(\eps)\), simply cover
each \(F^\ell\) at scale \(\eps\) and sum over \(\ell\). These terms are
independent of \(k\); in the usual compact-generator setting they are
finite by continuity of composition and compactness/totally boundedness
of \(F\).  Since \(B(k,F)=\bigcup_{\ell\le k}F^\ell\), the two covers
together cover \(B(k,F)\) for every \(k\ge0\) (for \(k<m(\eps)\) only the
shallow part is used), which proves the bound for all \(k\).  Neither
compactness of \(A\) nor of \(K\) was used, only invariance and
boundedness.
\end{proof}

\begin{example}[Cantor attractor (compact, free, contractive)] \label{ex:p1-cantor}
Let \(\xdom=[0,1]\) and \(F={f_0(x)=x/3, f_1(x)=(x+2)/3}\). Then
\(\sup\lip(f_i)=1/3=c<1\). The attractor %
\(A\) is the middle--third Cantor set \(C\) (compact, \(F\)-invariant).
By \cref{cond:p1-ucont} (with \(L=0\), \(K=[0,1]\)), for every \(\eps>0\)
and every \(k\), \[
N\left(B(k,F),\eps\right) \le N\left(C,\eps/2\right)
+\sum_{\ell<m(\eps)}N(F^\ell,\eps),
\qquad m(\eps)=\lceil\log_{3}(2/\eps)\rceil,
\] independent of \(k\). (Indeed, every \(w\in F^\ell\) has the
form \(w(x)=3^{-\ell}x+b_w\) and hence is \(\eps/2\)-close to the
constant map \(x\mapsto b_w\in C\).)
\end{example}

\paragraph{Remarks.}
(1) When \(\xdom\) is compact, one may take \(A=K=\xdom\) and \(L=0\) in
\cref{cond:p1-ucont}.
(2) A single saturating map \(\sigma\), such as \(\tanh\), the logistic
function, or clipping, provides the bounded absorbing set
\(K=\sigma(\xdom)\) with \(L=1\).
(3) Deep words are \emph{almost constant} (their images have diameter \(\le c^{\ell-L}\diam(K)\)),so at scale \(\eps\) only the landing point \(w(a_0)\in A\) matters; hence the dependence on \(k\) disappears once \(\ell\) exceeds an \(\eps\)-dependent ``memory length'' \(m(\eps)\).

\subsection{Polynomial growth under nilpotent control (P2)}
\label{sec:app-polynomial}

The next condition covers layer classes that are not contractive but whose
compositions are organized by a group of polynomial growth.  It gives the
fixed-scale bound \(N(B(k,F),d_\infty,\eps)\lesssim(1+k/\eps)^D\) and the
diameter bound \(D_k(S)\lesssim k\) used in \cref{prop:profiles}(ii) and
(iv).

\begin{namedcondition}{P2}[Nilpotent control gives polynomial growth]\label{cond:p2-nilp}
Let \((H,d_H)\) be a compactly generated group with identity \(e\), and
assume the length \(g\mapsto d_H(e,g)\) is subadditive:
\(d_H(e,gh)\le d_H(e,g)+d_H(e,h)\).
Assume that its metric balls have polynomial entropy of degree \(D\ge0\):
there is \(1\le C_H<\infty\) such that for all \(R\ge0\) and \(\delta>0\),
\[
N\bigl(B_H(e,R),d_H,\delta\bigr)
\le
C_H\left(1+\frac{R}{\delta}\right)^D
\]
(in particular these covering numbers are finite).
Suppose \(H\) acts on \(\xdom\) through a homomorphism
\(\alpha:H\to C(\xdom,\xdom)\), and suppose there is a bounded set
\(S\subset H\) such that \(F\subset \alpha(S)\). Finally assume the
orbit map is Lipschitz in the uniform metric:
there is \(0<L_\alpha<\infty\) such that for all \(g,h\in H\),
\[
d_\infty\bigl(\alpha(g),\alpha(h)\bigr)\le L_\alpha d_H(g,h).
\]
Then, with \(R_S:=\sup_{s\in S}d_H(e,s)\) and
\(C:=C_H\max(1,R_SL_\alpha)^D\), for every \(\eps>0\) and every
\(k\ge0\),
\[
N\left(B(k,F),d_\infty,\eps\right)
\le
C\left(1+\frac{k}{\eps}\right)^D ,
\]
and moreover \(\diam_\infty B(k,F)\le 2L_\alpha R_Sk\), hence
\(D_k(S)\le 2L_\alpha R_Sk\) for every sample \(S\).
\end{namedcondition}

\begin{proof}
Let \(R_S:=\sup_{s\in S}d_H(e,s)<\infty\). Any composition of at most
\(k\) maps from \(F\) has a representative \(\alpha(g)\) with
\(g=s_k\cdots s_1\) and \(s_i\in S\). By subadditivity,
\[
d_H(e,g)\le \sum_{i=1}^k d_H(e,s_i)\le kR_S.
\]
Thus
\[
B(k,F)\subset \alpha\bigl(B_H(e,kR_S)\bigr).
\]
The \(L_\alpha\)-Lipschitz bound on \(\alpha\) gives
\[
N\left(B(k,F),d_\infty,\eps\right)
\le
N\left(B_H(e,kR_S),d_H,\eps/L_\alpha\right).
\]
The assumed polynomial entropy bound on \(H\) gives
\(C_H(1+kR_SL_\alpha/\eps)^D\le C_H\max(1,R_SL_\alpha)^D(1+k/\eps)^D\),
which is the claim (for \(k=0\) as well).  The diameter bound is proved in remark (2) below.
\end{proof}

\begin{example}[Upper-triangular unipotent group (non-compact, nilpotent)] \label{ex:p2-noncpt-ut}
Let
\[
H=UT_d(\RR)=\{I_d+N:\ N \text{ is strictly upper triangular}\}
\]
with a right-invariant Riemannian metric, and let \(\xdom=H\).  Let
\(\alpha(g)\) be left translation, \(\alpha(g)(x)=gx\). Then
\[
d_\infty(\alpha(g),\alpha(h))=d_H(g,h),
\]
because \(d_H\) is right-invariant. For any compact \(S\subset H\) and
\(F=\alpha(S)\), \cref{cond:p2-nilp} gives
\[
  N(B(k,F),d_\infty,\eps)\lesssim (1+k/\eps)^D,
  \qquad
  D=\frac{d(d-1)(d+1)}{6}.
\]
Each generator \(\alpha(g)\), \(x\mapsto gx\), is a linear map of the
ambient matrix space \(\RR^{d\times d}\), so by
\cref{prop:implementation}(c) every word in \(B(k,F)\) is computed exactly by
a ReLU network of depth \(k\) and width \(2d^2\); no approximation of the
transitions, and hence no exponential width, is needed to implement this
example.\end{example}

\paragraph{Remarks.}
(1) The Guivarc'h--Bass homogeneous dimension 
    bridges nilpotency with polynomial volume and entropy growth.
    Connected simply connected nilpotent Lie groups with left-invariant
    Riemannian metrics, and finitely generated nilpotent groups with word
    metrics, satisfy the displayed entropy assumption with this dimension.
    See \citet{Breuillard2014} and \citet[Chapter~VII]{DelaHarpe2000geometricgroup} for the growth theory of nilpotent groups used here.
(2) The inclusion \(B(k,F)\subset\alpha(B_H(e,kR_S))\) in the proof also
controls the diameter of the word ball: for \(g,h\in B_H(e,kR_S)\),
\(d_\infty(\alpha(g),\alpha(h))\le L_\alpha d_H(g,h)\le 2L_\alpha R_Sk\), so
\(\diam_\infty(B(k,F))\le 2L_\alpha R_Sk\) and a fortiori
\(D_k(S)\le 2L_\alpha R_Sk\).  On a compact state space the diameter is of
course bounded by \(\diam(\xdom)\).  These layerwise bounds are the input to
\cref{prop:profiles}(ii) and (iv).

\subsection{Exponential growth under separation (E1, E1', E2)}
\label{sec:app-exponential}

The conditions below give the lower bound \(N(B(k,F),d_\infty,\eps)\ge r^k\)
for a class of \(r\) generators at a fixed resolution.  All of them are
proved with \cref{lem:probes-packing}: one exhibits, for each depth,
\(r^k\) compositions whose values at suitable probe points are uniformly
separated.  E1 and E1' use one probe point for all words; E2 uses one probe
per word and applies on compact state spaces where the maps expand on
part of the space and reset elsewhere.  With a bounded diameter these
conditions give the \(\sqrt k\) profile of \cref{prop:profiles}(iii), and
\cref{cor:matching} shows that the profile is attained when the separation
is visible at the output layer.

\begin{namedcondition}{E1}[Free semigroup with one-point uniform separation] \label{cond:e1-free-iso}
Let \(F=\{f_1,\dots,f_r\}\subset C(\xdom,\xdom)\) with \(r\ge2\). Assume:
\begin{enumerate}
\def\labelenumi{\arabic{enumi}.}
\item
  (Uniform separation at a base point) There exist
  \(x_\ast\in \xdom\) and \(\delta>0\) such that for all \(k\) and all
  distinct words \(u,v\in[r]^k\), writing
  \(f_u:=f_{i_k}\circ\cdots\circ f_{i_1}\) for \(u=(i_1,\dots,i_k)\),
  $d\big(f_u(x_\ast),f_v(x_\ast)\big) \ge \delta$.
\end{enumerate}
Then, for every \(k\) and every \(\varepsilon<\delta/2\),
\[
N\big(B(k,F),d_\infty,\varepsilon\big) \ge r^k.
\]
In particular \(u\mapsto f_u\) is injective on \([r]^k\) for every \(k\)
(freeness at each length), so no separate freeness hypothesis is needed.
\end{namedcondition}

\begin{proof}
Apply \cref{lem:probes-packing} with \(P=\{x_\ast\}\). For each fixed
\(k\), the \(r^k\) words of length \(k\) give \(r^k\) maps whose images
at \(x_\ast\) are \(\delta\)-separated; hence for
\(\varepsilon<\delta/2\) at least \(r^k\) balls are needed to cover
\(B(k,F)\). %
\end{proof}

\begin{namedcondition}{E1'}[Uniform same-length coding] \label{cond:e1p-theoremC}
Let \(F=\{f_1,\dots,f_r\}\subset C(\xdom,\xdom)\) with \(r\ge2\), and
suppose (uniform same-length separation) that there exist
\(x_\ast\in \xdom\) and \(c>0\) such that, for all words \(u\ne v\)
of the same length,
\[
d\big(u(x_\ast),v(x_\ast)\big) \ge c .
\]
Then, for every \(k\) and every \(\varepsilon < c/2\), \[
N\big(B(k,F),d_\infty,\varepsilon\big) \ge r^k.
\]
This is \cref{cond:e1-free-iso} with \(\delta=c\): neither isometry nor
freeness nor finiteness of \(d_\infty\) on \(\langle F\rangle\) is needed
for the lower bound.  The isometric setting in which it is used below,
\(F\subset\iso(\xdom)\) with \(d_\infty(g,h)<\infty\) for
\(g,h\in\langle F\rangle\) (as when \(d\) is bi-invariant,
\(d(axb,ayb)=d(x,y)\)), only serves to make the uniform metric finite on
the semigroup.
\end{namedcondition}

\begin{proof}
Apply \cref{lem:probes-packing} with \(P=\{x_\ast\}\), exactly as for
\cref{cond:e1-free-iso}. %
\end{proof}

\begin{example}[Free group with a bi-invariant metric (non-compact, free, isometric)] \label{ex:e1-free-bounded}
Let \(\xdom=G=F_r\) \((r\ge2)\). Let \[
S := \left\{ x a_i x^{-1}, x a_i^{-1} x^{-1} \;\middle|\; x\in F_r, i=1,\dots,r \right\}
\] and define the \emph{conjugacy-invariant} word metric \[
d_S(g,h):=|g^{-1}h|_S,
\] the shortest length in the alphabet \(S\). This metric is
\emph{bi-invariant}, hence all left translations \(L_g\) are
isometries and, crucially, \[
d_\infty(L_g,L_h)
=\sup_{x} d_S(gx,hx)
=\sup_{x} d_S \big(e, x^{-1}g^{-1}hx\big)
=d_S(e,g^{-1}h)<\infty
\] (the supremum is independent of \(x\)).

Let \(F=\{L_{a_1},\dots,L_{a_r}\}\). The positive semigroup is free.
With \(x_\ast=e\), for words \(u\neq v\) of the same length, \[
d_S\big(L_u(x_\ast),L_v(x_\ast)\big)=d_S(u,v)
=|u^{-1}v|_S \ge 1,
\] so \cref{cond:e1p-theoremC} applies with \(c=1\): for every \(\varepsilon<1/2\), \[
N\big(B(k,F),d_\infty,\varepsilon\big) \ge r^k.
\]
\end{example}
With the standard word metric instead, \(d_\infty(L_g,L_h)=\infty\) for \(g\ne h\); the lower bound would still hold, since only the probe \(x_\ast=e\) is used, but the bi-invariant metric keeps \(d_\infty\) finite.

\paragraph{Ping-pong coding.}
The classical \emph{Ping-Pong} (\emph{Table-Tennis}, or \emph{Schottky}) lemma \citep[e.g.][Chapter~II.B]{DelaHarpe2000geometricgroup} is a sufficient condition for freeness of generators, which itself does not immediately imply the exponential growth since the generators might be contractive. Therefore, the condition below states the uniform coding separation used in the entropy argument directly; the examples verify it through expanding branches and resets.

Here a \emph{chamber} \(U_i\subset\xdom\) is the active region for the
generator \(f_i\).  A \emph{coding core} \(V_i\subset U_i\) is a smaller
nonempty set on which \(f_i\) has enough coverage to realize the next
symbol.  The \emph{anchors} form a finite set \(A=\{a_1,\dots,a_r\}\)
of reset states; the points \(a_i\) need not be pairwise distinct unless
an example says so.  A \emph{marker} \(q\in\xdom\) is the target state
for correctly coded trajectories, and it is required to be separated from
the anchor set.  A \emph{probe} \(x_u\in\xdom\) is an input point chosen
for a word \(u\); it is constructed from the chambers and is not fixed
across all words.

\begin{namedcondition}{E2}[Ping-pong coding] \label{cond:e2-pingpong}
Let \(F=\{f_1,\dots,f_r\}\subset C(\xdom,\xdom)\) with \(r\ge2\).
Assume that there exist sets \(V_i\subset U_i\subset\xdom\),
anchors \(a_i\in\xdom\), a marker \(q\in\xdom\), and a constant
\(\alpha>0\) such that, with \(A:=\{a_1,\dots,a_r\}\) and
\[
Q:=\{q\}\cup\bigcup_{j=1}^r V_j,
\]
the following conventions and conditions hold.
The marker \(q\) is not an anchor, quantitatively enforced by item~4
below.  No global location assumption such as \(A\subset\bigcup_i U_i\)
or \(A\cap\bigcup_i U_i=\emptyset\) is imposed; anchors may lie inside
or outside chambers.  The required anchor condition is the invariance in
item~3.
\begin{enumerate}
\def\labelenumi{\arabic{enumi}.}
\item
  (Disjoint chambers)
  \[
  U_i\cap U_j=\emptyset\qquad (i\ne j).
  \]
\item
  (Full coding cores) For every \(i\),
  \[
  Q\subset f_i(V_i)
  \]
  (in particular each \(V_i\) is nonempty, since \(q\in f_i(V_i)\)).
\item
  (Reset and anchor invariance) For every \(i\),
  \[
  f_i(\xdom\setminus U_i)=\{a_i\},
  \qquad
  f_i(A)\subset A .
  \]
\item
  (Marker--anchor separation)
  \[
  \dist(\{q\},A)\ge \alpha .
  \]
\end{enumerate}
Then, for every \(k\) and every \(\varepsilon<\alpha/2\),
\[
N\big(B(k,F),d_\infty,\varepsilon\big) \ge r^k.
\]
Moreover, the map \(u\mapsto f_u\) is injective on words of each fixed
length.
\end{namedcondition}

\begin{proof}
Fix \(k\) and a word \(u=i_k\cdots i_1\in[r]^k\).  Set \(z_k=q\).
Since \(Q\subset f_{i_s}(V_{i_s})\), we may choose recursively
\(z_{s-1}\in V_{i_s}\) such that
\[
f_{i_s}(z_{s-1})=z_s,\qquad s=k,k-1,\dots,1.
\]
Put \(x_u:=z_0\).  Then \(f_u(x_u)=q\).  Now let
\(v=j_k\cdots j_1\ne u\), and let \(t\) be the first index, in the
order of application, with \(j_t\ne i_t\).  The first \(t-1\) steps
agree, so the \(v\)-trajectory is at \(z_{t-1}\in V_{i_t}\subset U_{i_t}\).
Since the chambers are disjoint, \(z_{t-1}\notin U_{j_t}\).  Hence
\(f_{j_t}\) resets this trajectory to \(a_{j_t}\in A\), and all later
generators keep it in \(A\) by anchor invariance.  Therefore
\[
d\big(f_u(x_u),f_v(x_u)\big)\ge \dist(\{q\},A)\ge \alpha .
\]
Thus, for \(P_k:=\{x_u\mid u\in[r]^k\}\), the \(r^k\) evaluation vectors
\(\{\eval_{P_k}(f_u)\mid u\in[r]^k\}\) are pairwise \(\alpha\)-separated.
The claim follows from \cref{lem:probes-packing}.
\end{proof}

\begin{example}[Ping-pong over the full shift (compact, expansive)] \label{ex:e2-pingpong-subshift}
Fix an alphabet \(\calA=[r] \cup \{\bullet\}\) with \(r\ge2\)
\emph{active} symbols \([r]\) and one extra \emph{padding} symbol \(\bullet\).
Let \[
\xdom=\calA^{\NN}=\{x=(x_0,x_1,x_2,\dots)\mid x_j\in\calA\}
\] be the (one-sided) full shift. Equip \(\xdom\) with the standard
ultrametric \(d_\theta\) for some fixed \(\theta\in(0,1)\): \[
d_\theta(x,y)=
\begin{cases}
0 & x=y\\
\theta^{\min\{n\ge 0 \mid x_n\neq y_n\}} & x\neq y
\end{cases}
\] Then \((\xdom,d_\theta)\) is compact, totally disconnected, and the left
shift \(\sigma(x)_n=x_{n+1}\) is \(L\)-Lipschitz with
\(L=\theta^{-1}>1\) (hence expansive).

For each \(a\in\calA\), write the clopen 1-cylinder
\(\ssq{a}=\{x\in \xdom\mid x_0=a\}\), and the \emph{anchor} sequence
\(\bar a=(a,a,a,\dots)\).

Define \(F=\{f_1,\dots,f_r\}\subset C(\xdom,\xdom)\) by
\[
f_a(x)=
\begin{cases}
\sigma(x) & x\in \ssq{a}\\[2mm]
\bar a & x\notin \ssq{a}
\end{cases}
\qquad a\in[r].
\] Because \(\ssq{a}\) is clopen, each \(f_a\) is continuous; on \(\ssq{a}\) it
is \(\sigma\) (Lipschitz constant \(\theta^{-1}>1\)), and on
\(\xdom\setminus\ssq{a}\) it is constant. Thus each \(f_a\) has an
expanding branch, although it is not globally expanding because of the
reset branch.

Let \(U_a:=V_a:=\ssq{a}\) (clopen chambers and coding cores),
\(a\in[r]\), let \(A:=\{\bar a\mid a\in[r]\}\), and let
\(q:=\bar\bullet\). Here \(q\notin\bigcup_{a\in[r]}U_a\), while each
anchor \(\bar a\) lies in its corresponding chamber \(U_a\). Then:

\begin{itemize}
\item
  Pairwise separation. If \(a\neq b\), then for any
  \(x\in U_a\), \(y\in U_b\), \(d_\theta(x,y)=\theta^{0}=1\). Hence
  \(\dist(U_a,U_b)=1\).
\item
  Expansion/coverage on own domain. For each \(a\),
  \(f_a|_{U_a}=\sigma\) maps \(U_a\) bijectively onto \(\xdom\) (surjective
  and expanding).
\item
  Reset off domain. On \(\xdom\setminus U_a\), \(f_a = \const_{\bar a}\)
  (exact reset; the ``reset diameter'' is \(0\)).
\item
  Anchor invariance and marker separation. For every \(a\in[r]\),
  \(f_a(A)\subset A\), and \(\dist(\{q\},A)=1\).
\end{itemize}

These data yield \cref{cond:e2-pingpong} with
\(\alpha=1\) (the chambers are even at distance \(1\)). Hence the
separation scale can be taken as \(\delta=1\).

Let a word \(u=a_k\cdots a_1\in[r]^k\). Define the
\emph{tail-padded probe} \[
x_u:=\big(a_1,a_2,\dots,a_k,\underbrace{\bullet,\bullet,\bullet,\dots}_{\text{all }\bullet}\big)\in \xdom.
\] Then, by construction, \[
f_u(x_u)=\sigma^k(x_u)=\bar\bullet,
\] because each of the first \(k\) steps sees the correct chamber and
applies \(\sigma\), peeling off the \(k\)-letter prefix and revealing
the all-\(\bullet\) tail.

If \(v=b_k\cdots b_1\neq u\), let \(t\) be the \emph{rightmost} index
with \(a_t\neq b_t\). When applying \(f_v\) to \(x_u\), the first
\(t-1\) letters match and act as shifts; at step \(t\) we apply
\(f_{b_t}\) to a sequence whose \(0\)-coordinate is \(a_t\neq b_t\), so
\(f_{b_t}\) resets to the anchor \(\bar{b_t}\). From then on, the state
never contains \(\bullet\) at the \(0\)-coordinate (subsequent resets
only use anchors \(\bar{b_s}\) with \(b_s\in[r]\), and
\(\sigma\) preserves \(0\)-coordinate \(b_s\) on the constant sequence
\(\bar{b_s}\)). Consequently \[
f_v(x_u)\neq \bar\bullet.
\] Therefore \[
d_\theta\big(f_u(x_u),f_v(x_u)\big)=d_\theta\big(\bar\bullet,f_v(x_u)\big)=1.
\] This shows \emph{freeness}: distinct words \(u\neq v\) define
distinct maps \(f_u\neq f_v\).

Fix \(k\in\NN\) and consider the finite probe set \[
P_k:=\{x_u \mid u\in[r]^k\}\subset \xdom,
\] of cardinality \(r^k\). For any distinct \(u,v\) we have just seen
that at the coordinate \(x_u\), \[
d_\theta\big(f_u(x_u),f_v(x_u)\big)=1\ge \delta.
\] Thus the \(r^k\) vectors \({E_{P_k}(f_u)}_{|u|=k}\subset \xdom^{P_k}\)
are pairwise \(1\)-separated in the max metric. By \cref{lem:probes-packing}, for every \(\varepsilon<\tfrac{1}{2}\), \[
N\big(B(k,F),d_\infty,\varepsilon\big) \ge r^k.
\]
\end{example}

\paragraph{Remarks.}
Only the disjointness of the chambers is used in the exact-reset proof.
The interval and symbolic examples below satisfy the stronger positive
separation \(\dist(U_i,U_j)\ge\Delta>0\), which is the robust form needed
if one allows approximate resets or interpolation collars.

\subsection{Memory-preserving expansion (E3)}
\label{sec:app-memory}

Expansion alone does not produce entropy growth beyond \(r^k\), because
later layers may erase earlier distinctions.  The construction below adds
a memory: each layer writes a symbol, shifts the stored symbols, and
expands them.  The entropy then accumulates multiplicatively across layers,
and the two corollaries give \(\log N\asymp k^2\) and \(\log N\asymp\lambda^{pk}\);
these are the two cases in which the layerwise bound of \cref{prop:envelope} is
attained (\cref{sec:proof-envelope}).

\begin{namedcondition}{E3}[Memory-preserving expansion]\label{cond:e3}
Let $E$ be a normed space, let $G\subset E$ be an arbitrary subset, and fix $\lambda>1$.
Set
\[
X:=\ell_\infty(E)
=
\Bigl\{x=(x_1,x_2,\dots): \sup_{j\ge 1}\|x_j\|_E<\infty\Bigr\},
\]
equipped with the bounded sup metric
\[
d_X(x,y):=\sup_{j\ge 1}\min\{1,\|x_j-y_j\|_E\}.
\]
We write \(d_\infty\) for the induced uniform metric on
\(C(X,X)\). This bounded state metric keeps the maps below at finite
uniform distance while preserving the small-scale expansion estimates
used in the lower bounds.
Define the reset map, the expander, and the writer family by
\[
r(x):=0,
\qquad
A(x_1,x_2,\dots):=(\lambda x_1,\lambda x_2,\dots),
\]
and, for each $u\in G$,
\[
g_u(x_1,x_2,\dots):=(u,x_1,x_2,\dots).
\]
Let
\[
F:=\{r,A\}\cup\{g_u:u\in G\}.
\]
For each $k\ge 1$ and each $u=(u_1,\dots,u_k)\in G^k$, define the word
\[
w_u
:=
A\circ g_{u_k}\circ A\circ g_{u_{k-1}}
\circ \cdots \circ
A\circ g_{u_1}\circ r.
\]
Let
\[
W_k:=\{w_u:u\in G^k\}\subset B(2k+1,F).
\]
Then the following hold.

\medskip
\noindent
{\rm (i)} For every $x\in X$,
\[
w_u(x)=
(\lambda u_k,\lambda^2 u_{k-1},\dots,\lambda^k u_1,0,0,\dots).
\]
In particular, each $w_u$ is a constant map.

\medskip
\noindent
{\rm (ii)} For $u=(u_1,\dots,u_k)$ and $v=(v_1,\dots,v_k)$,
\[
d_\infty(w_u,w_v)
=
\max_{1\le j\le k}
\min\{1,\lambda^{k-j+1}\|u_j-v_j\|_E\}.
\]

\medskip
\noindent
{\rm (iii)} For every \(\varepsilon>0\),
\[
N(W_k,d_\infty,\varepsilon)
\le
\prod_{j=1}^k
N \left(G,\|\cdot\|_E,\frac{\varepsilon}{\lambda^{k-j+1}}\right),
\]
and for every \(0<\varepsilon<1/2\),
\[
\prod_{j=1}^k
M \left(G,\|\cdot\|_E,\frac{2\varepsilon}{\lambda^{k-j+1}}\right)
\le
N(W_k,d_\infty,\varepsilon).
\]

Consequently, by monotonicity of the (external) covering number under
\(W_k\subset B(2k+1,F)\),
\[
N(B(2k+1,F),d_\infty,\varepsilon)\ge N(W_k,d_\infty,\varepsilon),
\]
so the word-ball entropy inherits a multiplicative layer-by-layer growth law.
\end{namedcondition}

\begin{proof}
We first prove (i). Since $r$ is the constant-zero map, the input $x$ is irrelevant after the first step. Starting from $0=(0,0,\dots)$,
\[
g_{u_1}(0)=(u_1,0,0,\dots),\qquad
A g_{u_1}(0)=(\lambda u_1,0,0,\dots).
\]
Applying $g_{u_2}$ and then $A$ gives
\[
A g_{u_2} A g_{u_1}(0)=(\lambda u_2,\lambda^2 u_1,0,0,\dots).
\]
Iterating this computation yields
\[
w_u(x)
=
(\lambda u_k,\lambda^2 u_{k-1},\dots,\lambda^k u_1,0,0,\dots),
\]
which proves (i).

Part (ii) now follows immediately from (i), because $w_u$ and $w_v$ are constant maps:
\[
d_\infty(w_u,w_v)
=
\sup_{x\in X} d_X\bigl(w_u(x),w_v(x)\bigr)
=
\max_{1\le j\le k}
\min\{1,\lambda^{k-j+1}\|u_j-v_j\|_E\}.
\]

For (iii), define two metrics on \(G^k\):
\[
\rho_k(u,v)
:=
\max_{1\le j\le k}\lambda^{k-j+1}\|u_j-v_j\|_E,
\qquad
\bar\rho_k(u,v)
:=
\max_{1\le j\le k}\min\{1,\lambda^{k-j+1}\|u_j-v_j\|_E\}.
\]
By part (ii), the parameterization $u\mapsto w_u$ is an isometric embedding of $(G^k,\bar\rho_k)$ into $(C(X,X),d_\infty)$, so
\[
N(W_k,d_\infty,\varepsilon)=N(G^k,\bar\rho_k,\varepsilon).
\]

For the upper bound, let $C_j$ be an $\varepsilon/\lambda^{k-j+1}$-cover of $G$ in $\|\cdot\|_E$.
Then the Cartesian product $C_1\times\cdots\times C_k$ is an $\varepsilon$-cover of $(G^k,\rho_k)$, because for every
$u=(u_1,\dots,u_k)$ one can choose $c_j\in C_j$ with
\[
\|u_j-c_j\|_E\le \frac{\varepsilon}{\lambda^{k-j+1}}
\]
for all $j$, and therefore
\[
\rho_k(u,c)\le \varepsilon,
\]
and hence \(\bar\rho_k(u,c)\le\varepsilon\).
Hence
\[
N(W_k,d_\infty,\varepsilon)
=
N(G^k,\bar\rho_k,\varepsilon)
\le
\prod_{j=1}^k
N \left(G,\|\cdot\|_E,\frac{\varepsilon}{\lambda^{k-j+1}}\right).
\]

For the lower bound, let $P_j$ be a maximal $2\varepsilon/\lambda^{k-j+1}$-packing of $G$ in $\|\cdot\|_E$.
Then for distinct points
\[
u=(u_1,\dots,u_k),\ v=(v_1,\dots,v_k)\in P_1\times\cdots\times P_k,
\]
there exists some $j$ with $u_j\neq v_j$, and for that $j$,
\[
\lambda^{k-j+1}\|u_j-v_j\|_E\ge 2\varepsilon.
\]
Therefore
\[
\rho_k(u,v)\ge 2\varepsilon,
\]
and since \(2\varepsilon<1\), also \(\bar\rho_k(u,v)\ge 2\varepsilon\).
Thus $P_1\times\cdots\times P_k$ is a $2\varepsilon$-packing of $(G^k,\bar\rho_k)$.
Any $\varepsilon$-cover must contain at least as many points as any $2\varepsilon$-packing, hence
\[
N(W_k,d_\infty,\varepsilon)
=
N(G^k,\bar\rho_k,\varepsilon)
\ge
\prod_{j=1}^k
M \left(G,\|\cdot\|_E,\frac{2\varepsilon}{\lambda^{k-j+1}}\right).
\]

Finally, since $W_k\subset B(2k+1,F)$ and external covering numbers are
monotone under inclusion,
\[
N(B(2k+1,F),d_\infty,\varepsilon)\ge N(W_k,d_\infty,\varepsilon).
\]
This completes the proof.
\end{proof}

\begin{corollary}[Super-exponential growth]
\label{cor:superexp}
Assume that \(N(G,\|\cdot\|_E,\delta)<\infty\) for all \(\delta>0\)
(automatic for compact \(G\)) and that there exist constants
$c_-,c_+,\delta_0>0$ such that, for all $0<\delta<\delta_0$,
\[
c_- \log(1/\delta)
\le
\log M(G,\|\cdot\|_E,\delta)
\le
\log N(G,\|\cdot\|_E,\delta)
\le
c_+ \log(1/\delta).
\]
Then for every fixed $\varepsilon$ with $0<\varepsilon<1/2$ and
$\varepsilon<\lambda\delta_0/2$, and for all $k\ge1$,
\[
C_1 k^2
\le
\log N(W_k,d_\infty,\varepsilon)
\le
C_2 k^2,
\qquad
C_1:=\frac{c_-\log\lambda}{2},
\quad
C_2:=c_+\bigl(\log\lambda-\log\varepsilon\bigr).
\]
Consequently,
\[
\log N(B(2k+1,F),d_\infty,\varepsilon)\ge C_1 k^2,
\]
so the word-ball covering number is at least \(\exp(\Omega(k^2))\).
\end{corollary}

\begin{proof}
By \Cref{cond:e3},
\[
\log N(W_k,d_\infty,\varepsilon)
\ge
\sum_{j=1}^k
\log M \left(G,\|\cdot\|_E,\frac{2\varepsilon}{\lambda^{k-j+1}}\right),
\]
and
\[
\log N(W_k,d_\infty,\varepsilon)
\le
\sum_{j=1}^k
\log N \left(G,\|\cdot\|_E,\frac{\varepsilon}{\lambda^{k-j+1}}\right).
\]
Since $0<\varepsilon<\lambda\delta_0/2$ and $\lambda>1$, every radius
$2\varepsilon/\lambda^{k-j+1}\le 2\varepsilon/\lambda$ and
$\varepsilon/\lambda^{k-j+1}$ lies below $\delta_0$; note that the $j=k$
radius does not shrink with $k$, so the smallness of $\varepsilon$ cannot
be traded for largeness of $k$.  Moreover $G\ne\emptyset$ by the packing
lower bound.  Hence the assumed entropy bounds apply to every factor, and
with
\[
\sum_{j=1}^k \log \left(\frac{\lambda^{k-j+1}}{a}\right)
=
(\log\lambda)\frac{k(k+1)}{2}+k\log(1/a)
\qquad(a>0)
\]
we obtain, using $2\varepsilon<1$ for the lower bound and $k+1\le 2k$,
$k\le k^2$ for the upper bound,
\[
\frac{c_-\log\lambda}{2}\,k^2
\le
\log N(W_k,d_\infty,\varepsilon)
\le
c_+\bigl(\log\lambda+\log(1/\varepsilon)\bigr)k^2 .
\]
\end{proof}

\begin{corollary}[Double-exponential growth]
\label{cor:doubleexp}
Assume that \(N(G,\|\cdot\|_E,\delta)<\infty\) for all \(\delta>0\) and
that there exist constants $p>0$, $c_-,c_+,\delta_0>0$ such that, for all $0<\delta<\delta_0$,
\[
c_- \delta^{-p}
\le
\log M(G,\|\cdot\|_E,\delta)
\le
\log N(G,\|\cdot\|_E,\delta)
\le
c_+ \delta^{-p}.
\]
Then for every fixed $\varepsilon$ with $0<\varepsilon<1/2$ and
$\varepsilon<\lambda\delta_0/2$, and for all $k\ge1$,
\[
C_1 \lambda^{pk}
\le
\log N(W_k,d_\infty,\varepsilon)
\le
C_2 \lambda^{pk},
\qquad
C_1:=c_-(2\varepsilon)^{-p},
\quad
C_2:=c_+\varepsilon^{-p}\frac{q}{q-1},
\quad q:=\lambda^p .
\]
Consequently,
\[
\log N(B(2k+1,F),d_\infty,\varepsilon)\ge C_1\lambda^{pk},
\]
i.e. the word-ball covering number is at least double-exponential in
\(k\).
\end{corollary}

\begin{proof}
Again by \Cref{cond:e3},
\[
\log N(W_k,d_\infty,\varepsilon)
\ge
\sum_{j=1}^k
\log M \left(G,\|\cdot\|_E,\frac{2\varepsilon}{\lambda^{k-j+1}}\right),
\]
and
\[
\log N(W_k,d_\infty,\varepsilon)
\le
\sum_{j=1}^k
\log N \left(G,\|\cdot\|_E,\frac{\varepsilon}{\lambda^{k-j+1}}\right).
\]
As in the proof of \cref{cor:superexp}, all radii lie below $\delta_0$ and
$G\ne\emptyset$, so the entropy bounds apply to every factor:
\[
c_-(2\varepsilon)^{-p}\sum_{r=1}^k q^{r}
\le
\log N(W_k,d_\infty,\varepsilon)
\le
c_+\varepsilon^{-p}\sum_{r=1}^k q^{r},
\qquad q=\lambda^p>1 .
\]
Since $q^k\le\sum_{r=1}^k q^r=q(q^k-1)/(q-1)\le\frac{q}{q-1}q^k$, this
proves the claim.
\end{proof}

\begin{example}[H\"older writers on a memory state space]
\label{ex:holder-memory}
Let $E=C([0,1]^d)$ with the sup norm, let $0<\alpha\le 1$, and define
\[
G
:=
\Bigl\{
u\in C^\alpha([0,1]^d):
\|u\|_\infty+[u]_{C^\alpha}\le 1
\Bigr\}.
\]
Set $X=\ell_\infty(E)$ and define $r$, $A$, and $g_u$ exactly as in \Cref{cond:e3}. %
Then there exist constants $c_-,c_+,\delta_0>0$ such that, for all $0<\delta<\delta_0$,
\[
c_- \delta^{-d/\alpha}
\le
\log M(G,\|\cdot\|_\infty,\delta)
\le
\log N(G,\|\cdot\|_\infty,\delta)
\le
c_+ \delta^{-d/\alpha}.
\]
Consequently, for every fixed $0<\varepsilon<\min\{1/2,\lambda\delta_0/2\}$,
\[
\log N(W_k,d_\infty,\varepsilon)
=
\Theta \bigl(\lambda^{(d/\alpha)k}\bigr).
\]
In particular,
\[
N(B(2k+1,F),d_\infty,\varepsilon)
\ge
\exp \Bigl(\Omega \bigl(\lambda^{(d/\alpha)k}\bigr)\Bigr),
\]
so the word-ball covering number is at least double-exponential in the depth parameter $k$.
\end{example}

\begin{proof}
The displayed entropy estimate is the standard metric entropy estimate for the unit H\"older ball in the sup norm.
Apply Corollary~\ref{cor:doubleexp} with $p=d/\alpha$.
\end{proof}

\section{The layerwise covering bound}
\label{sec:proof-envelope}

This appendix proves the layerwise covering bound of
\cref{sec:growth-conditions}, derives the profiles stated after it, and
records the elementary bound on the diameter of the reachable states that
accompanies it.  Throughout, \(F\subset C(\xdom,\xdom)\) consists of
\(\Lambda\)-Lipschitz maps with \(\Lambda\ge0\), and
\(S_m(\Lambda):=\sum_{i=0}^{m-1}\Lambda^i\) (\(S_0=0\), \(S_1=1\)) is the
geometric sum of \cref{prop:implementation}(b).  Covering numbers are
external (\cref{sec:setting}).

\subsection{Statement and proof}

\begin{theorem}[\cref{prop:envelope}, restated]
\label{prop:envelope-restated}
For every \(k\ge0\) and \(\eps>0\),
\begin{equation}
N\bigl(B(k,F),d_\infty,\eps\bigr)
\le
1+\sum_{m=1}^{k}N\bigl(F,d_\infty,\eps/S_m(\Lambda)\bigr)^{m}
\le
1+k\,N\bigl(F,d_\infty,\eps/S_k(\Lambda)\bigr)^{k},
\label{eq:envelope-restated}
\end{equation}
where the last inequality uses that \(S_m(\Lambda)\le S_k(\Lambda)\) for
\(m\le k\) (for \(k=0\) both sides equal \(1\)).  The centres of the covers
of \(F\) need not belong to \(F\) nor be Lipschitz; only the Lipschitz
constant of the maps in \(F\) is used, and \(F=\emptyset\) is allowed.
\end{theorem}

\begin{proof}
\(B(k,F)=\{\id\}\cup\bigcup_{m=1}^kF^{m}\), where \(F^m\) is the set of
compositions of exactly \(m\) maps of \(F\), and covering numbers are
subadditive under unions.  Fix \(1\le m\le k\) and let
\(\{c_1,\dots,c_M\}\subset C(\xdom,\xdom)\) be an
\(\eps/S_m(\Lambda)\)-cover of \(F\) in \(d_\infty\) with
\(M=N(F,d_\infty,\eps/S_m(\Lambda))\).  For \(f_m\circ\cdots\circ f_1\in F^m\)
choose indices with \(d_\infty(f_j,c_{i_j})\le\eps/S_m(\Lambda)\).  The
telescoping estimate in the proof of \cref{prop:implementation}(b), applied
with the abstract layers \(f_j\) and the implemented layers \(c_{i_j}\),
gives
\[
d_\infty\bigl(f_m\circ\cdots\circ f_1,\;c_{i_m}\circ\cdots\circ c_{i_1}\bigr)
\le
\frac{\eps}{S_m(\Lambda)}\sum_{j=1}^{m}\Lambda^{m-j}
=\eps ;
\]
that estimate uses only \(\lip(f_j)\le\Lambda\).  Hence the \(M^m\)
compositions \(c_{i_m}\circ\cdots\circ c_{i_1}\) form an \(\eps\)-cover of
\(F^m\), which proves the first inequality.  For the second, \(m\le k\)
implies \(S_m(\Lambda)\le S_k(\Lambda)\), hence
\(\eps/S_k(\Lambda)\le\eps/S_m(\Lambda)\) and
\(N(F,d_\infty,\eps/S_m(\Lambda))^m\le N(F,d_\infty,\eps/S_k(\Lambda))^k\)
because covering numbers are nonincreasing in the radius and at least \(1\).
\end{proof}

\subsection{Profiles from the layerwise bound}

\begin{corollary}[Profiles from the layerwise bound]
\label{cor:envelope-profiles}
Assume \(D_k(S)\le\overline D\) for all \(k\) with \(\overline D>0\), let
\(k\ge1\), and assume \(N(F,d_\infty,\eps)<\infty\) for all \(\eps>0\).
Since \(\mathsf V_k(S)\) is formed with internal \(d_S\)-covering numbers
while the layerwise bound bounds external \(d_\infty\)-covering numbers, the
transfer \(N(A,d_S,\eps)\le N(A,d_\infty,\eps/2)\) of
\cref{sec:proof-profiles} is used, which is the origin of the factors \(2\)
below.
\begin{enumerate}
\item[(a)]
(Parametric layers.)  If \(\log N(F,d_\infty,\eps)\le p\log(C/\eps)\) for all
\(0<\eps\le C\), with \(p\ge0\) and \(C\ge\overline D\), then for
\(0<\eps\le C\)
\[
\log N\bigl(B(k,F),d_\infty,\eps\bigr)
\le
\log(k+1)+kp\bigl[\log(C/\eps)+\log S_k(\Lambda)\bigr],
\]
and, with \(\Lambda_+:=\max\{1,\Lambda\}\) and \(S_k(\Lambda)\le k\Lambda_+^{k}\),
\begin{equation}
\mathsf V_k(S)
\le
\overline D\Bigl(
\sqrt{\log(k+1)}
+\sqrt{kp\log k}
+k\sqrt{p\log\Lambda_+}
+\sqrt{kp}\,\bigl(\sqrt{\log(2C/\overline D)}+\tfrac{\sqrt\pi}{2}\bigr)
\Bigr).
\label{eq:envelope-parametric}
\end{equation}
Thus \(\mathsf V_k(S)=O(\sqrt{kp\log(kC/\overline D)})\) if \(\Lambda\le1\)
and \(\mathsf V_k(S)=O(k\sqrt{p\log\Lambda})\) if \(\Lambda>1\).
\item[(b)]
(Nonparametric layers.)  If \(\log N(F,d_\infty,\eps)\le c\,\eps^{-q}\) for
all \(\eps>0\), with \(c\ge0\) and \(0<q<2\), then
\(\log N(B(k,F),d_\infty,\eps)\le\log(k+1)+kc\,S_k(\Lambda)^q\eps^{-q}\) for
all \(\eps>0\), and
\begin{equation}
\mathsf V_k(S)
\le
\overline D\sqrt{\log(k+1)}
+\sqrt{kc}\,\bigl(2S_k(\Lambda)\bigr)^{q/2}\,\frac{\overline D^{\,1-q/2}}{1-q/2}.
\label{eq:envelope-nonparametric}
\end{equation}
Thus \(\mathsf V_k(S)=O(k^{(1+q)/2})\) if \(\Lambda\le1\) and
\(\mathsf V_k(S)=O(k^{1/2}\Lambda^{qk/2})\) if \(\Lambda>1\).
\end{enumerate}
\end{corollary}

\begin{proof}
(a) For \(0<\eps\le C\) and \(k\ge1\) we have
\(\eps/S_k(\Lambda)\le\eps\le C\), so the hypothesis applies at radius
\(\eps/S_k(\Lambda)\) and \eqref{eq:envelope-restated} gives
\(\log N(B(k,F),\eps)\le\log(1+kN^k)\le\log(k+1)+k\log N\) with
\(N:=N(F,d_\infty,\eps/S_k(\Lambda))\), \(\log N\le p[\log(C/\eps)+\log S_k(\Lambda)]\);
here \(1+kN^k\le(k+1)N^k\) because \(N\ge1\).  Since \(\Lambda^i\le\Lambda_+^{k}\)
for \(i<k\), \(S_k(\Lambda)\le k\Lambda_+^{k}\), so
\(\log S_k(\Lambda)\le\log k+k\log\Lambda_+\).  Taking square roots with
\(\sqrt{a+b}\le\sqrt a+\sqrt b\),
\[
\sqrt{\log N(B(k,F),\eps)}
\le
\sqrt{\log(k+1)}+\sqrt{kp\log k}+k\sqrt{p\log\Lambda_+}
+\sqrt{kp}\sqrt{\log(C/\eps)},
\]
for every \(0<\eps\le C\).  Since
\(N(B(k,F),d_S,\eps)\le N(B(k,F),d_\infty,\eps/2)\) and \(D_k(S)\le\overline D\),
\(\mathsf V_k(S)\) is bounded by the integral of the right-hand side at
radius \(\eps/2\) over \((0,\overline D]\), where
\(\int_0^{\overline D}\sqrt{\log(2C/\eps)}\,\dd\eps
\le\overline D\bigl(\sqrt{\log(2C/\overline D)}+\sqrt\pi/2\bigr)\) by
\cref{lem:log-split} applied to
\(\log(2C/\eps)=\log(2C/\overline D)+\log(\overline D/\eps)\); this gives
\eqref{eq:envelope-parametric}.

(b) Now \(\log N(B(k,F),\eps)\le\log(k+1)+kc\,(S_k(\Lambda)/\eps)^q\) for
all \(\eps>0\); at radius \(\eps/2\) the second term is
\(kc(2S_k(\Lambda))^q\eps^{-q}\), and
\(\int_0^{\overline D}\eps^{-q/2}\,\dd\eps=\overline D^{\,1-q/2}/(1-q/2)\)
because \(q<2\).  The asymptotics follow from \(S_k(\Lambda)\le k\) for
\(\Lambda\le1\) and \(S_k(\Lambda)\le\Lambda^{k}/(\Lambda-1)\) for
\(\Lambda>1\).
\end{proof}

\subsection{Sharpness}

The memory-preserving construction of \cref{cond:e3} attains both
layerwise bounds.  Its layer class \(F=\{r,A\}\cup\{g_u:u\in G\}\) consists of
\(\lambda\)-Lipschitz maps for the bounded metric \(d_X\) (the reset and the
writers are \(1\)-Lipschitz, and
\(\min\{1,\lambda a\}\le\lambda\min\{1,a\}\) for \(\lambda\ge1\)), and
\(N(F,d_\infty,\eps)\le2+N(G,\|\cdot\|_E,\eps)\) for \(\eps<1\).  If \(G\) has
finite-dimensional entropy, \(\log N(G,\delta)\asymp\log(1/\delta)\), then
case (a) gives \(\log N(B(k,F),\eps)\lesssim k^2\), and \cref{cor:superexp}
gives the matching lower bound \(\gtrsim k^2\) (its words \(W_k\) lie in
\(B(2k+1,F)\), and external covering numbers are monotone under inclusion).
If \(\log N(G,\delta)\asymp\delta^{-p}\), case (b) gives
\(\log N(B(k,F),\eps)\lesssim k\lambda^{pk}\), and \cref{cor:doubleexp} gives
\(\gtrsim\lambda^{pk}\).  Thus the layerwise bound is attained up to polynomial
factors, and no better general bound in terms of \(N(F,\cdot)\) and
\(\Lambda\) alone is possible.

\subsection{Diameter of the reachable states}

On a bounded state space \(D_k(S)\le\diam(\xdom)\) for every \(k\).  On an
unbounded space the diameter can grow with the depth, and the growth is
again governed by \(\Lambda\).

\begin{lemma}[Reachable radius]
\label{lem:reachable-radius}
Let \(x_0\in\xdom\), \(R\ge0\), \(c\ge0\), and suppose \(d(f(x_0),x_0)\le c\)
for all \(f\in F\).  Then, with \(\Lambda_+:=\max\{1,\Lambda\}\), every
\(g\in B(k,F)\) maps the ball \(B(x_0,R)\) into
\(B\bigl(x_0,\Lambda_+^{k}R+c\,S_k(\Lambda)\bigr)\).  Consequently, if
\(S\subset B(x_0,R)\), then
\(D_k(S)\le2\bigl(\Lambda_+^{k}R+c\,S_k(\Lambda)\bigr)\): bounded in \(k\) if
\(\Lambda<1\), at most linear if \(\Lambda=1\), and at most exponential if
\(\Lambda>1\).
\end{lemma}

\begin{proof}
For \(f\in F\) and \(d(x,x_0)\le\rho\),
\(d(f(x),x_0)\le d(f(x),f(x_0))+d(f(x_0),x_0)\le\Lambda\rho+c\).  Iterating
\(m\le k\) times from \(\rho_0=R\) gives
\(\rho_m\le\Lambda^mR+c\sum_{i<m}\Lambda^i\), which is at most
\(\Lambda_+^{k}R+cS_k(\Lambda)\) since \(\Lambda^m\le\Lambda_+^k\) and
\(S_m\le S_k\).  The diameter bound follows from the triangle inequality.
\end{proof}

The exponential factor \(\Lambda^kR\) is the product of the layerwise
Lipschitz constants that appears in norm-based generalization bounds for
fully connected networks \citep{Neyshabur2015,Bartlett2017}.  In the present
bound it enters only through the diameter, that is, through the integration
range of the entropy integral, and only when the state space is unbounded;
bounded activations, normalization layers, or a projection onto a bounded
set remove it, after which \cref{cor:envelope-profiles} applies with
\(\overline D=\diam(\xdom)\).

\section{Variance profiles}
\label{sec:proof-profiles}

The following proposition converts the fixed-scale growth mechanisms of
\cref{sec:growth-conditions}, together with a bound on the diameter, into
the depth profiles of the entropy integral listed at the end of
\cref{sec:growth-conditions};
the list after the proposition records which case applies to each
mechanism and to the layerwise bound.

\subsection{Statement}

\begin{proposition}[Variance profiles from growth and diameter]
\label{prop:profiles}
Let \(F\subset C(\xdom,\xdom)\), write \(N_k(\eps):=N(B(k,F),d_\infty,\eps)\),
and let \(\overline D\ge0\) (\(\overline D>0\) in (ii)), \(D_1>0\),
\(C_0\ge1\), and \(D\ge0\) be constants.
\begin{enumerate}
\item[(i)]
(Saturation.)  If \(D_k(S)\le\overline D\) and
\(N_k(\eps)\le N_\infty(\eps)\) for all \(k\) and \(\eps>0\), with
\(\mathsf V_\infty:=\int_0^{\overline D}\sqrt{\log N_\infty(\eps)}\,\dd\eps<\infty\),
then \(\mathsf V_k(S)\le\mathsf V_\infty\) for all \(k\).
\item[(ii)]
(Polynomial growth, bounded diameter.)  If \(D_k(S)\le\overline D\) and
\(N_k(\eps)\le C_0(1+k/\eps)^D\) for all \(k\ge1\) and \(\eps>0\), then
\(\mathsf V_k(S)
\le
\overline D\sqrt{D}\bigl(\sqrt{\log(1+k/\overline D)}+\tfrac{\sqrt\pi}{2}\bigr)
+\overline D\sqrt{\log C_0}
=O(\sqrt{D\log k})\).
\item[(iii)]
(Exponential growth, bounded diameter.)  If \(D_k(S)\le\overline D\) and
\(\log N_k(\eps)\le \alpha k+\psi(\eps)\) with
\(\Psi:=\int_0^{\overline D}\sqrt{\psi(\eps)}\,\dd\eps<\infty\), then
\(\mathsf V_k(S)\le\overline D\sqrt{\alpha k}+\Psi=O(\sqrt k)\); in
particular, if \(|F|=r\ge2\), then \(N_k(\eps)\le r^{k+1}\) and
\(\mathsf V_k(S)\le\overline D\sqrt{(k+1)\log r}\).
\item[(iv)]
(Polynomial growth, linearly growing diameter.)  If \(D_k(S)\le D_1 k\) and
\(N_k(\eps)\le C_0(1+k/\eps)^D\), then
\(\mathsf V_k(S)
\le
D_1k\bigl(\sqrt{D}(\sqrt{\log(1+1/D_1)}+\tfrac{\sqrt\pi}{2})+\sqrt{\log C_0}\bigr)
=O(k\sqrt D)\).
\end{enumerate}
Consequently, \(\var(k,n)\) is of order \(n^{-1/2}\), \(\sqrt{D\log k/n}\),
\(\sqrt{k/n}\), and \(k\sqrt{D/n}\) in the four cases, respectively.
\end{proposition}

\subsection{Profiles by mechanism}

In each item, the first entry is the fixed-scale entropy
\(\log N(B(k,F),d_\infty,\eps)\), the second the entropy integral
\(\mathsf V_k(S)\), the third the estimation term \(\var(k,n)\) with its
letter (L for logarithmic, P for polynomial), and the last the case of
\cref{prop:profiles} or of \cref{cor:envelope-profiles} and the matching
lower bound where one is available.  A bounded diameter is assumed except
in the last item.
\begin{itemize}
\item P1, P1' (compact equicontinuous dynamics; contraction to a bounded
  set): \(O(1)\); \(O(1)\); \(n^{-1/2}\) (L); case (i), saturated.
\item P2 (nilpotent control), compact state space: \(D\log(1+k/\eps)\);
  \(O(\sqrt{D\log k})\); \(\sqrt{\log k/n}\) (L); case (ii), matched by
  \cref{cor:matching}(ii).
\item E1', E2 (\(r\) separated generators): \(\asymp k\log r\);
  \(O(\sqrt{k\log r})\); \(\sqrt{k/n}\) (P, \(\varp=1\)); case (iii),
  matched by \cref{cor:matching}(i).
\item Layerwise bound, parametric layers, \(\Lambda\le1\): \(kp\log(kC/\eps)\);
  \(O(\sqrt{kp\log k})\); \(\sqrt{k\log k/n}\) (P, \(\varp=1\));
  \cref{cor:envelope-profiles}(a), the order of the VC-dimension bounds of
  \citet{Bartlett2019}.
\item Layerwise bound, parametric layers, \(\Lambda>1\):
  \(pk(k\log\Lambda+\log(C/\eps))\); \(O(k\sqrt{p\log\Lambda})\);
  \(k\sqrt{p/n}\) (P, \(\varp=2\)); \cref{cor:envelope-profiles}(a),
  attained by E3 (\cref{cor:superexp}).
\item Layerwise bound, nonparametric layers with entropy \(c\eps^{-q}\),
  \(\Lambda>1\): \(kc(S_k(\Lambda)/\eps)^q\); \(O(\sqrt k\,\Lambda^{qk/2})\);
  exponential; \cref{cor:envelope-profiles}(b), attained by E3
  (\cref{cor:doubleexp}).
\item P2, non-compact state space (\(D_k\le2L_\alpha R_Sk\)):
  \(D\log(1+k/\eps)\); \(O(k\sqrt D)\); \(k\sqrt{D/n}\) (P, \(\varp=2\));
  case (iv).
\end{itemize}

\subsection{Proof}

We now prove the elementary estimates behind \cref{prop:profiles}.  Throughout,
\(N_k(\eps):=N(B(k,F),d_\infty,\eps)\), and we use that
\(N(B(k,F),d_S,\eps)\le N_k(\eps)\) because \(d_S\le d_\infty\)
(\cref{sec:growth-conditions}); this holds for external and for internal
covering numbers alike, since a \(d_\infty\)-cover of \(B(k,F)\) is a
\(d_S\)-cover with the same centres.  If the entropy integral of
\cref{thm:hidden-decomp} is formed with internal covering numbers while the
growth conditions of \cref{sec:growth-conditions-proof} bound external ones,
use \(N^{\rm int}(A,d_S,\eps)\le N(A,d_S,\eps/2)\le N(A,d_\infty,\eps/2)\)
instead; this replaces \(C_0\) by \(2^DC_0\) in cases (ii) and (iv),
\(\psi(\eps)\) by \(\psi(\eps/2)\) in case (iii), and \(N_\infty(\eps)\) by
\(N_\infty(\eps/2)\) in case (i), and changes nothing else.

\begin{lemma}
\label{lem:log-split}
For \(\overline D>0\), \(k\ge0\), and \(0<\eps\le\overline D\),
\[
\log\Bigl(1+\frac{k}{\eps}\Bigr)
\le
\log\Bigl(1+\frac{k}{\overline D}\Bigr)+\log\frac{\overline D}{\eps},
\qquad
\int_0^{\overline D}\sqrt{\log\frac{\overline D}{\eps}}\,\dd\eps
=\overline D\,\Gamma(3/2)=\frac{\sqrt\pi}{2}\,\overline D .
\]
\end{lemma}

\begin{proof}
For \(0<\eps\le\overline D\),
\((1+k/\overline D)(\overline D/\eps)=\overline D/\eps+k/\eps\ge 1+k/\eps\),
which gives the first inequality after taking logarithms.  For the integral,
substitute \(\eps=\overline D t\) and then \(t=e^{-s}\):
\[
\int_0^{\overline D}\sqrt{\log(\overline D/\eps)}\,\dd\eps
=\overline D\int_0^1\sqrt{\log(1/t)}\,\dd t
=\overline D\int_0^\infty s^{1/2}e^{-s}\,\dd s
=\overline D\,\Gamma(3/2).
\]
\end{proof}

\begin{proof}[Proof of \cref{prop:profiles}]
(i) Since the integrand and the upper limit are both dominated,
\(\mathsf V_k(S)\le\int_0^{\overline D}\sqrt{\log N_\infty(\eps)}\,\dd\eps=\mathsf V_\infty\).

(ii) Using \(\sqrt{a+b+c}\le\sqrt a+\sqrt b+\sqrt c\) and \cref{lem:log-split},
for \(0<\eps\le\overline D\),
\[
\sqrt{\log N_k(\eps)}
\le
\sqrt{\log C_0}
+\sqrt{D\log(1+k/\overline D)}
+\sqrt{D\log(\overline D/\eps)} .
\]
Integrating over \((0,\overline D]\) and applying the identity in
\cref{lem:log-split} to the last term gives the claim.

(iii) Here \(\sqrt{\log N_k(\eps)}\le\sqrt{\alpha k}+\sqrt{\psi(\eps)}\), and
integration over \((0,\overline D]\) gives
\(\mathsf V_k(S)\le\overline D\sqrt{\alpha k}+\Psi\).  If \(|F|=r\ge2\), then
\(B(k,F)\) has at most \(\sum_{m=0}^k r^m\le r^{k+1}\) elements (this uses
\(r\ge2\)), so
\(N_k(\eps)\le r^{k+1}\) at every scale and one may take
\(\alpha=\log r\) with \(\psi\equiv\log r\), or directly
\(\log N_k(\eps)\le(k+1)\log r\), which yields
\(\mathsf V_k(S)\le\overline D\sqrt{(k+1)\log r}\).

(iv) Apply \cref{lem:log-split} with \(\overline D\) replaced by \(D_1k\):
for \(0<\eps\le D_1k\),
\(\log(1+k/\eps)\le\log(1+1/D_1)+\log(D_1k/\eps)\), hence
\begin{align*}
\mathsf V_k(S)
&\le
\int_0^{D_1k}\Bigl(\sqrt{\log C_0}+\sqrt{D\log(1+1/D_1)}+\sqrt{D\log(D_1k/\eps)}\Bigr)\dd\eps
\\
&=
D_1k\Bigl(\sqrt{\log C_0}+\sqrt{D\log(1+1/D_1)}+\sqrt D\,\tfrac{\sqrt\pi}{2}\Bigr).
\end{align*}
The statements about \(\var(k,n)\) follow by multiplying with
\(12A_HL/\sqrt n\).
\end{proof}

\subsection{Diameter bounds by mechanism}

Under \cref{cond:p1} the state space is compact, so
\(D_k(S)\le\diam_\infty(B(k,F))\le\diam(\xdom)\).  Under
\cref{cond:p1-ucont} with the bounded-metric convention
\(d_b=\min\{1,d\}\), \(D_k(S)\le1\); without that convention, all words of
length at least \(L\) take values in the bounded absorbing set \(K\), so the
sub-ball of such words has diameter at most \(\diam(K)\).  Under
\cref{cond:p2-nilp}, the proof gives \(B(k,F)\subset\alpha(B_H(e,kR_S))\)
and the orbit map is \(L_\alpha\)-Lipschitz, so
\(D_k(S)\le d_\infty\)-\(\diam\,\alpha(B_H(e,kR_S))\le 2L_\alpha R_Sk\);
on a compact state space this is improved to \(\diam(\xdom)\).  Under
\cref{cond:e2-pingpong} the state space is compact, and the construction of
\cref{cond:e3} uses a metric bounded by \(1\).  Under
\cref{cond:e1-free-iso} with the standard word metric, \(d_\infty\) may be
infinite; the bi-invariant variant \cref{ex:e1-free-bounded} has \(D_k\le k\), and
after passing to \(d_b\) the diameter is at most \(1\).

\section{The four regimes and the balancing depths}
\label{sec:proof.tradeoff}

This appendix defines the four regimes of \cref{sec:optdepth}, states
their balancing depths and balanced rates, and derives them.

\subsection{Two laws and four regimes}

The main text names two of the four combinations of bias and variance laws; here we treat all four.  The approximation error
decays with depth $k$ either exponentially or polynomially,
\begin{align*}
\bias(k) = \exp(\Theta(-\biase k))\quad\text{or}\quad
\Theta(k^{-\biasp}),
\end{align*}
with parameters $\biase,\biasp>0$; the letters E and P refer to these two
cases.  The estimation term follows one of the two profiles produced by
\cref{prop:profiles}, root-logarithmic or root-polynomial growth,
\begin{align*}
\var(k,n) \lesssim
\sqrt{\log (k)/n}
\quad\text{or}\quad
\sqrt{k^\varp/n},
\end{align*}
with parameter $\varp>0$; the letters L and P refer to these two cases.  The
root-logarithmic profile is case (ii) of \cref{prop:profiles}, and it also
covers the saturated case (i), for which the true dependence is constant;
case (iii) is the root-polynomial profile with \(\varp=1\), and case (iv)
with \(\varp=2\).  Combining the two bias laws with the two variance profiles
gives rise to the four typical regimes EL, EP, PL, and PP listed in
\eqref{eq:tradeoff-table}: the first letter describes the bias and the second the
variance.  They are representative diagnostic cases rather than an
exhaustive list of all possible depth dependences; for instance, the
sub-exponential bias decay \(\exp(-\Theta(\sqrt k))\) of Picard-type neural
operators (\cref{sec:examples-regime}) combined with a saturated or
root-logarithmic profile is called the SubEL regime there.

The balancing depths and balanced rates are the following, with the
derivations below:
\begin{equation}
\label{eq:tradeoff-table}
\begin{aligned}
&\text{EL}\ (\bias=e^{-\biase k},\ \var=\sqrt{\log k/n}):
&&\kopt=\tfrac{1}{2\biase}(\log n-\log\log\log n),
&&\gen(\kopt,n)\asymp\sqrt{\tfrac{\log\log n}{n}};\\
&\text{EP}\ (\bias=e^{-\biase k},\ \var=\sqrt{k^\varp/n}):
&&\kopt=\tfrac{1}{2\biase}(\log n-\varp\log\log n),
&&\gen(\kopt,n)\asymp\sqrt{\tfrac{(\log n)^{\varp}}{(2\biase)^{\varp}n}};\\
&\text{PL}\ (\bias=k^{-\biasp},\ \var=\sqrt{\log k/n}):
&&\kopt\sim\Bigl(\tfrac{2\biasp n}{\log(2\biasp n)}\Bigr)^{\frac{1}{2\biasp}},
&&\gen(\kopt,n)\asymp\sqrt{\tfrac{\log n}{2\biasp n}};\\
&\text{PP}\ (\bias=k^{-\biasp},\ \var=\sqrt{k^\varp/n}):
&&\kopt\asymp n^{\frac{1}{2\biasp+\varp}},
&&\gen(\kopt,n)\asymp n^{-\frac{\biasp}{2\biasp+\varp}}.
\end{aligned}
\end{equation}
Lower-order terms in \(\kopt\) are omitted in the EL and EP rows.

The worked examples are placed in these regimes in the summary list of
\cref{sec:examples-regime}.

\subsection{Width and other one-layer parameters}

\begin{remark}[Width and other one-layer parameters]
\label{rem:width}
The display \eqref{eq:tradeoff-table} fixes the hidden-layer class \(F\) and
the output-layer class \(H\) and varies only the depth.  Width, and more generally any
parameter of a single layer, is not absent from the framework: if \(F_w\)
denotes the admissible hidden layers at width \(w\), then \(w\) enters the
one-layer covering numbers \(N(F_w,d_\infty,\eps)\), hence those of
\(B(k,F_w)\) through \cref{prop:envelope}, where \(p\asymp w^2\) for a
fully connected layer, and possibly the output-layer class \(H_w\), the
approximation term, and the implementation error
(\cref{sec:example-implementation} quantifies the last route).  Writing \(\calH_{k,w}=H_w\circ B(k,F_w)\),
the balancing depth \(\kopt(n,w)\) is obtained by minimizing the full bound
over \(k\) for each fixed \(w\); the display does not assert that it is
independent of \(w\), and its dependence on \(w\) need not be monotone
because width can change both the bias and the variance.  A joint
optimization over depth and width, and a resource-matched comparison with
wide shallow networks, are outside the present analysis.
\end{remark}

\subsection{The balancing principle}

Throughout, we minimize the upper bound
\[
\gen(k,n)\lesssim\bias(k)+\var(k,n)
\]
over the depth \(k\), treated as a positive real number and rounded to the
nearest integer at the end.  Since \(\bias(k)\) is nonincreasing and
\(\var(k,n)\) is nondecreasing in \(k\), the minimizer is located where the
two terms are of the same order, and the balanced value is then of the order
of either term.  The following lemma makes this precise.

\begin{lemma}[Balancing principle]
\label{lem:balancing}
Let \(\bias\ge0\) be nonincreasing and \(\var(\cdot,n)\ge0\) be nondecreasing
on a set \(\mathcal K\) of depths, and let \(\gen(k,n):=\bias(k)+\var(k,n)\).
If \(k_0\in\mathcal K\) balances the two terms, \(\bias(k_0)=\var(k_0,n)\),
then \(\gen(k,n)\ge\bias(k_0)\) for every \(k\in\mathcal K\), while
\(\gen(k_0,n)=2\bias(k_0)\); hence \(k_0\) minimizes the bound over
\(\mathcal K\) up to the factor \(2\).
\end{lemma}

\begin{proof}
For \(k\le k_0\) we have \(\gen(k,n)\ge\bias(k)\ge\bias(k_0)\), and for
\(k\ge k_0\) we have \(\gen(k,n)\ge\var(k,n)\ge\var(k_0,n)=\bias(k_0)\).
\end{proof}

In the PP regime the balancing equation is solved exactly, so
\cref{lem:balancing} applies verbatim.  In the other three regimes the
displayed \(\kopt\) balances the two terms only asymptotically; the balanced
rates are nevertheless two-sided, of the form \(\Theta(\cdot)\), with
explicit constants.  For all sufficiently large \(n\), the EP rate lies
between \(1\) and \(1+(2\alpha)^{-\gamma/2}\) times
\(n^{-1/2}(\log n)^{\gamma/2}\), the EL rate between \(1\) and
\(1+\sqrt{1+|\log 2\alpha|}\) times \(\sqrt{\log\log n/n}\), and the PL rate
between \(\sqrt{1/(4\beta)}\) and \(2\sqrt{(1+|\log 2\beta|)/(2\beta)}\)
times \(\sqrt{\log n/n}\).  If the two terms do not cross, the minimizer
lies at a boundary of \(\mathcal K\); in the four regimes below they cross
for all large \(n\).

\subsection{Derivations by regime}

\subsubsection{EP: exponentially decaying bias, polynomially growing variance}

Let \(\bias(k)=e^{-\alpha k}\) and \(\var(k,n)=n^{-1/2}k^{\gamma/2}\) with
\(\alpha,\gamma>0\).  Taking logarithms, the two terms are of the same order
exactly when
\[
\alpha k=\tfrac12\log n-\tfrac{\gamma}{2}\log k+O(1).
\]
As \(n\to\infty\) the solution satisfies \(\log k=\log\log n+O(1)\), because
\(\log k\ll\log n\); substituting this into the right-hand side gives
\[
\kopt=\frac{1}{2\alpha}\bigl(\log n-\gamma\log\log n+O(1)\bigr).
\]
Evaluating either term at \(\kopt\) yields the balanced rate
\[
\gen(\kopt,n)\asymp n^{-1/2}(\log n)^{\gamma/2},
\]
more precisely \((2\alpha)^{-\gamma/2}n^{-1/2}(\log n)^{\gamma/2}\) up to a
factor bounded above and below by positive constants.

\subsubsection{EL: exponentially decaying bias, logarithmically growing variance}

Let \(\bias(k)=e^{-\alpha k}\) and \(\var(k,n)=\sqrt{\log k/n}\) with
\(\alpha>0\).  The two terms are of the same order exactly when
\[
\alpha k=\tfrac12\log n-\tfrac12\log\log k+O(1).
\]
Since \(\log k=\log\log n+O(1)\) at the solution, and hence
\(\log\log k=\log\log\log n+O(1)\), we obtain
\[
\kopt=\frac{1}{2\alpha}\bigl(\log n-\log\log\log n+O(1)\bigr),
\qquad
\gen(\kopt,n)\asymp\sqrt{\frac{\log\log n}{n}} .
\]

\subsubsection{PP: polynomially decaying bias, polynomially growing variance}

Let \(\bias(k)=k^{-\beta}\) and \(\var(k,n)=n^{-1/2}k^{\gamma/2}\) with
\(\beta,\gamma>0\).  The two terms are equal exactly when
\(k^{\beta+\gamma/2}=n^{1/2}\), that is, at
\[
\kopt=n^{1/(2\beta+\gamma)},
\qquad\text{where}\qquad
\gen(\kopt,n)\asymp n^{-\beta/(2\beta+\gamma)} .
\]
Here the balancing equation is solved exactly, so \cref{lem:balancing}
shows that \(\kopt\) minimizes the bound up to the factor \(2\).

\subsubsection{PL: polynomially decaying bias, logarithmically growing variance}

Let \(\bias(k)=k^{-\beta}\) and \(\var(k,n)=\sqrt{\log k/n}\) with
\(\beta>0\).  Squaring both terms, they are of the same order exactly when
\(k^{2\beta}\log k\asymp n\).  The equation \(k^{2\beta}\log k=n\) is solved
by \(k=\exp(W(2\beta n)/(2\beta))\), where \(W\) is the Lambert function.
The Lambert function can be avoided by taking
\[
\kopt:=\Bigl(\frac{2\beta n}{\log(2\beta n)}\Bigr)^{1/(2\beta)},
\qquad\text{for which}\qquad
{\kopt}^{2\beta}\log\kopt=n\Bigl(1-\frac{\log L}{L}\Bigr),
\quad L:=\log(2\beta n),
\]
so that \({\kopt}^{2\beta}\log\kopt/n\to1\) as \(n\to\infty\).  For this
choice, \(\bias(\kopt)=\sqrt{L/(2\beta n)}\) and, for \(L\ge1\),
\(\var(\kopt,n)=\sqrt{(L-\log L)/(2\beta n)}\le\sqrt{L/(2\beta n)}\), so
that
\[
\gen(\kopt,n)\asymp\sqrt{\frac{\log n}{2\beta n}} .
\]

\paragraph{Ordering of the EP and PL rates.}
The EP and PL rates in \eqref{eq:tradeoff-table} are not uniformly ordered without
specifying \(\gamma\): EP gives \(n^{-1/2}(\log n)^{\gamma/2}\), whereas PL
gives \(n^{-1/2}(\log n)^{1/2}\), up to constants.  Thus EP is no worse
than PL for \(\gamma\le1\), while PL is no worse than EP for \(\gamma>1\).

\section{Worked examples: teacher--student models, neural operators, and the implementation layer}
\label{sec:examples-regime}

This appendix opens the series of worked examples.  Each example
instantiates the framework in the same order: the setting (state space,
hidden-layer class, output layer), the variance profile obtained from a
growth mechanism of \cref{sec:growth-conditions-proof}, the bias against a
target class, the balancing depth, and, where available, a computed
illustration, the implementation error, and the relation to the
literature.  The present appendix treats the two examples that need no
new construction, the contractive teacher--student model and Picard-type
neural operators, and then the implementation layer of \cref{thm:bv}, which
is used by all examples.  The deep ReLU, chain-of-thought, and solver
examples follow in \cref{sec:app-relu,sec:app-cot,sec:app-ode}.

\paragraph{Summary of the worked examples.}
For each example we record the variance profile, the bias, and the
balancing depth; ``proved'' means derived within the framework under the
stated hypotheses, ``known'' means an existing architecture-specific rate
inserted into the framework, and ``assumed'' means an assumption on the
target.
\begin{itemize}
\item Contractive teacher--student model (\cref{sec:examples-regime}):
  saturated profile (proved, P1/P1'); bias \(D\lambda^k\) (proved); EL,
  \(\kopt\asymp\log n/(2\log\lambda^{-1})\).
\item Fixed-point refinement, iterative denoising (\cref{sec:app-ode}):
  saturated profile with explicit constants (proved); bias
  \((1-h\mu)^k\) (proved); EL, \(\kopt\le\log n/(2h\mu)\).
\item Fixed-horizon integration, neural ODE, diffusion sampler
  (\cref{sec:app-ode}): \(\sqrt{\log k}\) profile with explicit constants
  (proved); bias \(C_ET^2k^{-p}\) (proved); PL, \(\kopt\asymp n^{1/(2p)}\).
\item Chain of thought, append-only scratchpad (\cref{sec:app-cot}):
  saturated profile with explicit constants (proved); bias \(\theta^k\)
  (proved); EL, \(\kopt\asymp\log n/(2\log\theta^{-1})\).
\item Chain of thought, branching (\cref{sec:app-cot}): \(\sqrt{\log k}\)
  profile on every sample (proved), \(\sqrt k\) in \(d_\infty\); bias \(0\)
  for \(k\ge k_0\) (proved) or \(e^{-\biase k}\) (assumed); \(\kopt=k_0\),
  or EL.
\item Picard-type neural operator (\cref{sec:examples-regime}): saturated
  profile (proved, P1'); bias \(e^{-c\sqrt k}\) (known); SubEL.
\item Deep ReLU network (\cref{sec:app-relu}): saturated profile for
  contractive or non-expanding layers on a bounded state space (proved),
  layerwise bound profile \(k\sqrt p\) for expanding layers (proved) with the
  \(\sqrt k\) lower bound realized by expand-and-reset layers (proved);
  bias \(\Lambda^k\) for a contractive teacher (proved), of Jackson type
  \(k^{-s/d}\) for H\"older or Besov targets (known); EL, respectively PP.
\end{itemize}

\subsection{Contractive teacher--student models}

\paragraph{Setting.}
A teacher-student setting refers to the scenario where the hypothesis class (student) and the target class (teacher) share the same compositional structure.
Let the depth-$k$ student class be $\calH_k := H \circ B(k,F)$ and take the target class to be the uniform closure of the infinitely deep compositional class,
$\Con := \calH_\infty := H \circ \overline{\bigcup_{m\ge 0}B(m,F)}^{\,d_\infty}.$
Assume the input space $\xdom$ is bounded with $\diam(\xdom)=D<\infty$; the output layer $H$ is $1$-Lipschitz; and the intermediate-layer semigroup $F\subset C(\xdom,\xdom)$ is \emph{contractive}, i.e., there exists $\lambda\in(0,1)$ such that $\lip(f)\le \lambda$ for all $f\in F$. A teacher of depth less than \(k\) is represented exactly by \(\calH_k\).  For a finite-depth teacher of depth at least \(k\), write \(c=h\circ u\circ v\) with $h\in H$, \(u\in F^k\), and \(v\in B(\ell,F)\).  Truncating the tail yields a depth-$k$ student $h\circ u\in\calH_k$ that approximates $c$ with error
\begin{align*}
\|h\circ u\circ v - h\circ u\|_\infty
&\le \lip(h)\lip(u) d_\infty(v,\mathrm{id})
\le 1\cdot \lambda^k \cdot D
= D\lambda^k,    
\end{align*}
since $\lip(u)\le \lambda^k$ and $d_\infty(v,\mathrm{id})\le D$. Passing to the uniform closure gives the same bound for teachers in \(\calH_\infty\). Hence the approximation error decays exponentially:
\[
\sup_{c\in\calH_\infty}\inf_{h\in\calH_k} \|c-h\|_\infty \le D\lambda^k .
\]
Moreover, under the compact-domain P1 condition, or under the non-compact
absorbing-set version P1', the contractive transition family has saturated
word-ball entropy: the hidden contribution to the variance upper bound is
independent of \(k\) at fixed scale, and in the compact/totally bounded case
the whole Dudley integral is depth-independent (case (i) of
\cref{prop:profiles}).  Thus this example is at least as favorable as the EL
proxy with root-logarithmic variance growth.

\subsection{Picard-type neural operators}

\paragraph{Setting and bias.}
\citet{furuya2025neuraloperator}. 
\citet{furuya2025neuraloperator} 
investigated the approximation error of Neural Operators (NOs) that learn the solution operator of nonlinear parabolic PDEs on a bounded domain $\xdom\subset\RR^d$.
By aligning a single hidden layer of the NO with one step of the Picard iteration for the PDE, a textbook iterative argument for solving differential equations, they show that the approximation error decays at a \emph{sub-exponential} rate of the form \(O(\exp(-c\sqrt{k}))\) in the network depth \(k\), for a problem-dependent constant \(c>0\).
Since Picard iteration is \emph{contractive} and generated by a single operator, the resulting transition dynamics are naturally treated by P1 on compact state spaces, or by P1' when the iteration has a compact absorbing set.  In either case, the transition word-ball entropy saturates rather than merely growing polynomially.
The function space for approximation is a \emph{mixed Lebesgue space} $L_t^{r}L_x^{s}$, 
which we can regard as the target class $\Con$.
A finite-sample generalization analysis of this Picard-type setting, including finite-observation reconstruction errors and long-time rollout, is developed by \citet{taniguchi2026picard}.

This NO setting can be viewed as a concrete instance of the teacher-student framework discussed earlier: both the hypothesis class and the target class share a hierarchical (iterative) structure, enabling efficient depth-driven approximation. It thus exemplifies the \emph{SubEL regime} of \cref{sec:optdepth}, sub-exponential bias decay with a saturated or root-logarithmic variance profile, in which depth is particularly advantageous. Together with the previous example, it suggests that depth is most effective when the underlying target structure, namely the target class $\Con$, is hierarchical, as in deep compositions or differential equations.

\subsection{The implementation layer}
\label{sec:example-implementation}
\label{sec:proof-implementation}

The implementation map in \cref{thm:bv} is the identity when the abstract class
is itself the implemented network class, and then \(\errimp(k)=0\).  The
distinction becomes useful when a mathematically convenient transition model
is only approximated by a finite network, for instance when \(\calH_k\) is a
class of polynomials or of smooth flows and \(\Himp\) is a finite-width ReLU
family.  The following statement gives three explicit implementation maps.

\begin{proposition}[Implementation with controlled uniform error]
\label{prop:implementation}
\begin{enumerate}
\item[(a)]
(Net-based.)  Let \(\calH_k\) be totally bounded in the uniform norm on
\(\xdom\), and let \(\calA\) be a class of functions that is uniformly dense
on \(\calH_k\); for example, on a compact \(\xdom\), one-hidden-layer
networks with a non-polynomial activation.  For every \(\eps>0\) there is a
finite implementation class \(\Himp^{\eps}\subset\calA\) with
\(|\Himp^{\eps}|\le N(\calH_k,\|\cdot\|_\infty,\eps/2)\) and an
implementation map with \(\errimp(k)\le\eps\).
\item[(b)]
(Layerwise.)  If every transition is \(\Lambda\)-Lipschitz, every output layer is
\(L_H\)-Lipschitz, and each generator and output layer is implemented within
uniform error \(\delta\) and \(\delta_H\), respectively
(\(\Lambda,\delta,\delta_H\ge0\)), then composing the
implemented layers gives
\(\errimp(k)\le\delta_H+L_H\,\delta\sum_{i=0}^{k-1}\Lambda^i\); this is
\(O(k\delta)\) for non-expanding transitions and \(O(\delta)\) for contractive
ones.
\item[(c)]
(Exact.)  If \(\xdom=\RR^d\) and the transitions are affine, as for the
nilpotent matrix actions of \cref{cond:p2-nilp}, then each transition is one
ReLU layer of width \(2d\) via \(x=\relu(x)-\relu(-x)\), every word in
\(B(k,F)\) is a depth-\(k\) ReLU network of width \(2d\), and \(\errimp(k)=0\).
\end{enumerate}
\end{proposition}

Part (a) shows that the implementation class is generally a finite class that
does not contain \(\calH_k\), with an error that can be made arbitrarily
small at the price of a larger class; its size is controlled by the covering
number of the abstract class.  Part (b) shows how a per-layer resolution
\(\delta\), for instance a width- or precision-dependent quantity, interacts
with depth: additively under layerwise stability and geometrically without
it.  Part (c) answers the implementation question for the Heisenberg and
upper-triangular examples of \cref{sec:growth-conditions-proof}: no
exponential width is needed, and the polynomial growth bounds of P2 apply to
a concrete ReLU network with zero implementation error.  Quantitative
finite-width results on manifolds and symmetric spaces
\citep{Kratsios2022,chen2019manifold,sonoda2022symmetric,Petersen2020} give
further examples of controlled \(\errimp\) under additional assumptions on
the target and the region being approximated; we do not claim that P2 alone
yields an efficient implementation of every nilpotent action.

\paragraph{Width through the implementation error.}
Part (b) makes the role of the width explicit.  Suppose the abstract
hidden layers are \(\Lambda\)-Lipschitz maps whose coordinates belong to a
smoothness class on the bounded state set, and that each of them is
implemented by a block of width \(w\) with uniform accuracy \(\delta(w)\);
for example \(\delta(w)\asymp w^{-s/d}\) for \(C^s\) layers on a bounded
subset of \(\RR^d\) \citep{DeVore1989,Yarotsky2017}, and \(\delta(w)\asymp
w^{-1/2}\) in \(L^2\) for layers of bounded variation norm
\citep{Barron1993,Jones1992,Pisier1981}.  Then, with a fixed output layer,
\[
\errimp(k)\le\delta_H+L_H\,\delta(w)\,S_k(\Lambda),
\qquad
S_k(\Lambda)=\sum_{i<k}\Lambda^i ,
\]
and the implementation error is below the estimation term \(\var(k,n)\) as
soon as \(\delta(w)\le\var(k,n)/(L_HS_k(\Lambda))\).  For non-expanding
layers and the root-polynomial profile \(\var(k,n)\asymp\sqrt{k/n}\) this
requires \(w\gtrsim(L_H\sqrt{kn})^{d/s}\): the width needed to realize a
depth-\(k\) abstract network without changing its regime grows polynomially
in \(k\) and \(n\), and for expanding layers geometrically in \(k\).  Width
therefore enters the depth trade-off through two routes, the one-layer
covering number \(p\asymp w^2\) in \cref{prop:envelope} and the per-layer
accuracy \(\delta(w)\) here, and both are explicit; a joint optimization
over \((k,w)\) would balance the three terms \(\bias(k)\), \(\var(k,n)\), and
\(\beta_\ell\errimp(k)\) simultaneously, which we do not pursue.

\paragraph{Proof of \cref{prop:implementation}.}
We restate \cref{prop:implementation} in slightly more general form and prove
it.  The point of the statement is that the implementation class in
\cref{thm:bv} is a genuine third object: it need not contain \(\calH_k\), and
the uniform implementation error can be traded against the size or the
resolution of the implementation.

\begin{proposition}[\cref{prop:implementation}, restated]
\label{prop:implementation-restated}
\begin{enumerate}
\item[(a)]
(Net-based implementation.)  Let \(\calH_k\subset\RR^{\xdom}\) be totally
bounded in the uniform norm, and let \(\calA\subset\RR^{\xdom}\) be
uniformly dense on \(\calH_k\): for every \(g\in\calH_k\) and \(\eta>0\)
there is \(a\in\calA\) with \(\|g-a\|_\infty\le\eta\) (no compactness of the
domain and no continuity is needed; on a compact \(K\) any dense
\(\calA\subset C(K)\) qualifies).  Then for every \(\eps>0\) there
exist a finite class \(\Himp^{\eps}\subset\calA\) with
\(|\Himp^{\eps}|\le N(\calH_k,\|\cdot\|_\infty,\eps/2)\) and an implementation map
\(\embed:\calH_k\to\Himp^{\eps}\) such that
\(\sup_{g\in\calH_k}\|g-\embed(g)\|_\infty\le\eps\).
\item[(b)]
(Layerwise implementation.)  Let \(F\subset C(\xdom,\xdom)\) with
\(\lip(f)\le\Lambda\) for all \(f\in F\) (\(\Lambda\ge0\)), and let \(H\subset C(\xdom)\) with
\(\lip(h)\le L_H\) for all \(h\in H\).  Suppose every \(f\in F\) is assigned
an implemented map \(\tilde f:\xdom\to\xdom\) with \(d_\infty(f,\tilde f)\le\delta\)
(\(\delta\ge0\))
and every \(h\in H\) an implemented \(\tilde h:\xdom\to\RR\) with
\(\|h-\tilde h\|_\infty\le\delta_H\).  For each element of \(\calH_k\) fix one
representation \(h\circ f_m\circ\cdots\circ f_1\) with \(m\le k\) and define
\(\embed(h\circ f_m\circ\cdots\circ f_1):=\tilde h\circ\tilde f_m\circ\cdots\circ\tilde f_1\).
Then
\[
\errimp(k)\le \delta_H+L_H\,\delta\sum_{i=0}^{k-1}\Lambda^i ,
\]
which is at most \(\delta_H+L_Hk\delta\) if \(\Lambda\le1\) and at most
\(\delta_H+L_H\delta/(1-\Lambda)\) if \(\Lambda<1\).
\item[(c)]
(Exact implementation of affine transitions.)  If \(\xdom=\RR^d\) and every
\(f\in F\) is affine, \(f(x)=Ax+b\), then each \(f\) is realized exactly by
one ReLU layer of width \(2d\), and every element of \(B(k,F)\) by a ReLU
network of depth \(k\) and width \(2d\).  With a linear output-layer class,
\(\calH_k\subset\Himp\) for the corresponding ReLU network class and
\(\errimp(k)=0\).
\end{enumerate}
\end{proposition}

\begin{proof}
(a) Let \(\{g_1,\dots,g_M\}\subset\calH_k\) be an \(\eps/2\)-net of
\(\calH_k\) in \(\|\cdot\|_\infty\) with centres in \(\calH_k\) and of
minimal cardinality \(M=N(\calH_k,\|\cdot\|_\infty,\eps/2)\) (internal
covering number).  By density on \(\calH_k\), choose
\(a_j\in\calA\) with \(\|a_j-g_j\|_\infty\le\eps/2\), set
\(\Himp^{\eps}:=\{a_1,\dots,a_M\}\), and let \(\embed(g):=a_{j(g)}\) where
\(j(g)\) is the index of a net point with \(\|g-g_{j(g)}\|_\infty\le\eps/2\).
Then \(\|g-\embed(g)\|_\infty\le\|g-g_{j(g)}\|_\infty+\|g_{j(g)}-a_{j(g)}\|_\infty\le\eps\),
and \(|\Himp^{\eps}|\le M\) since the \(a_j\) need not be distinct.

(b) Fix a representation \(h\circ f_m\circ\cdots\circ f_1\) and \(x\in\xdom\).
Put \(y_0:=x\), \(y_j:=\tilde f_j(y_{j-1})\), and
\(G_j:=f_m\circ\cdots\circ f_{j+1}\) for \(0\le j\le m\), with
\(G_m:=\id\); then \(\lip(G_j)\le\Lambda^{m-j}\).  Set
\(T_j:=G_j(y_j)\), so that \(T_0=f_m\circ\cdots\circ f_1(x)\) and
\(T_m=\tilde f_m\circ\cdots\circ\tilde f_1(x)\).  Since
\(T_{j-1}=G_j(f_j(y_{j-1}))\) and \(T_j=G_j(\tilde f_j(y_{j-1}))\),
\[
d(T_{j-1},T_j)\le\Lambda^{m-j}\,d\bigl(f_j(y_{j-1}),\tilde f_j(y_{j-1})\bigr)
\le\Lambda^{m-j}\delta .
\]
By the triangle inequality,
\(d(T_0,T_m)\le\delta\sum_{j=1}^m\Lambda^{m-j}=\delta\sum_{i=0}^{m-1}\Lambda^i\).
Finally,
\[
|h(T_0)-\tilde h(T_m)|
\le
|h(T_0)-h(T_m)|+|h(T_m)-\tilde h(T_m)|
\le
L_H\,\delta\sum_{i=0}^{m-1}\Lambda^i+\delta_H ,
\]
and taking the supremum over \(x\), over the representations with
\(m\le k\) (the sum is nondecreasing in \(m\) because \(\Lambda\ge0\)), and
over \(h\in H\) gives the claim.

(c) Since \(x=\relu(x)-\relu(-x)\) coordinatewise, one has
\(Ax+b=[A,-A]\,\relu\bigl([I_d;-I_d]\,x\bigr)+b\), a ReLU layer of width
\(2d\) followed by an affine map.  Composing \(k\) such blocks, and merging
each output affine map with the next input affine map, gives a depth-\(k\)
ReLU network of width \(2d\) that computes \(f_k\circ\cdots\circ f_1\)
exactly.  A linear output layer composed with it is again such a network.
\end{proof}

\paragraph{Remarks.}
In (a), \(\calA\) may be taken to be one-hidden-layer networks with a
non-polynomial activation \citep{Cybenko1989,Hornik1989,Leshno1993,Pinkus.survey},
or finite expansions in a universal reproducing kernel; the required widths
depend on \(\eps\) and on the regularity of the net points \(g_j\), and
quantitative rates require additional assumptions on the target and the
domain \citep{Barron1993,Kratsios2022,chen2019manifold,sonoda2022symmetric,Petersen2020}.
In (b), the bound displays the interaction between per-layer resolution
\(\delta\) (for instance a width- or precision-dependent quantity) and depth:
it is additive in \(k\) for non-expanding transitions and geometric without
layerwise stability.  The implemented maps \(\tilde f\) need not be
continuous or Lipschitz; only the abstract transitions are used to propagate
the error.  In (c), the nilpotent examples of \cref{cond:p2-nilp}, such as
\cref{ex:p2-noncpt-ut}, fall under this case, so their polynomial growth
bounds apply verbatim to a concrete ReLU implementation with zero
implementation error; on a torus
the torus itself is the state space, and the same affine layers act after
reduction modulo \(\ZZ^d\).

\section{Worked example: deep ReLU networks}
\label{sec:app-relu}

This appendix places the fully connected ReLU network, the architecture on
which most of the theory of depth has been developed
(\cref{sec:lit-overview}), in the state-transition model, and shows that
the Lipschitz constant of the layers alone moves the network across the
growth mechanisms of \cref{sec:growth-conditions}: contractive layers
saturate, non-expanding layers on a bounded state space saturate as well,
and expanding layers can realize exponential growth.  The ReLU examples in
the literature are then read as instances of the resulting regimes, and a
computed illustration (\cref{fig:relu-example}) shows the three growth
behaviours for three two-layer families that differ only in their
Lipschitz constant.

\subsection{Setting}
\label{sec:relu-setting}

Let \(K\subset\RR^m\) be a nonempty bounded set of diameter \(D_K\), for
instance \(K=[-1,1]^m\), let \(R_K\ge0\) be a bound with \(\|x\|\le R_K\) for
all \(x\in K\), and let \(\Pi_K\) be a
\(1\)-Lipschitz retraction of \(\RR^m\) onto \(K\), such as the
coordinatewise clipping for a box.  A hidden layer is a ReLU block of
width \(w\) that maps the state space to itself,
\[
f_\vartheta(x):=\Pi_K\bigl(V\relu(Wx+b)+c\bigr),
\qquad
\vartheta=(W,b,V,c)\in\RR^{w\times m}\times\RR^{w}\times\RR^{m\times w}\times\RR^{m},
\]
where \(\relu\) acts coordinatewise, and the hidden-layer class is
\[
F_\Lambda:=\{f_\vartheta:\ \|W\|_{\rm op},\|V\|_{\rm op}\le\beta_W,\ \|b\|,\|c\|\le\beta,\ \lip(f_\vartheta)\le\Lambda\},
\]
with \(p:=2mw+w+m\) parameters, where \(\beta_W,\beta,\Lambda\ge0\).  The retraction keeps the state space
bounded, which bounded activations or normalization layers achieve in
practice, and it is the identity whenever the block output stays in
\(K\).  The output layer is linear,
\(H=\{x\mapsto\langle u,x\rangle:\|u\|\le R\}\), so
\cref{ass:sg-increment-main} holds with depth-independent constants
(\cref{prop:hilbert-sg}), and a depth-\(k\) network
\(h\circ f_{\vartheta_k}\circ\cdots\circ f_{\vartheta_1}\) is a fully
connected ReLU network of depth \(2k\) with clipped states of dimension
\(m\) after every second layer; the Lipschitz constraint restricts the
class to blocks of Lipschitz constant at most \(\Lambda\) (the product
\(\|V\|_{\rm op}\|W\|_{\rm op}\) is always an upper bound for it).

\begin{lemma}[Covering numbers of one ReLU block]
\label{lem:relu-layer-covering}
For \(f_\vartheta,f_{\vartheta'}\in F_\Lambda\),
\[
d_\infty(f_\vartheta,f_{\vartheta'})
\le
\beta_W R_K\|W-W'\|_{\rm op}+\beta_W\|b-b'\|+(\beta_WR_K+\beta)\|V-V'\|_{\rm op}+\|c-c'\|,
\]
and hence, for every \(\eps>0\),
\(\log N(F_\Lambda,d_\infty,\eps)\le p\log(1+C_F/\eps)\) with
\(C_F:=2\,(2\beta_WR_K+\beta_W+\beta+1)\max\{\beta_W,\beta\}\);
in particular \(\log N(F_\Lambda,d_\infty,\eps)\le p\log(1/\eps)+O(p)\) for
\(\eps\le C_F\).
\end{lemma}

\begin{proof}
For \(x\in K\), write \(z=\relu(Wx+b)\) and \(z'=\relu(W'x+b')\); then
\(\|z-z'\|\le R_K\|W-W'\|_{\rm op}+\|b-b'\|\) and
\(\|z'\|\le\beta_WR_K+\beta\).  Hence
\(\|Vz+c-V'z'-c'\|\le\|V\|_{\rm op}\|z-z'\|+\|V-V'\|_{\rm op}\|z'\|+\|c-c'\|\),
which gives the displayed bound after applying the \(1\)-Lipschitz
retraction.  The parameter tuple lives in a real normed space of dimension \(p\)
(operator norms on \(W,V\), Euclidean norms on \(b,c\), the maximum over
the four blocks), in which the parameter set is contained in the ball of
radius \(\max\{\beta_W,\beta\}\), and the displayed estimate shows that
\(\vartheta\mapsto f_\vartheta\) is \((2\beta_WR_K+\beta_W+\beta+1)\)-Lipschitz
from this norm to \(d_\infty\).  A ball of radius \(\rho\) in a real normed
space of dimension \(p\) has a \(\delta\)-separated subset of at most
\((1+2\rho/\delta)^p\) points (compare the volumes of disjoint balls of
radius \(\delta/2\)), hence a \(\delta\)-cover of that size; the image of
such a cover under the Lipschitz map is an \(\eps\)-cover of \(F_\Lambda\)
when \(\delta=\eps/(2\beta_WR_K+\beta_W+\beta+1)\), which gives the stated
\(C_F\).
\end{proof}

\subsection{Variance: three growth regimes from the Lipschitz constant}
\label{sec:relu-regimes}

\begin{proposition}[Depth profiles of ReLU networks by Lipschitz constant]
\label{prop:relu-regimes}
Let \(F=F_\Lambda\) on the bounded set \(K\), and let
\(\mathsf V_k(S)\) be the entropy integral of \cref{thm:hidden-decomp}.
\begin{enumerate}
\item[(i)]
(Contractive, \(0<\Lambda<1\).)  \Cref{cond:p1-ucont} applies with the
invariant set \(K\) itself, so
\(\sup_kN(B(k,F_\Lambda),d_\infty,\eps)<\infty\) for every \(\eps>0\), and
\(\mathsf V_k(S)=O(1)\).  Explicitly, with
\(m(\eps):=\lceil\log(2D_K/\eps)/\log(1/\Lambda)\rceil\),
\(\log N(B(k,F_\Lambda),d_\infty,\eps)\le\log\bigl(N(K,\eps/2)+m(\eps)N(F_\Lambda,d_\infty,(1-\Lambda)\eps)^{m(\eps)}\bigr)\),
which does not depend on \(k\).
\item[(ii)]
(Non-expanding, \(\Lambda\le1\).)  If \(K\) is compact,
\cref{cond:p1} applies (non-expanding generators on a compact state
space), so again \(\sup_kN(B(k,F_\Lambda),d_\infty,\eps)<\infty\) and
\(\mathsf V_k(S)=O(1)\); the explicit layerwise bound of \cref{prop:envelope}
gives the cruder but quantitative bound
\(\mathsf V_k(S)\le D_K\bigl(\sqrt{\log(k+1)}+\sqrt{kp\log k}+\sqrt{kp}\,(\sqrt{\log(1+2C_F/D_K)}+\sqrt\pi/2)\bigr)=O(\sqrt{kp\log k})\)
for \(k\ge1\), which needs only boundedness of \(K\).
\item[(iii)]
(Expanding, \(\Lambda>1\).)  \Cref{prop:envelope} gives
\(\mathsf V_k(S)\le D_K\bigl(\sqrt{\log(k+1)}+\sqrt{kp\log k}+k\sqrt{p\log\Lambda}+\sqrt{kp}\,(\sqrt{\log(1+2C_F/D_K)}+\sqrt\pi/2)\bigr)=O(k\sqrt{p\log\Lambda})\),
and this order cannot be improved in general: for \(m=1\), \(K=[0,1]\)
with the clipping retraction, width \(w\ge4\), \(\beta_W\ge41\),
\(\beta\ge2\), and every \(\Lambda\ge20\), the class \(F_\Lambda\) contains
the two piecewise-linear expand-and-reset maps of \cref{sec:relu-computed},
which satisfy \cref{cond:e2-pingpong}
with two generators, so \(N(B(k,F_\Lambda),d_\infty,\eps)\ge2^k\) for all
\(\eps<1/16\) and \(\mathsf V_k(S)\) grows at least like \(\sqrt k\) on a
sample that contains the probe points (\cref{lem:probes-packing}).
\end{enumerate}
\end{proposition}

\begin{proof}
(i) Every map in \(F_\Lambda\) is \(\Lambda\)-Lipschitz with \(\Lambda<1\)
and maps \(K\) into \(K\), so \(K\) is a nonempty invariant set and a
bounded absorbing set with \(L=0\); \cref{cond:p1-ucont} gives the bound
with \(N(F_\Lambda^{\,\ell},d_\infty,\eps)\le N(F_\Lambda,d_\infty,(1-\Lambda)\eps)^{\ell}\)
by the telescoping estimate of \cref{prop:envelope}, and the right-hand
side is finite by \cref{lem:relu-layer-covering}.  (ii) The maps are
non-expanding self-maps of the compact set \(K\), which is condition 2c of
\cref{cond:p1}.  The layerwise bound statements in (ii) and (iii) are
\cref{cor:envelope-profiles}(a) with \cref{lem:relu-layer-covering}; since
the lemma bounds the entropy by \(p\log(1+C_F/\eps)\) rather than by
\(p\log(C_F/\eps)\), the proof of the corollary is repeated with this
hypothesis, which changes \(\log(2C/\overline D)\) into
\(\log(1+2C_F/D_K)\) and nothing else.  For the lower bound, a continuous
piecewise-linear map \([0,1]\to[0,1]\) with \(q\) interior breakpoints is
a ReLU block \(x\mapsto V\relu(Wx+b)+c\) of width \(q+1\), and
the two maps of \cref{sec:relu-computed} have Lipschitz constant \(20\),
are ReLU blocks of width \(4\) with weight norms at most \(41\) and bias
norms at most \(2\), and satisfy \cref{cond:e2-pingpong} with separation
\(1/8\), as verified there; the entropy lower bound is \cref{cond:e2-pingpong}, and the transfer
to the sample is \cref{lem:probes-packing}.
\end{proof}

The three cases are the three rows of the balance in \cref{sec:optdepth}
on the variance side.  Contractive and non-expanding ReLU layers on a
bounded state space have a saturated profile, so their estimation bound
does not grow with the depth; expanding layers have a polynomial profile at
best, between \(\sqrt k\) and \(k\), so their estimation bound grows with
the depth and has to be balanced against the bias.  The dependence on the
width is through \(p=2mw+w+m\) in the explicit bounds and through the
one-layer covering numbers in the saturated ones.

\subsection{Bias and balancing}
\label{sec:relu-bias}

The bias depends on the target class, and the ReLU literature provides
three cases.  For a teacher--student target, \(\Con=H\circ\overline{\langle
F_\Lambda\rangle}\) with \(\Lambda<1\), the truncation argument of
\cref{sec:examples-regime} gives \(\bias(k)\le\beta_\ell RD_K\Lambda^k\),
and with case (i) the network is in the EL regime: a contractive ReLU
network may be run to convergence.  For H\"older, Sobolev, or Besov
targets, the approximation error of depth-\(k\) ReLU networks decays
polynomially in the number of parameters and hence in the depth
\citep{Yarotsky2017,Siegel2023,suzuki2018besov-adaptivity}, so the network
is in the PP regime with the profile of case (ii) or (iii): this is the
setting of the VC-dimension and nonparametric analyses reviewed in
\cref{sec:lit-overview}, whose polynomial depth dependence is the layerwise bound
of \cref{prop:envelope}.  For the hierarchical compositions of
\citet{Schmidt-Hieber2020}, the approximation error has an exponentially
decaying compositional part and the covering-number bounds grow
logarithmically in the depth, close to the PL regime.  In each case the
framework does not change the known rate; it identifies which of the three
mechanisms of \cref{prop:relu-regimes} the corresponding proof uses, and
the Lipschitz constant of the layers as the quantity that decides between
them.

\subsection{Computed illustration (\cref{fig:relu-example})}
\label{sec:relu-computed}

\Cref{fig:relu-example} is computed exactly, without randomness, in the
following setting.  The state space is \(K=[0,1]\) with the clipping retraction,
and each family consists of two one-hidden-layer ReLU maps
\([0,1]\to[0,1]\); the three families differ only in their Lipschitz
constant.
\begin{itemize}
\item Contractive (\(\Lambda=1/4\)): \(f_0(x)=x/4\) and \(f_1(x)=(x+3)/4\),
  the affine maps of a Cantor-type attractor, as in \cref{ex:p1-cantor}
  with ratio \(1/4\).
\item Non-expanding (\(\Lambda=1\)): \(f_0(x)=\min\{1,\relu(x+\tfrac1{10})\}\)
  and \(f_1(x)=\min\{1,\relu(x+\tfrac1{10\sqrt2})\}\), translations by two
  incommensurable positive steps followed by clipping; each is a ReLU
  network of width \(2\).  Since both steps are positive, the compositions
  commute exactly, \(f_1\circ f_0=f_0\circ f_1=\min\{1,\relu(x+\tfrac1{10}+\tfrac1{10\sqrt2})\}\),
  and a composition of \(n_0\) steps of the first kind and \(n_1\) of the
  second is the clipped translation by \(n_0/10+n_1/(10\sqrt2)\): this is
  the abelian mechanism P2 with two generators, and the sums stay below
  \(1\) for \(n_0+n_1\le9\).
\item Expanding (\(\Lambda=20\)): the piecewise-linear maps with
  breakpoints \((0,\tfrac38),(\eta,0),(\tfrac14-\eta,1),(\tfrac14,\tfrac38),(1,\tfrac38)\)
  and \((0,\tfrac58),(\tfrac34,\tfrac58),(\tfrac34+\eta,0),(1-\eta,1),(1,\tfrac58)\),
  \(\eta=1/32\).  Each is affine from its coding core
  \(V_0=[\eta,\tfrac14-\eta]\), respectively \(V_1=[\tfrac34+\eta,1-\eta]\),
  onto \([0,1]\), and constant equal to its anchor \(3/8\), respectively
  \(5/8\), outside its chamber \(U_0=[0,\tfrac14]\), respectively
  \(U_1=[\tfrac34,1]\).  This is \cref{cond:e2-pingpong} with the marker
  \(q=1/2\): the chambers are disjoint, \(f_i(V_i)=[0,1]\), every point
  outside \(U_i\) is reset to the anchor \(a_i\), the anchors lie outside
  both chambers, and \(\dist(q,\{a_0,a_1\})=1/8\).  Each map is a ReLU
  block of width \(4\) (\cref{prop:relu-regimes}(iii)).
\end{itemize}
The left three panels show the graphs of all eight compositions of depth
\(k=3\) of each family: contractive compositions are almost constant,
non-expanding compositions are clipped translations, and expanding
compositions are separated by their behaviour on nested cores of width
\((3/16)^{k}\).

The right panel counts the compositions of depth at most \(k\) that are
distinguishable at resolution \(\eps=10^{-2}\), that is, the size of a
greedy \(\eps\)-net of \(B(k,F)\) in the sup distance over a finite
evaluation set, for \(k\le9\).  For the contractive and the non-expanding
family the evaluation set is the uniform grid of \(4001\) points in
\([0,1]\).  For the expanding family the grid alone cannot resolve the
coding cores beyond depth \(5\) (their width is \((3/16)^k\)), so the
evaluation set is the grid together with the probe points \(x_w\) of every
word \(w\) of length at most \(9\), computed as
\(x_w=g_{i_1}\circ\cdots\circ g_{i_k}(q)\) with \(g_i\) the inverse of
\(f_i\) on \(V_i\), at which \(f_w(x_w)=q\) while every other word of the
same length maps \(x_w\) to an anchor.  On this set the count equals
\(2^{k+1}-1\), the number of words, as guaranteed by
\cref{cond:e2-pingpong}; on the grid alone it would follow \(2^{k+1}-1\)
up to depth \(5\) and then saturate, which is the empirical-entropy effect
of \cref{sec:cot-computed}.  The three families realize the three growth
cases of \cref{prop:relu-regimes} and of \cref{sec:growth-conditions}:
\begin{itemize}
\item the contractive count is \(1,3,7,15,31\) for \(k\le4\) and stays at
  \(31\) afterwards, since compositions of depth \(\ge5\) are within
  \(4^{-5}<\eps\) of constant maps; the level is the covering number of the
  attractor at resolution \(\eps\), a property of the attractor and not of
  the depth (case (i), P1');
\item the non-expanding count is \(1,3,6,10,15,21,28,35,42,49\): the
  triangular numbers \((k+1)(k+2)/2\) up to \(k=6\), that is, one map per
  pair \((n_0,n_1)\) with \(n_0+n_1\le k\), and slightly fewer afterwards
  because some sums fall within \(\eps\) of each other; this is the
  polynomial fixed-scale growth of the abelian mechanism P2 (case (ii),
  which by P1 saturates once the sums exceed \(1\));
\item the expanding count grows like \(2^k\) in the uniform distance
  (case (iii), E2).
\end{itemize}
\subsection{Implementation and literature}
The hidden-layer class \(F_\Lambda\) is itself a class of ReLU blocks, so
the implementation map is the identity and \(\errimp(k)=0\); the
implementation error becomes relevant only when a ReLU block approximates
a layer of another type, which is the situation of
\cref{prop:implementation}(b).  The depth dependences established for ReLU
networks in the literature, and their place among the mechanisms of this
paper, are reviewed in \cref{sec:lit-overview}.

\section{Worked example: chain-of-thought style symbolic computation}
\label{sec:app-cot}

This appendix applies the framework to a stylized model of step-by-step
symbolic computation, in which each hidden layer is one reasoning step
acting on a scratchpad and the depth \(k\) is the number of steps.  The
purpose is to show that the statistical cost of the number of steps can be
read off from the metric growth of the compositions of steps, separately
from any particular implementation (for example a recurrently applied
transformer block).  Two step families are analyzed: an append-only
scratchpad, which is contractive and lands in the EL regime, and a branching
computation that reads the current symbol, which is expansive, has
exponentially growing uniform entropy, and yet has root-logarithmic
empirical entropy on every sample, because a sample of size \(n\) can
exercise only finitely many branches.  Throughout, ``chain of thought'' refers to
this abstract model and not to a specific language model; the relation to
the literature on chain-of-thought reasoning is discussed at the end.

\subsection{Setting}
\label{sec:cot-setting}

\paragraph{Scratchpad space.}
Let \(\calA\) be a finite alphabet with \(|\calA|=m\ge2\) and let
\(\xdom=\calA^{\NN}=\{x=(x_0,x_1,\dots):x_j\in\calA\}\) be the space of
infinite symbol sequences, equipped with the ultrametric
\[
d_\theta(x,y)=\theta^{n(x,y)}\quad(x\ne y),
\qquad
n(x,y):=\min\{j\ge0: x_j\ne y_j\},
\]
for a fixed \(\theta\in(0,1)\), and \(d_\theta(x,x)=0\).  Then
\((\xdom,d_\theta)\) is compact with \(\diam(\xdom)=1\), and two sequences
are within distance \(\theta^{L}\) if and only if they agree in the first
\(L\) symbols (see \cref{sec:growth-conditions-proof} for the standard
facts used below).  A finite scratchpad of length \(N\) can be embedded by
padding with a fixed symbol.

\paragraph{Output layers.}
The answer is read from the scratchpad through an output layer.  We use two
linear output-layer classes on explicit feature maps, so that
Assumption~\ref{ass:sg-increment-main} holds with depth-independent
constants by \cref{prop:hilbert-sg}.
\begin{itemize}
\item
(Window output layer.)  For \(L\ge1\), let \(\Phi_L:\xdom\to\RR^{\calA^L}\)
be the one-hot encoding of the window \((x_0,\dots,x_{L-1})\), and
\(H_L:=\{x\mapsto\langle w,\Phi_L(x)\rangle:\|w\|_2\le R\}\).
\item
(Geometrically weighted output layer.)  Let
\(\Phi_\infty:\xdom\to\ell^2(\NN;\RR^{\calA})\),
\(\Phi_\infty(x):=(\theta^j e_{x_j})_{j\ge0}\), where \(e_a\) is the
one-hot vector of \(a\in\calA\), and
\(H_\infty:=\{x\mapsto\langle w,\Phi_\infty(x)\rangle:\|w\|_{\ell^2}\le R\}\).
\end{itemize}

\begin{lemma}
\label{lem:cot-output}
\(\Phi_L\) is \(\sqrt2\,\theta^{1-L}\)-Lipschitz and \(\Phi_\infty\) is
\(\sqrt{2/(1-\theta^2)}\)-Lipschitz with respect to \(d_\theta\), and
\(\|\Phi_L\|\le1\), \(\|\Phi_\infty\|\le(1-\theta^2)^{-1/2}\).
Consequently, Assumption~\ref{ass:sg-increment-main} holds for every
hidden-layer class on \(\xdom\), with \(A_H=1\) and \(L=\sqrt2\,\theta^{1-L}\)
for \(H_L\), and with \(A_H=1\) and \(L=\sqrt{2/(1-\theta^2)}\) for
\(H_\infty\); moreover \(\erad_S(H_L)\le R/\sqrt n\) and
\(\erad_S(H_\infty)\le R/\sqrt{n(1-\theta^2)}\).
\end{lemma}

\begin{proof}
If \(d_\theta(x,y)\le\theta^L\), the first \(L\) symbols agree and
\(\Phi_L(x)=\Phi_L(y)\).  Otherwise \(d_\theta(x,y)\ge\theta^{L-1}\) and
\(\|\Phi_L(x)-\Phi_L(y)\|\le\sqrt2\le\sqrt2\,\theta^{1-L}d_\theta(x,y)\).
For \(\Phi_\infty\), with \(n=n(x,y)\),
\(\|\Phi_\infty(x)-\Phi_\infty(y)\|^2=\sum_{j\ge n}\theta^{2j}\|e_{x_j}-e_{y_j}\|^2\le 2\theta^{2n}/(1-\theta^2)\),
so \(\|\Phi_\infty(x)-\Phi_\infty(y)\|\le\sqrt{2/(1-\theta^2)}\,d_\theta(x,y)\).
The norm bounds are immediate.  The increment condition follows from
\cref{prop:hilbert-sg}, and the Rademacher bounds from
\(\erad_S(H)=R\,\EE_\sigma\|n^{-1}\sum_i\sigma_i\Phi(X_i)\|\le R\sup\|\Phi\|/\sqrt n\).
\end{proof}

\paragraph{Loss and learning problem.}
Labels are real-valued, the loss is bounded and \(\beta_\ell\)-Lipschitz as
in \cref{sec:setting}, and the learner minimizes the empirical risk over
\(\calH_k=H\circ B(k,F)\), that is, over output layers composed with
programs of at most \(k\) steps.

\subsection{Append-only steps: saturation and the EL regime}
\label{sec:cot-append}

\paragraph{Steps.}
A write step prepends a symbol:
\(g_a(x):=(a,x_0,x_1,\dots)\) for \(a\in\calA\), and
\(F_{\rm w}:=\{g_a:a\in\calA\}\).  For a word \(u=(u_1,\dots,u_j)\in\calA^j\)
write \(g_u:=g_{u_j}\circ\cdots\circ g_{u_1}\), so that
\(g_u(x)=(u_j,u_{j-1},\dots,u_1,x_0,x_1,\dots)\): the scratchpad records
the steps in reverse order and the input is pushed back.  Each \(g_a\)
satisfies \(d_\theta(g_a(x),g_a(y))=\theta\,d_\theta(x,y)\), so \(F_{\rm w}\)
is a contractive hidden-layer class with \(\lip(g_a)=\theta<1\), and
\cref{cond:p1} applies.

\begin{lemma}[Saturation for append-only steps]
\label{lem:cot-append-growth}
For \(\eps>0\) let \(\ell(\eps):=\max\{0,\lceil\log(1/\eps)/\log(1/\theta)\rceil\}\).
Then for all \(k\ge0\) and all \(\eps>0\),
\[
N\bigl(B(k,F_{\rm w}),d_\infty,\eps\bigr)\le m^{\min\{k,\ell(\eps)\}+1},
\qquad
\mathsf V_k(S)\le
\sqrt{\log m}\Bigl(\frac{\sqrt\pi}{2\sqrt{\log(1/\theta)}}+\sqrt2\Bigr)
=:\mathsf V_\infty .
\]
In particular \(\var(k,n)\le 12A_HL\,\mathsf V_\infty/\sqrt n\) for every
\(k\) (case (i) of \cref{prop:profiles}).
\end{lemma}

\begin{proof}
If \(|u|\ge\ell\) and \(u'\) consists of the last \(\ell\) letters of \(u\),
then \(g_u(x)\) and \(g_{u'}(x)\) agree in the first \(\ell\) symbols for
every \(x\), so \(d_\infty(g_u,g_{u'})\le\theta^{\ell}\).  With
\(\ell=\ell(\eps)\) we have \(\theta^{\ell}\le\eps\), hence the compositions
of length at most \(\min\{k,\ell(\eps)\}\) form an \(\eps\)-cover of
\(B(k,F_{\rm w})\), and there are at most
\(\sum_{j\le\min\{k,\ell(\eps)\}}m^j\le m^{\min\{k,\ell(\eps)\}+1}\) of them.
Since \(\diam(\xdom)=1\) and \(d_S\le d_\infty\),
\[
\mathsf V_k(S)\le\int_0^1\sqrt{(\ell(\eps)+1)\log m}\,\dd\eps
\le\sqrt{\log m}\int_0^1\Bigl(\sqrt{\tfrac{\log(1/\eps)}{\log(1/\theta)}}+\sqrt2\Bigr)\dd\eps ,
\]
using \(\ell(\eps)+1\le\log(1/\eps)/\log(1/\theta)+2\) and
\(\sqrt{a+b}\le\sqrt a+\sqrt b\).  The identity
\(\int_0^1\sqrt{\log(1/\eps)}\,\dd\eps=\sqrt\pi/2\) (\cref{lem:log-split})
gives the claim.
\end{proof}

\paragraph{Bias.}
Take the target class to be the teacher--student class of arbitrarily long
programs with the geometrically weighted output layer,
\(\Con:=H_\infty\circ\overline{\langle F_{\rm w}\rangle}^{\,d_\infty}\).
For a teacher \(c=h_w\circ g_u\circ g_v\) with \(|u|=k\) and any \(v\),
\cref{lem:cot-output} and \(\lip(g_u)=\theta^k\) give
\[
|c(x)-h_w(g_u(x))|
\le
R\sqrt{\tfrac{2}{1-\theta^2}}\;d_\theta\bigl(g_u(g_v(x)),g_u(x)\bigr)
\le
R\sqrt{\tfrac{2}{1-\theta^2}}\;\theta^k ,
\]
and the same bound holds on the uniform closure.  By
\eqref{eq:approx-transfer},
\(\bias(k)=\errmodel(k)\le\beta_\ell R\sqrt{2/(1-\theta^2)}\,\theta^k\):
steps performed long ago contribute geometrically less to the answer.

\begin{proposition}[Append-only scratchpad]
\label{prop:cot-append}
In the setting above, with \(H=H_\infty\) and \(F=F_{\rm w}\), the
excess-risk bound of \cref{thm:bv} is, up to the depth-independent terms,
\[
\gen(k,n)\lesssim
\beta_\ell R\sqrt{\tfrac{2}{1-\theta^2}}\,\theta^k
+\frac{12\sqrt{2/(1-\theta^2)}\,\mathsf V_\infty}{\sqrt n},
\]
which is the EL regime with \(\biase=\log(1/\theta)\) and saturated variance.
The bound is minimized at \(\kopt=\frac{\log n}{2\log(1/\theta)}+O(1)\), with
balanced value of order \(n^{-1/2}\).
\end{proposition}

\begin{proof}
Combine \cref{thm:bv,thm:hidden-decomp}, \cref{lem:cot-output}, and
\cref{lem:cot-append-growth}; the balancing step is that of the EL row of
\eqref{eq:tradeoff-table} with a constant variance profile, which removes the
\(\log\log\) factors.
\end{proof}

With the window output layer \(H_L\) instead, the bias vanishes exactly for
\(k\ge L\), because the answer depends only on the last \(L\) writes; then
\(\kopt=L\) and additional steps cost nothing in the bound.  In both cases
the number of useful steps is small, of order \(\log n\) or a constant, and
further steps do not increase the estimation bound: append-only scratchpads are forgetful at the
output layer, and this forgetting is exactly the contraction that saturates
the entropy.

\subsection{Branching computation: exponential uniform entropy, logarithmic empirical entropy}
\label{sec:cot-branch}

\paragraph{Steps.}
Let \(\calA=[r]\cup\{\bullet\}\) with \(r\ge2\) active symbols and one
padding symbol.  A guarded step \(f_a\), \(a\in[r]\), reads the current
symbol: if it equals \(a\) the step consumes it and exposes the rest of the
scratchpad, \(f_a(x)=\sigma(x)=(x_1,x_2,\dots)\); otherwise the step fails
and resets the scratchpad to the constant error state \(\bar a=(a,a,\dots)\).
Let \(F_{\rm b}:=\{f_a:a\in[r]\}\).  This is the ping--pong family of
\cref{ex:e2-pingpong-subshift}, which satisfies \cref{cond:e2-pingpong} with
separation \(1\): on its chamber each \(f_a\) is the shift, with Lipschitz
constant \(\theta^{-1}>1\), so \(F_{\rm b}\) is expansive rather than
contractive.  A program \(u=(u_1,\dots,u_j)\) succeeds on exactly those inputs
whose first \(j\) symbols are \(u_1,\dots,u_j\), and then outputs the
remainder of the scratchpad.  Explicitly, writing \([u]\) for the cylinder
set of inputs starting with \(u\) and \(f_u:=f_{u_j}\circ\cdots\circ f_{u_1}\)
for \(j\ge1\) (\(f_{()}=\id\)),
\begin{equation}
f_u(x)=\sigma^{j}(x)\ \text{ if } x\in[u],
\qquad
f_u(x)=\bar u_j\ \text{ otherwise},
\label{eq:cot-branch-closed-form}
\end{equation}
since once a guard fails the scratchpad is the constant state of the failed
symbol, and each later guard either keeps it (if its symbol agrees) or
replaces it by its own constant state, so the final state is \(\bar u_j\).

\begin{lemma}[Exponential growth for branching steps]
\label{lem:cot-branch-growth}
For every \(k\ge0\) and \(\eps\in(0,1/2)\),
\[
r^k\le N\bigl(B(k,F_{\rm b}),d_\infty,\eps\bigr)\le r^{k+1},
\qquad
\mathsf V_k(S)\le\sqrt{(k+1)\log r},
\]
and hence \(\var(k,n)\le 12A_HL\sqrt{(k+1)\log r/n}\) (case (iii) of
\cref{prop:profiles}).
\end{lemma}

\begin{proof}
The lower bound is \cref{ex:e2-pingpong-subshift}.  The upper bound holds
because \(F_{\rm b}\) is finite, so \(B(k,F_{\rm b})\) has at most
\(\sum_{j\le k}r^j\le r^{k+1}\) elements (using \(r\ge2\)), and \(\diam(\xdom)=1\)
(\cref{prop:profiles}(iii)).
\end{proof}

\paragraph{Empirical entropy: what a sample can see.}
The exponential growth in \cref{lem:cot-branch-growth} is certified in the
uniform metric, and \cref{ex:e2-pingpong-subshift} shows that it is real.
The entropy integral of \cref{thm:hidden-decomp}, however, is taken in the
empirical metric \(d_S\), and for branching computation the two metrics
differ by an exponential factor, for every sample.  The reason is
\eqref{eq:cot-branch-closed-form}: programs of the same length have
pairwise disjoint success cylinders, so a sample of size \(n\) meets at most
\(n\) of them, and every program whose cylinder contains no sample point is
indistinguishable on the sample from a constant map.

\begin{lemma}[Empirical saturation for branching steps]
\label{lem:cot-branch-sample}
For every sample \(S=(X_1,\dots,X_n)\), every \(k\ge0\), and every
\(\eps>0\),
\[
N\bigl(B(k,F_{\rm b}),d_S,\eps\bigr)
\le
1+r+\sum_{j=1}^{k}\min\bigl\{r^j,\;n,\;\lfloor\eps^{-2}\rfloor\bigr\}
\le
1+r+k\min\{n,\eps^{-2}\},
\]
and consequently
\begin{gather*}
\mathsf V_k(S)
\le
\sqrt{\log(k+1)}+\sqrt{\log(1+r)}+\sqrt{2\log2}+\sqrt{\pi/2},
\\
\var(k,n)\le\frac{12A_HL}{\sqrt n}\Bigl(\sqrt{\log(k+1)}+\sqrt{\log(1+r)}+\sqrt{2\log2}+\sqrt{\pi/2}\Bigr),
\end{gather*}
the root-logarithmic profile (case (ii) of \cref{prop:profiles}), for every
sample and without any assumption on the input distribution.  (The constant
\(\sqrt{2\log2}\) comes from the internal/external transfer
\(N(B,d_S,\eps)\le N^{\rm ext}(B,d_S,\eps/2)\), since \(\mathsf V_k(S)\) is
formed with internal covering numbers.)
\end{lemma}

\begin{proof}
Let \(\hat P_n([u]):=n^{-1}\#\{i:X_i\in[u]\}\) be the empirical mass of the
success cylinder of \(u\).  By \eqref{eq:cot-branch-closed-form} and
\(\diam(\xdom)=1\),
\[
d_S(f_u,\bar u_j)^2
=\frac1n\sum_{i:X_i\in[u]}d_\theta\bigl(\sigma^j(X_i),\bar u_j\bigr)^2
\le\hat P_n([u]) .
\]
For fixed \(j\) the cylinders \([u]\), \(|u|=j\), are pairwise disjoint,
hence \(\sum_{|u|=j}\hat P_n([u])\le1\); therefore fewer than \(\eps^{-2}\)
programs of length \(j\) have \(\hat P_n([u])>\eps^2\), and also at most
\(n\) programs of length \(j\) have \(\hat P_n([u])>0\).  Every other program
of length \(j\) is within \(d_S\)-distance \(\eps\) of one of the \(r\)
constant maps \(\bar a\), \(a\in[r]\).  Taking as centres the identity, the
\(r\) constant maps, and, for each \(1\le j\le k\), the at most
\(\min\{r^j,n,\lfloor\eps^{-2}\rfloor\}\) programs with
\(\hat P_n([u])>\eps^2\), gives an external \(\eps\)-cover of
\(B(k,F_{\rm b})\) of the stated cardinality.  For the entropy integral, the internal covering number at radius
\(\eps\) is at most the external one at radius \(\eps/2\), which is at most
\(1+r+4k\eps^{-2}\le(1+r)(k+1)\,4\eps^{-2}\) for \(\eps\le1\), so
\(\sqrt{\log N}\le\sqrt{\log(1+r)}+\sqrt{\log(k+1)}+\sqrt{2\log2}+\sqrt{2\log(1/\eps)}\);
integrate over \((0,1]\) (recall \(D_k(S)\le\diam(\xdom)=1\)) and use
\(\int_0^1\sqrt{\log(1/\eps)}\,\dd\eps=\sqrt\pi/2\) (\cref{lem:log-split}).
\end{proof}

The lemma does not contradict the lower bound of
\cref{lem:cot-branch-growth}: the \(r^k\) programs of length \(k\) are
\(1\)-separated in \(d_\infty\), but all except at most \(n\) of them agree
with a constant map on the sample.  The output-layer separation condition of
\cref{cor:matching} (\cref{ass:readout-realization-main} with
depth-independent \(\kappa\)) therefore fails for this family, and the
\(\sqrt{k/n}\) upper bound obtained from the uniform entropy is not sharp
for any fixed sample size.  This is the general phenomenon behind
\cref{thm:hidden-decomp}: the empirical entropy of \(B(k,F)\), not its
uniform entropy, is the statistical cost of depth, and the transfer
\(d_S\le d_\infty\) used in \cref{sec:growth-conditions} is a one-sided
device that may lose an exponential factor.

\paragraph{Bias and balancing.}
Two target models are natural.
\begin{itemize}
\item
(Exact program.)  If the target is \(c=h\circ f_{u^\ast}\) for one program
\(u^\ast\) of length \(k_0\) and \(h\in H\), then \(c\in\calH_k\) for all
\(k\ge k_0\) and \(\bias(k)=0\) there.  The bound of \cref{thm:bv} is then
minimized at \(k=k_0\), with value of order
\(A_HL\sqrt{\log k_0/n}+R/\sqrt n\) by \cref{lem:cot-branch-sample}, and
additional steps beyond \(k_0\) cost only \(\sqrt{\log k/n}\).
\item
(Geometric relevance.)  If longer programs matter geometrically less, in the
sense that \(\bias(k)\le C_{\rm b}e^{-\biase k}\) for some
\(C_{\rm b},\biase>0\), then the bound is in the EL regime: by the EL row
of \eqref{eq:tradeoff-table}, \(\kopt=\frac{1}{2\biase}\log n+O(\log\log\log n)\)
and the balanced value is of order \(\sqrt{\log\log n/n}\).  In contrast to
the append-only case, this decay is an assumption on the target rather than
a consequence of the step geometry, because the guarded steps do not
contract.
\end{itemize}

\begin{proposition}[Branching computation]
\label{prop:cot-branch}
In the setting above, with \(H=H_L\) or \(H_\infty\) and \(F=F_{\rm b}\),
the excess-risk bound of \cref{thm:bv} is, up to the depth-independent
terms,
\[
\gen(k,n)\lesssim\bias(k)+\frac{12A_HL}{\sqrt n}\Bigl(\sqrt{\log(k+1)}+\sqrt{\log(1+r)}+\sqrt{2\log2}+\sqrt{\pi/2}\Bigr)
\]
for every sample.  With an exact program of length \(k_0\) the bound is
minimized at \(\kopt=k_0\); under geometric relevance it is in the EL
regime with \(\kopt\asymp\log n\).
\end{proposition}

\subsection{Computed illustration (\cref{fig:cot-example})}
\label{sec:cot-computed}

\Cref{fig:cot-example} is computed in the following setting.  The alphabet is \(\calA=\{1,2,3\}\cup\{\bullet\}\), so \(r=3\)
active symbols and one padding symbol (drawn in grey); the ultrametric has
\(\theta=1/2\); scratchpads are displayed through their first \(12\)
symbols.

\paragraph{Left panel (append-only steps).}
The input \(x\) is a uniformly random sequence of active symbols, and
\(y\) equals \(x\) except in its first symbol.  Both are run through the
program \(u=(1,2,3,1,2,2)\) of \cref{sec:cot-append}: after \(k\) steps the
scratchpad is \((u_k,\dots,u_1,x_0,x_1,\dots)\).  Each pair of rows shows
the two scratchpads after \(k=0,\dots,6\) steps; the black bar marks
\(n(g_u(x),g_u(y))\), the first position at which they differ, which
equals \(k\).  Hence \(d_\theta(g_u(x),g_u(y))=\theta^k\), the contraction
behind \cref{lem:cot-append-growth}, and the output layers of
\cref{sec:cot-setting} see the two states as \(\theta^k\)-close.

\paragraph{Middle panel (branching steps).}
The upper input starts with the six symbols of \(u\) followed by random
symbols, so every guarded step \(f_{u_j}\) succeeds and consumes its
symbol: after \(k\) steps the scratchpad is \(\sigma^k(x)\) padded with
\(\bullet\) at the end.  The lower input equals the upper one except in
its third symbol, so the third step fails and resets the scratchpad to the
constant state \(\bar u_3\), which the remaining steps turn into
\(\bar u_k\) as in \eqref{eq:cot-branch-closed-form}.  The black bar marks
the first difference between the two rows; it stays at position \(0\)
once the lower input has failed, so the two states are at distance \(1\),
which is the separation used in \cref{lem:cot-branch-growth}.

\paragraph{Right panel (distinguishable programs on a sample).}
A sample of \(n=200\) uniformly random sequences of active symbols of
length \(12\) is drawn.  For each \(k\le10\), all programs of length at
most \(k\) are enumerated (\(\sum_{j\le k}3^j\) programs, the dashed
curve), and two programs are identified when their outputs agree in the
first \(\ell=3\) symbols at every sample point, that is, when their sample
restrictions are within \(\theta^3\) in the metric
\(\max_id_\theta(f(X_i),g(X_i))\).  Since the metric is an ultrametric, the
classes of this relation are the balls of radius \(\theta^3\), so the
number of classes is the covering number of \(B(k,F)\) at resolution
\(\theta^3\) in this sample metric, which dominates the covering number in
\(d_S\).  The count is computed exactly by encoding the first three output
symbols at every sample point as one integer and counting distinct rows.
For branching steps (red squares) the count follows the uniform count
\(3^k\) while \(3^k\le n\) and then grows linearly, by about \(n\) per unit
of depth, as predicted by \cref{lem:cot-branch-sample}: programs whose
success cylinder contains no sample point restrict to one of the \(r\)
constant maps.  For append-only steps (blue circles) the count saturates
at \(\sum_{j\le3}3^j=40\), because programs longer than \(\ell=3\) agree
with a shorter one in the first three symbols
(\cref{lem:cot-append-growth}).  For \(k=10\) the three counts are
\(88573\), \(1184\), and \(40\).

\subsection{Implementation}

Each step is a fixed map, so a recurrently applied block that implements the
step family realizes \(B(k,F)\) with the same block at every depth; the
implementation error of an approximate block is controlled by
\cref{prop:implementation}(b).  For append-only steps the propagation
constant is \(\Lambda=\theta<1\), so a per-step error \(\delta\) yields
\(\errimp(k)\le\delta_H+L_H\delta/(1-\theta)\), uniformly in \(k\).  For
branching steps the propagation constant on the chambers is
\(\Lambda=\theta^{-1}>1\), and the layerwise bound grows like
\(\delta\theta^{-k}\); an exact implementation of the guard is required for
long programs.  Thus the two step families differ numerically, forgetting is stable and
branching is not, while statistically both are benign on any finite sample:
the price of branching is the root-logarithmic profile of
\cref{lem:cot-branch-sample} rather than the saturated one, together with
the need for an exact implementation of the guard.
In summary, both step families act on the same scratchpad space with
fixed linear output layers, so the output-layer constants are
depth-independent, and only the growth of \(B(k,F)\) differs.  For
append-only steps, \(B(k,F_{\rm w})\) saturates in \(d_\infty\) and in
\(d_S\) (\cref{cond:p1}), \(\var(k,n)\asymp n^{-1/2}\), the bias is at most
\(\theta^k\) (proved), and the regime is EL with \(\kopt\asymp\log n\).  For
branching steps, \(r^k\le N(B(k,F_{\rm b}),d_\infty,\eps)\le r^{k+1}\)
(\cref{cond:e2-pingpong}) but
\(N(B(k,F_{\rm b}),d_S,\eps)\le1+r+k\min\{n,\eps^{-2}\}\) on every sample
(\cref{lem:cot-branch-sample}), so \(\var(k,n)\asymp\sqrt{\log k/n}\); the
bias is \(0\) for \(k\ge k_0\) (proved) or \(e^{-\biase k}\) (assumed), and
the balancing depth is \(k_0\), or \(\asymp\log n\) in the EL regime.

\subsection{Relation to the literature}

Chain-of-thought prompting \citep{Wei2022cot} lets a language model produce
intermediate steps before the answer.  Expressivity analyses show that
allowing a number of intermediate steps enlarges the class of problems a
transformer can solve, in particular inherently serial ones
\citep{li2024cot}, and formal frameworks for length generalization
\citep{huang2025lengthgen} and generalization analyses for training
transformers with chain of thought \citep{li2025cot} address specific
architectures.  The present example is complementary and does not concern
a particular architecture: it isolates the statistical cost of the number of
steps as the growth of the compositions of step maps, and shows that this
cost is saturated for forgetful scratchpads and root-logarithmic in the
number of steps for branching computation, independently of how the steps
are implemented; the exponential uniform entropy of branching computation is
not paid on any finite sample.

\section{Worked example: unrolled iterative solvers and samplers}
\label{sec:app-ode}

Many deep models are unrolled iterations: fixed-point solvers, iterative
denoisers, and numerical integrators of ordinary differential equations
(ODEs), including the samplers of score-based diffusion and flow-matching
models.  In each case a hidden layer is one iteration and the depth is the
number of iterations.  This appendix instantiates the framework for two such
families.  The first, fixed-point refinement, is contractive and lands in the
EL regime with saturated variance.  The second, fixed-horizon integration
with a step size that decreases with the depth, is not contractive but
stable, and lands in the PL regime; its variance saturates by an
Arzel\`a--Ascoli argument.  The example is summarized as follows.

\begin{example}[Unrolled iterations]
\label{ex:solvers-main}
The state space is a compact convex \(K\subset\RR^d\), a hidden layer is
one projected step \(x\mapsto\Pi_K(x+h\,s(x))\) with a vector field \(s\)
from a class \(\calS\), and the output layer is linear.  For strongly
monotone fields the profile saturates and the bias decays geometrically,
so \(\kopt\asymp\log n\) and further iterations do not increase the
estimation bound (EL; \cref{prop:ode-fixedpoint}).  For Euler steps of a
stable field over a fixed horizon the profile is \(\sqrt{\log k/n}\) and
the bias is the discretization error \(k^{-p}\), so
\(\kopt\asymp n^{1/(2p)}\) (PL; \cref{prop:ode-horizon}).
\end{example}

\subsection{Setting}
Throughout, \(K\subset\RR^d\) is a compact convex
set with diameter \(D_K\), \(\Pi_K\) is the Euclidean projection onto \(K\),
which maps \(\RR^d\) into \(K\), is the identity on \(K\), and is
\(1\)-Lipschitz, and the output-layer class is
\(H:=\{x\mapsto\langle w,x\rangle:\|w\|\le R\}\) on \(K\), for which
Assumption~\ref{ass:sg-increment-main} holds with \(A_H=1\) and \(L=1\)
(\cref{prop:hilbert-sg}) and \(\erad_S(H)\le R\sup_K\|x\|/\sqrt n\).  Any
fixed Lipschitz feature map may be inserted before the linear layer without
changing the depth dependence.

\subsection{Fixed-point refinement: iterative denoising}
\label{sec:ode-fixedpoint}

\paragraph{Layers.}
Let \(\calS\) be a class of vector fields \(s:K\to\RR^d\) of the form
\(s=\nabla\phi\) with \(\phi\) \(\mu\)-strongly concave and \(\Lambda_s\)-smooth
on \(K\), \(0<\mu\le\Lambda_s\) (for instance the score \(\nabla\log p\) of a strongly log-concave
density, or the residual \(D(x)-x\) of a denoiser \(D\) satisfying the
corresponding monotonicity conditions).  Fix a step size
\(0<h\le 2/(\mu+\Lambda_s)\) and let
\[
F_{\rm fp}:=\bigl\{T_s: x\mapsto\Pi_K\bigl(x+h\,s(x)\bigr)\ :\ s\in\calS\bigr\}.
\]

\begin{lemma}
\label{lem:fp-contraction}
Every \(T_s\in F_{\rm fp}\) is \(\lambda\)-Lipschitz on \(K\) with
\(\lambda:=1-h\mu<1\).  Consequently \cref{cond:p1} applies:
\(\sup_kN(B(k,F_{\rm fp}),d_\infty,\eps)<\infty\) for every \(\eps>0\),
\(D_k(S)\le D_K\), and \(\mathsf V_k(S)\le\mathsf V_\infty<\infty\) for all
\(k\) whenever \(\int_0^{D_K}\sqrt{\log N(\overline{\langle F_{\rm fp}\rangle},d_\infty,\eps)}\,\dd\eps<\infty\).
\end{lemma}

\begin{proof}
Write \(\Delta:=x-y\) and \(\Delta_s:=s(x)-s(y)\).  Since \(-\phi\) is
\(\mu\)-strongly convex and \(\Lambda_s\)-smooth, \(s=\nabla\phi\) is
\(\mu\)-strongly monotone, \(\langle\Delta_s,\Delta\rangle\le-\mu\|\Delta\|^2\),
and co-coercive \citep[Theorem~2.1.12]{Nesterov2004}:
\[
\langle\Delta_s,\Delta\rangle
\le
-\frac{\mu\Lambda_s}{\mu+\Lambda_s}\|\Delta\|^2
-\frac{1}{\mu+\Lambda_s}\|\Delta_s\|^2 .
\]
Strong monotonicity and Cauchy--Schwarz give \(\|\Delta_s\|\ge\mu\|\Delta\|\).
Hence, for \(0<h\le2/(\mu+\Lambda_s)\), so that
\(h^2-2h/(\mu+\Lambda_s)\le0\),
\begin{align*}
\|\Delta+h\Delta_s\|^2
&\le
\Bigl(1-\frac{2h\mu\Lambda_s}{\mu+\Lambda_s}\Bigr)\|\Delta\|^2
+\Bigl(h^2-\frac{2h}{\mu+\Lambda_s}\Bigr)\|\Delta_s\|^2
\\
&\le
\Bigl(1-\frac{2h\mu\Lambda_s}{\mu+\Lambda_s}
+\mu^2h^2-\frac{2h\mu^2}{\mu+\Lambda_s}\Bigr)\|\Delta\|^2
=(1-h\mu)^2\|\Delta\|^2 ,
\end{align*}
and \(\Pi_K\) is \(1\)-Lipschitz; moreover \(0\le1-h\mu<1\) because
\(h\mu\le2\mu/(\mu+\Lambda_s)\le1\).  The remaining claims are
\cref{cond:p1} and \cref{prop:profiles}(i).
\end{proof}

\paragraph{Explicit entropy bound.}
The saturation can be quantified through \cref{cond:p1-ucont} with
\(A=K\) and \(m(\eps)=\lceil\log(2D_K/\eps)/\log(1/\lambda)\rceil\):
\begin{gather*}
N\bigl(B(k,F_{\rm fp}),d_\infty,\eps\bigr)
\le
N(K,\eps/2)+\sum_{j<m(\eps)}N\bigl(F_{\rm fp}^{\,j},d_\infty,\eps\bigr),
\\
N\bigl(F_{\rm fp}^{\,j},d_\infty,\eps\bigr)\le N\bigl(F_{\rm fp},d_\infty,(1-\lambda)\eps\bigr)^{j},
\end{gather*}
where the last inequality follows from the telescoping estimate of
\cref{prop:implementation-restated}(b): replacing each of \(j\) layers by an
\((1-\lambda)\eps\)-close layer changes the composition by at most
\((1-\lambda)\eps\sum_{i<j}\lambda^i\le\eps\).  Hence
\(\log N_\infty(\eps)\lesssim d\log(D_K/\eps)+m(\eps)\log N(F_{\rm fp},d_\infty,(1-\lambda)\eps)\),
and if the one-layer class is parametric with
\(\log N(F_{\rm fp},d_\infty,\eps)\le p\log(C_F/\eps)\), then
\(\log N_\infty(\eps)\lesssim (d+p)\log^2(C/\eps)\), whose square root is
integrable at \(0\).  Thus \(\mathsf V_\infty\lesssim D_K\sqrt{d+p}\) up to
logarithmic factors, independently of \(k\).

\paragraph{Bias.}
Take the target class to be the teacher--student class
\(\Con:=H\circ\overline{\langle F_{\rm fp}\rangle}^{\,d_\infty}\), which
contains in particular the maps \(x\mapsto\langle w,x^\ast_s\rangle\) sending
each input to the fixed point \(x^\ast_s\) of \(T_s\), the limit of the
refinement.  By the truncation argument of \cref{sec:examples-regime}, with
\(\lip(h)\le R\) and \(\lip(u)\le\lambda^k\) for \(u\in F_{\rm fp}^{\,k}\),
\[
\sup_{c\in\Con}\inf_{f\in\calH_k}\|c-f\|_\infty\le R\,D_K\,\lambda^k,
\qquad\text{hence}\qquad
\bias(k)\le\beta_\ell R D_K(1-h\mu)^k
\]
by \eqref{eq:approx-transfer}.

\begin{proposition}[Fixed-point refinement]
\label{prop:ode-fixedpoint}
In the setting above, the excess-risk bound of \cref{thm:bv} is, up to
depth-independent terms,
\(\gen(k,n)\lesssim\beta_\ell RD_K(1-h\mu)^k+12\,\mathsf V_\infty/\sqrt n\),
the EL regime with \(\biase=\log(1/(1-h\mu))\ge h\mu\) and saturated
variance.  It is minimized at
\(\kopt=\frac{\log n}{2\log(1/(1-h\mu))}+O(1)\le\frac{\log n}{2h\mu}+O(1)\),
with balanced value of order \(n^{-1/2}\).
\end{proposition}

The number of refinement iterations should therefore grow logarithmically
with the sample size, at a rate set by the contraction modulus \(h\mu\), and
additional iterations do not increase the estimation bound.  This is the abstract
content of the observation that unrolled denoisers and equilibrium models
can be run to convergence without a statistical penalty
\citep{bai2019deq,romano2017red}; \citet{taniguchi2026picard} develop the
same mechanism for Picard iterations of parabolic equations.

\subsection{Fixed-horizon integration: neural ODEs and diffusion samplers}
\label{sec:ode-horizon}

\paragraph{Layers.}
Let \(\calS_T\) be a class of time-dependent vector fields
\(s:K\times[0,T]\to\RR^d\) with \(\|s\|\le M_s\), \(\Lambda_s\)-Lipschitz in
\(x\), and \(\Lambda_\tau\)-Lipschitz in \(\tau\).  A layer is one explicit
Euler step of size \(h\in[0,h_0]\) at time stamp \(\tau\in[0,T]\):
\[
F_T:=\bigl\{T_{s,h,\tau}:x\mapsto\Pi_K\bigl(x+h\,s(x,\tau)\bigr)\ :\ s\in\calS_T,\ h\in[0,h_0],\ \tau\in[0,T]\bigr\},
\]
and \(\id=T_{s,0,\tau}\in F_T\).  A \emph{scheme} of at most \(k\) steps is a
composition \(T_{s,h_k,\tau_k}\circ\cdots\circ T_{s,h_1,\tau_1}\) with a
single drift \(s\), consistent time stamps \(\tau_i=\sum_{j<i}h_j\), and
total time \(\sum_ih_i\le T\); let \(B_T(k)\subset B(k,F_T)\) be the set of
schemes of at most \(k\) steps, together with \(\id\).  The schemes form a
nested family, \(B_T(k)\subset B_T(k+1)\), by padding with steps of size
\(0\), and the depth-\(k\) hypothesis class is \(H\circ B_T(k)\).  All
statements of \cref{sec:bounds} apply to this subclass with \(F=B_T(k)\) in
\cref{thm:hidden-decomp}.

\begin{lemma}[Stability and saturation]
\label{lem:ode-saturation}
Every scheme in \(\bigcup_kB_T(k)\) is \(e^{\Lambda_sT}\)-Lipschitz on
\(K\).  Consequently \(\bigcup_kB_T(k)\) is equicontinuous on the compact
set \(K\), hence totally bounded in \(d_\infty\) (Ascoli's theorem), and
\[
\sup_{k}N\bigl(B_T(k),d_\infty,\eps\bigr)=:N_\infty(\eps)<\infty
\quad(\eps>0),
\qquad
D_k(S)\le D_K .
\]
If moreover \(\int_0^{D_K}\sqrt{\log N_\infty(\eps)}\,\dd\eps<\infty\), then
\(\mathsf V_k(S)\le\mathsf V_\infty<\infty\) for all \(k\) (case (i) of
\cref{prop:profiles}).
\end{lemma}

\begin{proof}
Each layer is \((1+h_i\Lambda_s)\)-Lipschitz, since \(\Pi_K\) is
\(1\)-Lipschitz and \(s(\cdot,\tau)\) is \(\Lambda_s\)-Lipschitz.  A scheme
is therefore \(\prod_i(1+h_i\Lambda_s)\le\exp(\Lambda_s\sum_ih_i)\le e^{\Lambda_sT}\)-Lipschitz.
A uniformly Lipschitz family of self-maps of a compact metric space is
equicontinuous, and Ascoli's theorem gives total boundedness in \(d_\infty\).
The rest is \cref{prop:profiles}(i).
\end{proof}

No contraction is used: only the stability of the integrator over the fixed
horizon.  Nor is the consistency of the time stamps used: the larger class
of all compositions of at most \(k\) layers from \(F_T\) with a common drift
and \(\sum_ih_i\le T\) is also \(e^{\Lambda_sT}\)-Lipschitz and saturates in
the same way.  The finiteness of \(\int\sqrt{\log N_\infty}\) is a condition
on the one-layer class, and for the schemes actually used by a solver, equal
steps \(h=T/m\), it is explicit.

\begin{lemma}[Explicit entropy of equal-step schemes]
\label{lem:ode-scheme-entropy}
Assume \(0<T\le h_0\).  For \(s\in\calS_T\) and \(m\ge1\) let
\(\Phi_{s,m}:=T_{s,T/m,\tau_m}\circ\cdots\circ T_{s,T/m,\tau_1}\) with
\(\tau_i=(i-1)T/m\) be the \(m\)-step equal-step Euler scheme, \(\Phi_{s,0}:=\id\),
and let \(E_T(k):=\{\Phi_{s,m}:s\in\calS_T,\ 0\le m\le k\}\subset B_T(k)\).
Then for all \(s\in\calS_T\), every drift \(s'\) (not necessarily in
\(\calS_T\)), and \(m\ge0\),
\[
d_\infty(\Phi_{s,m},\Phi_{s',m})\le Te^{\Lambda_sT}\,\|s-s'\|_\infty,
\qquad
\|s-s'\|_\infty:=\sup_{x\in K,\,\tau\in[0,T]}\|s(x,\tau)-s'(x,\tau)\|,
\]
and consequently, for every \(k\ge0\) and \(\eps>0\),
\[
N\bigl(E_T(k),d_\infty,\eps\bigr)
\le
1+k\,N\bigl(\calS_T,\|\cdot\|_\infty,\eps e^{-\Lambda_sT}/T\bigr).
\]
If moreover \(N(\calS_T,\|\cdot\|_\infty,\rho)<\infty\) for all \(\rho>0\)
and the integrand below is integrable, then
\[
\mathsf V_k(S)
\le
D_K\sqrt{\log(k+1)}+\int_0^{D_K}\sqrt{\log N\bigl(\calS_T,\|\cdot\|_\infty,\eps e^{-\Lambda_sT}/(2T)\bigr)}\,\dd\eps
\]
for the hypothesis class \(H\circ E_T(k)\) (the factor \(1/2\) is the
internal/external transfer of \cref{sec:proof-profiles}).  If the drift
class is parametric, \(\log N(\calS_T,\|\cdot\|_\infty,\eps)\le p\log(C/\eps)\),
the last integral is at most
\(D_K\sqrt p\,(\sqrt{\log(2CTe^{\Lambda_sT}/D_K)}+\sqrt\pi/2)\), and the
variance profile is \(O(\sqrt{\log k})\), case (ii) of \cref{prop:profiles}.
\end{lemma}

\begin{proof}
Fix \(x\in K\) and let \(y_i\), \(y_i'\) be the Euler iterates of \(s\) and
\(s'\) started at \(y_0=y_0'=x\), with \(h=T/m\).  Since \(\Pi_K\) is
\(1\)-Lipschitz and \(s(\cdot,\tau)\) is \(\Lambda_s\)-Lipschitz (only the
Lipschitz constant of \(s\) is used),
\[
\|y_i-y_i'\|
\le
\|y_{i-1}-y_{i-1}'\|+h\|s(y_{i-1},\tau_i)-s'(y_{i-1}',\tau_i)\|
\le
(1+h\Lambda_s)\|y_{i-1}-y_{i-1}'\|+h\|s-s'\|_\infty ,
\]
so \(\|y_m-y_m'\|\le h\|s-s'\|_\infty\sum_{i<m}(1+h\Lambda_s)^i
\le mh\,e^{\Lambda_smh}\|s-s'\|_\infty=Te^{\Lambda_sT}\|s-s'\|_\infty\).
This is the telescoping estimate of \cref{prop:implementation}(b) with
per-layer error \(h\|s-s'\|_\infty\) and \(\Lambda=1+h\Lambda_s\).  Hence,
for each \(m\ge1\), the image of an \(\eps e^{-\Lambda_sT}/T\)-cover of
\(\calS_T\) under \(s\mapsto\Phi_{s,m}\) is an \(\eps\)-cover of
\(\{\Phi_{s,m}:s\in\calS_T\}\); adding the identity and summing over
\(1\le m\le k\) gives the covering bound.  Since \(1+kN\le(k+1)N\) for
\(N\ge1\), \(\sqrt{\log(1+kN)}\le\sqrt{\log(k+1)}+\sqrt{\log N}\),
\(N(E_T(k),d_S,\eps)\le N(E_T(k),d_\infty,\eps/2)\), and \(D_k(S)\le D_K\),
integration over \((0,D_K]\) gives the bound on \(\mathsf V_k(S)\).  The
condition \(T\le h_0\) ensures that the equal steps \(T/m\) are admissible,
so that \(E_T(k)\subset B_T(k)\).  The parametric case follows from \cref{lem:log-split}
as in \cref{cor:envelope-profiles}(a).
\end{proof}

The equal-step schemes are nested, \(E_T(k)\subset E_T(k+1)\), so the
balancing argument of \cref{sec:optdepth} applies to \(H\circ E_T(k)\)
verbatim; the estimate is stated for this subclass because a solver with a
prescribed step count uses exactly these schemes.

\paragraph{Bias.}
Let the target be the endpoint map of the exact flow of some
\(s^\ast\in\calS_T\), composed with an output layer:
\(\Con:=\{x\mapsto\langle w,\Phi_T^{s^\ast}(x)\rangle:\|w\|\le R\}\), where
\(\Phi_T^{s^\ast}(x)\) is the solution at time \(T\) of
\(\dot x(t)=s^\ast(x(t),t)\), \(x(0)=x\).  Assume that the exact trajectories
and the Euler trajectories started in \(K\) stay in a region where \(\Pi_K\)
is inactive (for instance \(K\) contains the \(M_sT\)-neighborhood of the
relevant inputs).  The classical global error bound for the explicit Euler
scheme with \(k\) equal steps \(h=T/k\) gives
\[
\sup_{x}\bigl\|\Phi_T^{s^\ast}(x)-T_{s^\ast,h,\tau_k}\circ\cdots\circ T_{s^\ast,h,\tau_1}(x)\bigr\|
\le
C_E\,\frac{T^2}{k},
\qquad
C_E:=\frac{(\Lambda_sM_s+\Lambda_\tau)\,e^{\Lambda_sT}}{2},
\]
so that \(\bias(k)\le\beta_\ell R\,C_ET^2/k\) by \eqref{eq:approx-transfer}.
If the layer class consists instead of one-step methods of order \(p\)
(for example Runge--Kutta steps), the same argument gives
\(\bias(k)\lesssim k^{-p}\).

\begin{proposition}[Fixed-horizon integration]
\label{prop:ode-horizon}
In the setting above, the excess-risk bound of \cref{thm:bv} is, up to
depth-independent terms,
\(\gen(k,n)\lesssim\beta_\ell RC_ET^2k^{-p}+12\,\mathsf V_\infty/\sqrt n\)
with \(p=1\) for Euler layers, the PL regime with \(\biasp=p\) and saturated
variance.  It is minimized at \(\kopt\asymp n^{1/(2p)}\), with balanced
value of order \(n^{-1/2}\).
\end{proposition}

The number of integration steps should therefore grow polynomially with the
sample size, while the statistical cost of the steps stays bounded.  This is
the depth trade-off of neural ODEs \citep{Chen2018a} and of stable
residual architectures \citep{Haber2017,E2017}, read in the state-transition
model: depth refines the discretization and does not enlarge the effective
hypothesis class beyond the equicontinuous family of stable schemes.

\paragraph{Diffusion and flow-matching samplers.}
The probability-flow ODE of a score-based diffusion model
\citep{song2021sde,ho2020ddpm} and the velocity field of a flow-matching
model \citep{lipman2023flowmatching} are time-dependent vector fields
\(s(x,\tau)\) whose learned approximants form the drift class \(\calS_T\);
a sampler with \(k\) steps is a scheme in \(B_T(k)\), and the number of
function evaluations is the depth.  When the learning task is the regression
of the endpoint map, for instance predicting a coordinate of the generated
sample from the initial noise, or the reconstruction of clean data along the
deterministic flow, \cref{prop:ode-horizon} applies verbatim under the
Lipschitz and boundedness conditions on the learned drifts.  Generative
quality is usually measured by a distance between distributions rather than
by a regression loss, and the present analysis does not address that
criterion; it addresses the statistical cost of the number of steps for the
learned map.

\begin{figure}[t]
    \centering
    \includegraphics[width=\linewidth]{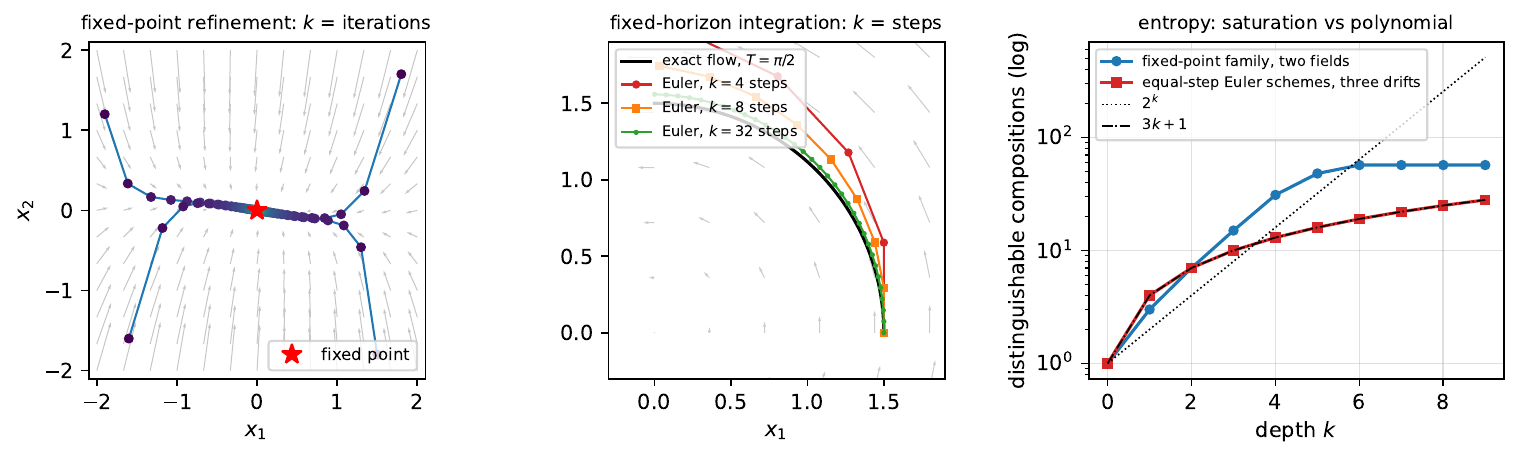}
    \caption{Unrolled solvers on \(K=[-2,2]^2\).  Left: fixed-point refinement, four trajectories (dots, one per iteration) and the fixed point (star).  Middle: Euler integration of a rotation field with \(4\), \(8\), and \(32\) steps (circles, squares, dots) and the exact flow (black).  Right: distinguishable compositions of depth at most \(k\) at resolution \(2\cdot10^{-2}\) for a fixed-point family with two fields (blue circles) and for equal-step Euler schemes with three drifts (red squares), with the references \(2^k\) (dotted) and \(3k+1\) (dash-dotted).  See \cref{sec:ode-computed}.}
    \label{fig:ode-example}
\end{figure}

\subsection{Computed illustration (\cref{fig:ode-example})}
\label{sec:ode-computed}

The two cases differ in one step.  A strongly monotone field, as the
score of a strongly log-concave density or the residual of a denoiser,
makes every layer a contraction, so the entropy of \(B(k,F)\) saturates
by \cref{thm:mechanisms-main}(ii) and the bias decays like \((1-h\mu)^k\)
because the iteration converges geometrically to its fixed point; a
denoiser or a deep equilibrium model may therefore be run to convergence
without a statistical penalty.  An Euler step of a stable but not
contractive field over a fixed horizon, as in a neural ODE or the sampler
of a diffusion model, keeps the compositions uniformly Lipschitz, so the
entropy grows only logarithmically in the number of steps, while the bias
is the discretization error of the scheme.  \Cref{fig:ode-example} shows
both families on \(K=[-2,2]^2\): fixed-point refinement of a strongly
monotone field (left, four trajectories converging to the fixed point),
Euler integration of a rotation field over a fixed horizon with \(4\),
\(8\), and \(32\) steps against the exact flow (middle), and the
distinguishable compositions of the two families at resolution
\(2\cdot10^{-2}\) (right): the fixed-point count follows \(2^k\) until the
contraction takes the compositions below the resolution and then
saturates, and the count of the equal-step Euler schemes is \(3k+1\),
linear in the depth (details below).

\Cref{fig:ode-example} is computed without randomness; its three panels
use the following setting.

\paragraph{Fixed-point refinement (left panel).}
The state space is \(K=[-2,2]^2\) with the Euclidean metric and
\(\Pi_K\) the coordinatewise clipping to \([-2,2]\).  The vector field is
\[
s(x)=-\bigl(x_1+\tfrac12\sin x_2,\ 4x_2+\tfrac12\sin x_1\bigr).
\]
Its Jacobian is
\(-\bigl[\begin{smallmatrix}1&\frac12\cos x_2\\ \frac12\cos x_1&4\end{smallmatrix}\bigr]\),
whose symmetric part has eigenvalues between \(0.92\) and \(4.08\), so
\(s\) is \(\mu\)-strongly monotone and \(\Lambda_s\)-Lipschitz with
\(\mu=0.9\) and \(\Lambda_s=4.1\); the contraction estimate of
\cref{lem:fp-contraction} holds for such fields, since its proof uses only
strong monotonicity and co-coercivity, which follow from these bounds on
the symmetric part.  The step size is \(h=1/(\mu+\Lambda_s)=0.2\), inside
the admissible range \((0,2/(\mu+\Lambda_s)]\), so every layer is
\((1-h\mu)=0.82\)-Lipschitz.  Four initial states,
\((1.8,1.7)\), \((-1.9,1.2)\), \((1.5,-1.8)\), \((-1.6,-1.6)\), are
iterated \(k=40\) times, and the fixed point \(x^\ast\) is computed by
\(400\) iterations from the first initial state; the largest distance
\(\max_i\|x^{(i)}_k-x^\ast\|\) over the four trajectories, the quantity
bounded by \(D_K\lambda^k\) in \cref{prop:ode-fixedpoint}, decays
geometrically at a rate close to \(1-h\mu\).

\paragraph{Fixed-horizon integration (middle panel).}
The vector field is the rotation \(s(x)=(-x_2,x_1)\), which is
\(\Lambda_s=1\)-Lipschitz, bounded by \(M_s=2\sqrt2\) on \(K\), and not
contractive; its exact flow over time \(t\) is the rotation by the angle
\(t\).  The horizon is \(T=\pi/2\), and the scheme is the explicit Euler
method with \(k\) equal steps \(h=T/k\), the equal-step scheme
\(\Phi_{s,k}\) of \cref{lem:ode-scheme-entropy}; the projection is inactive
along the displayed paths.  The panel shows the paths from the initial
state \((1.5,0)\) for \(k=4,8,32\) and the exact quarter circle; the
endpoint error \(\max_i\|\Phi_{s,k}(x^{(i)})-\Phi_T^{s}(x^{(i)})\|\) over the
three initial states \((1.5,0)\), \((1,0.5)\), \((0.6,-1.2)\) decays like
\(k^{-1}\), the global error of the Euler method,
which is the bias of \cref{prop:ode-horizon} with \(p=1\).

\paragraph{Distinguishable compositions (right panel).}
The right panel compares the entropy of the two families in the same way
as \cref{fig:relu-example,fig:cot-example}: it counts the compositions of
depth at most \(k\) that are distinguishable at resolution
\(\eps=2\cdot10^{-2}\), that is, the size of a greedy \(\eps\)-net in the
metric \(\max_i\|f(x_i)-g(x_i)\|\) over the \(21\times21\) grid of points
\(x_i\) of \(K\), for \(k\le9\).  The fixed-point family has two layers,
\(x\mapsto\Pi_K(x+h's_j(x))\) with the strongly monotone fields
\(s_1(x)=-(2x_1+\tfrac12\sin x_2,\,4x_2+\tfrac12\sin x_1)\) and
\(s_2(x)=-(4x_1+\tfrac12\cos x_2,\,2x_2+\tfrac12\cos x_1)\), whose
symmetric Jacobians have eigenvalues between \(1.9\) and \(4.1\), and the
step \(h'=2/(1.9+4.1)=1/3\), so that every layer is \(0.37\)-Lipschitz; all
\(2^k\) words of length \(k\) are formed.  The count is
\(1,3,7,15,31,48,57\) for \(k\le6\) and stays at \(57\) afterwards: the
words follow \(2^k\) until the contraction \(0.37^k\) falls below the
resolution, and then saturate at the entropy of the attractor of the two
fields, which is mechanism P1' (\cref{lem:fp-contraction}).  The
fixed-horizon family is the class \(E_T(k)\) of \cref{lem:ode-scheme-entropy}
with the drift class \(\{\omega\,(-x_2,x_1):\omega\in\{0.8,1,1.2\}\}\),
that is, the equal-step Euler schemes with \(m\le k\) steps and one of
three angular speeds.  The count is \(3k+1\): one map per pair (drift,
number of steps), all distinguishable at this resolution, which is the
linear growth \(1+k\,N(\calS_T,\cdot)\) of \cref{lem:ode-scheme-entropy}
and hence the \(\sqrt{\log k}\) profile.  The two curves are the two
variance profiles of the examples, saturated and logarithmic; the two
bias laws, geometric for the contractive iteration and \(k^{-1}\) for the
Euler scheme, are computed in the same setting and are the other side
of the balance in \cref{sec:optdepth}.

\subsection{Implementation}
For both families, \cref{prop:implementation}(b) controls the implementation
error of approximate layers.  For fixed-point refinement,
\(\Lambda=1-h\mu<1\) and \(\errimp(k)\le\delta_H+R\delta/(h\mu)\) uniformly
in \(k\).  For fixed-horizon schemes, \(\Lambda_i=1+h_i\Lambda_s\) and the
telescoping sum is bounded by \(\delta\,k\,e^{\Lambda_sT}\), so
\(\errimp(k)\le\delta_H+R\delta ke^{\Lambda_sT}\) grows only linearly in the
number of steps; with \(\delta\) the per-step accuracy of the learned drift
this is the usual stability of one-step methods.  In both cases the
implementation error does not change the regime of the depth trade-off.

\subsection{Relation to the literature}
Unrolled fixed-point iterations are the model of deep equilibrium
networks \citep{bai2019deq} and of regularization by denoising
\citep{romano2017red}, and Picard-type iterations of the same kind
underlie the neural operators analyzed by \citet{taniguchi2026picard}.
Fixed-horizon integration is the model of neural ODEs \citep{Chen2018a}
and of stable residual architectures \citep{Haber2017,E2017}, and the
samplers of score-based diffusion and flow-matching models
\citep{song2021sde,ho2020ddpm,lipman2023flowmatching} are its instances
with a learned drift.  In each case the depth is the number of iterations
or steps; the present appendix supplies the statistical side of the
question of how many of them to use, which those works treat as a
numerical or empirical choice.

\end{document}